\documentclass[11pt]{article}

\usepackage[final]{acl}

\usepackage{times}
\usepackage{latexsym}

\usepackage[T1]{fontenc}
\usepackage[utf8]{inputenc}

\usepackage{microtype}

\usepackage{inconsolata}

\usepackage{graphicx}

\usepackage{amsmath}
\usepackage{enumitem}
\usepackage{booktabs}

\usepackage{cuted}
\usepackage{capt-of}
\usepackage{fontawesome5}   % \faGithub logo for the resource links
\definecolor{rtLow}{HTML}{1BAF7A}   % category 1, low starting loss (green)
\definecolor{rtFast}{HTML}{2A78D6}  % category 2, faster drop (blue)
\definecolor{rtSlow}{HTML}{EB6834}  % category 3, slower drop (orange)
\newcommand{\low}[1]{{\color{rtLow!80!black}#1}}
\newcommand{\fast}[1]{{\color{rtFast!80!black}#1}}
\newcommand{\slow}[1]{{\color{rtSlow!80!black}#1}}
\definecolor{rtModerate}{HTML}{9B4F96} % moderate drop (plum); CVD-distinct from the other three (worst-case dE 11.2)
\newcommand{\moderate}[1]{{\color{rtModerate!80!black}#1}}
\title{Predicting Memorization Before Fine-Tuning Language Models}

\author{
  \textbf{Jaydeep Borkar\textsuperscript{1}},
  \textbf{Niloofar Mireshghallah\textsuperscript{2}}, 
  \textbf{David A. Smith\textsuperscript{1}}
\\
  \textsuperscript{1}Northeastern University,
  \textsuperscript{2}Carnegie Mellon University 
\\
 \small{
   \textbf{Correspondence:} \href{mailto:borkar.j@northeastern.edu}{borkar.j@northeastern.edu}
 }
 \\[5pt]
 \normalsize{
   \faGithub~\href{https://github.com/jaydeepborkar/predicting-memorization}{code}
   \qquad
   \raisebox{-0.18em}{\includegraphics[height=1em]{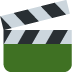}}~\href{https://huggingface.co/spaces/jaydeepb/predicting_memorization}{demo}
   \qquad
   \raisebox{-0.18em}{\includegraphics[height=1em]{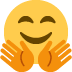}}~\href{https://huggingface.co/collections/jaydeepb/predicting-memorization-fine-tuned-models}{models}
 }
}

\begin{document}
\maketitle
%\vspace{-8pt}
\begin{strip}
\centering
\includegraphics[width=\textwidth]{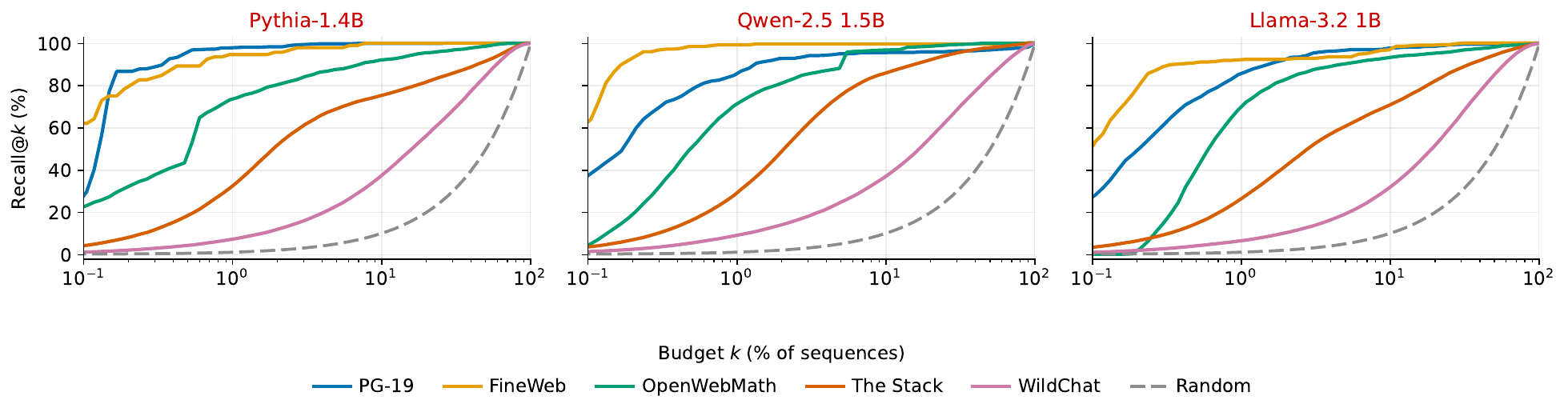}
\captionof{figure}{Our classifier flags memorization cheaply. We show recall@$k$ on the held-out dataset of 1M sequences. Recall@$k$ is the fraction of memorized sequences found among the top $k\%$ ranked by our classifier. On most datasets, inspecting a small top fraction already recovers most memorization, and recall stays far above the random baseline (recall@$k = k$). The $x$-axis is log-scaled and, since $k$ is a percentage, goes to 100.}
\label{fig:recallk}
\end{strip}

\begin{abstract}
Language models are known to memorize and later regurgitate their training data, raising serious concerns when it is sensitive. Fine-tuning data is often private and proprietary, so a practitioner benefits from knowing which sequences will be memorized before training begins. Existing methods for predicting fine-tuning memorization still require partially training the target model. We show that memorization in a target model can be predicted without fine-tuning that model at all. We compute features from the pretrained base model and fit a classifier on memorization labels from a single fine-tuning run to predict the sequences future runs will memorize. This classifier reaches an ROC-AUC of at least 0.9 in most cases across six domains (FineWeb, Enron, PG-19, The Stack, OpenWebMath, WildChat) and three model families (Pythia, Qwen-2.5, Llama-3.2). Moreover, a classifier trained on a single domain, model family, \& model size predicts memorization in unseen domains, families, and sizes. We further find that the \emph{learnability} of a sequence shapes how it gets memorized. Finally, we study \emph{scaling laws for memorization} on a Qwen-2.5 ladder from 0.5B to 14B, and find that both the memorization rate and our classifier's precision follow a power law in model size, with memorization growing as $N^{0.39}$ to $N^{0.83}$.

\end{abstract}

\section{Introduction}
\label{sec:intro}

Large language models (LLMs) are known to memorize and regurgitate portions of their training data verbatim~\citep{gpt2_extraction}. This is problematic when the memorized data is sensitive, such as personal information~\citep{panda2024teach,scalable_extraction}, proprietary data~\citep{wang2025verileakynavigatingipprotection}, or copyrighted material~\citep{karamolegkou-etal-2023-copyright, chen-etal-2024-copybench, cooper2026extractingmemorizedpiecescopyrighted, alignmentwhackamolefinetuning}. An extensive body of work has studied memorization empirically in both pretraining~\citep{tirumala2022memorization, counterfactual, emergent_memorization} and fine-tuning settings~\citep{mireshghallah-etal-2022-empirical, zeng-etal-2024-exploring, code_finetuning}. This line of work has characterized how memorization manifests during training~\citep{jagielski2023measuring, borkar-etal-2025-privacy}, quantifying its extent~\citep{quantifying_memorization, morris2026how}, and categorizing the different forms it takes~\citep{ippolito-etal-2023-preventing, prashanth2025recite}. Studying memorization after it has occurred has given us important learnings on factors that drive memorization, better ways to measure it, and the consequences of it. However, from a practical point of view, it is of limited utility for a practioner who may want to know what sequences will be memorize before starting the training. The question of predicting memorization before training finishes is an open important question~\citep{biderman2026positiondontjustfix, potts2026predictingmemorization}. We study this question in a fine-tuning setting: \emph{given only the pretrained base model and a fine-tuning dataset, can we predict which sequences will be memorized before fine-tuning begins?}  

We show that memorization in fine-tuned language models can be predicted \emph{a priori}. We train a classifier on memorization labels from a single fine-tuning run and use it to predict which sequences new fine-tuning runs on unseen data will memorize. The classifier uses only features computed from the corresponding pretrained base model, so the target model never needs to be fine-tuned. We focus on fine-tuning because practitioners fine-tune models far more often than they pretrain them. Furthermore, fine-tuning datasets are typically small and fully under their control, so flagged sequences can be inspected or removed before training. In contrast, existing approaches for predicting fine-tuning memorization still require partially training the target model~\citep{slack-al-moubayed-2025-early, dentan2025predictingmemorizationlargelanguage}, and in pretraining, memorization has proven hard to forecast even from smaller models~\citep{emergent_memorization}.

We consider three classes of features. The first depends on the data alone and captures how compressible a sequence is~\citep{gpt2_extraction}. The second comes from a forward pass through the base model and captures how surprising a sequence is~\citep{secret-sharer,gpt2_extraction}. The third comes from a backward pass and captures how easily the base model fits a sequence, since gradient-based signals are known to reflect how examples are learned~\citep{data_diet} and memorized sequences are ones the model fits most easily~\citep{borkar2026memorizationdynamicsknowledgedistillation}. We measure compressibility with zlib entropy~\citep{zlib}, surprise with perplexity-based features (perplexity and per-token loss variance), and ease of fitting with gradient-based features (gradient norm and one-step loss change). We conduct our study across six datasets covering various domains, such as email~\citep[Enron:][]{enron}, personal/conversational text~\citep[WildChat:][]{wildchat}, books~\citep[PG-19:][]{pg-19}, code~\citep[The Stack:][]{TheStack}, math~\citep[OpenWebMath:][]{openwebmath}, and general web text~\citep[FineWeb:][]{fineweb}. We experiment with three model families---Pythia 1.4B~\citep{pythia}, Qwen-2.5 1.5B~\citep{qwen2025qwen25technicalreport}, and Llama-3.2 1B~\citep{llama-3.2}---and, for our scaling analysis, additionally use the Qwen-2.5 family across sizes from 0.5B to 14B. We report the following findings:

\begin{enumerate}
    \item Using only signals from the pretrained base model, we can predict which examples will be memorized prior to finetuning with high success rate (AUC of at least 0.9 in the majority of cases across datasets and models). We also find that removing the sequences predicted as most likely to be memorized from the fine-tuning dataset substantially reduces memorization, with almost no loss in model utility (Section~\ref{sec:memorization_is_predictable}). 
    \item We study memorization through the lens of \emph{learnability} and find that not all sequences are memorized the same way. Some already have a low loss under the base model, some start with a high loss that drops quickly after a single gradient step, and others start with a high loss that drops slowly, which makes them the hardest to predict (Section~\ref{sec:learnability}).
     \item A classifier trained to predict memorization in one domain, model family, and model size can predict memorization in other domains, model families, and model sizes (Section~\ref{sec:generalization}).
    \item We study how memorization scales with model size on a Qwen-2.5 ladder from 0.5B to 14B, and find that both the memorization rate and the precision of our predictor follow a power law (Section~\ref{sec:scaling_laws}).
\end{enumerate}

We release our fine-tuned models, code, trained classifiers, and an interactive demo for this paper.

\section{Related Work}
\label{sec:related_work} 

\paragraph{Memorization in Language Models}: as coined by ~\citet{secret-sharer}, refers to the phenomenon when language models reproduce text that they memorized from their training data. This finding has led to a large-scale study to understand memorization in various stages of modern LLM training. Notably, a lot of prior works have studied general memorization in pre-trained base models~\cite{gpt2_extraction, quantifying_memorization, tirumala2022memorization,counterfactual, ippolito-etal-2023-preventing, scalable_extraction, hayes-etal-2025-measuring}. Extraction of memorized data can have consequences for privacy~\citep{carlini2022the,borkar-etal-2025-privacy, furukawa2026reconstructionpersonallyidentifiableinformation} and copyright~\citep{karamolegkou-etal-2023-copyright, chen-etal-2024-copybench, alignmentwhackamolefinetuning, cooper2026extractingmemorizedpiecescopyrighted}. Memorization during fine-tuning has also been well-studied, including its susceptibility to different fine-tuning methods~\citep{mireshghallah-etal-2022-empirical}, understanding it across different fine-tuning tasks~\citep{zeng-etal-2024-exploring}, \& analyzing the nature of the data that can be extracted~\citep{borkar2023learndataleakageunlearning}.

\paragraph{Predicting memorization}: In the pretraining setting, \citet{emergent_memorization} find that memorization in a large pretrained model is difficult to forecast reliably from smaller models or earlier checkpoints. In the same setting, \citet{dontforgetdecomposingtraining} decompose each training step and predict memorization from features such as how well an example's own gradient update aligns with lowering its loss, though these features are computed from model checkpoints collected during training.

In the fine-tuning setting, existing approaches still require some amount of training of the target model before they can make a prediction. \citet{dentan2025predictingmemorizationlargelanguage} study predicting memorization for discriminative classification tasks, but only after fine-tuning until the median loss has dropped by 95\%, while \citet{slack-al-moubayed-2025-early} use an \emph{n}-gram memorization signal as an early-stopping criterion during instruction tuning. A related line of work identifies factors correlated with memorization: \citet{borkar-etal-2025-privacy} link \emph{n}-gram overlap to the memorization of emails, and \citet{borkar2026memorizationdynamicsknowledgedistillation} study predictive factors in distilled models trained from a larger teacher with a KL-divergence objective. To the best of our knowledge, none of these works predict memorization in a to-be-fine-tuned target model from base-model features and labels from a single fine-tuning run, before any training of the target.

\section{Experimental Setup}
\label{sec:exp_setup}
Let $M$ be a pretrained base model and $d$ a fine-tuning dataset consisting of sequences $x \in d$. Let $f(x; M)$ denote a feature vector computed for sequence $x$ using only the unmodified base model $M$, without any fine-tuning. Let $y_x \in \{0, 1\}$ denote whether sequence $x$ is memorized after fine-tuning $M$ on $d$. We train a classifier $C$ to predict $y_x$ from $f(x; M)$, using labels $\{y_x\}_{x \in d}$ obtained from a single fine-tuning run of $M$ on a 1M-sequence sample $d$. We evaluate $C$ under four setups, namely (i) 5-fold cross-validation within $d$; (ii) on a disjoint, independently drawn 1M-sequence sample $d'$ from the same source dataset (e.g., a second, non-overlapping 1M-sequence sample from FineWeb); (iii) on a fine-tuning run from a different source dataset entirely (e.g., a classifier trained on FineWeb, evaluated against a run on Enron); and (iv) on a fine-tuning run from a different model family (e.g., applying a classifier trained on the (Pythia, FineWeb) run to a Qwen or Llama model fine-tuned on FineWeb). Next, we detail our fine-tuning, memorization measurement, features, and classifier evaluation setups.

\paragraph{Fine-tuning Setup} We fine-tune each base model checkpoint on a total of 1M sequences with a sequence length of 128 tokens. We use the AdamW optimizer~\citep{loshchilov2018decoupled} with a learning rate of $5 \times 10^{-5}$, cosine decay~\citep{loshchilov2017sgdr}, and a linear warmup of 50 steps. We apply a weight decay of $1 \times 10^{-2}$. We use a per-device batch size of 32 with gradient accumulation of 32 steps and fine-tune for one epoch. We fine-tune three base models, Pythia 1.4B, Qwen 2.5 (0.5B to 14B), and Llama 3.2 1B, on each of six datasets: Enron, WildChat, FineWeb, PG-19, The Stack, and OpenWebMath.

\paragraph{Measuring Memorization} We use the definition of discoverable memorization~\citep{quantifying_memorization, scalable_extraction}, where we give the first 50 tokens as a prefix to the model and generate the next 50 tokens using greedy decoding. If the generation matches the ground truth, the sequence is considered memorized. Some fine-tuning sequences can already be reproduced verbatim by the base model before fine-tuning~\citep{prashanth2025recite, liu2025language}. Such sequences would be labeled as memorized after fine-tuning even though fine-tuning did not cause it. Since we aim to predict memorization caused by fine-tuning, we remove every memorized sequence that the base model already reproduces. We remove these sequences from both the classifier's training and evaluation sets in our main experiments (Sections~\ref{sec:memorization_is_predictable}, \ref{sec:learnability}, and~\ref{sec:scaling_laws}), and call this the contamination-removed setup.   

\paragraph{Features} We categorize our features into three categories based on what properties of a sequence they capture: 1) how compressible a sequence is, 2) how surprising it is to the base model, and 3) how easily the model fits it.
\begin{enumerate}[leftmargin=*, topsep=0pt, itemsep=2pt]
    \item \textbf{Text-based features} we use zlib entropy~\citep{zlib} to measure the compressibility of the sequence $x$. Prior work also associates compressibility with memorization~\citep{gpt2_extraction, prashanth2025recite}
    \item \textbf{Loss-based features} need only a forward pass through the base model with parameters $\theta$:
    \begin{enumerate}
        \item \textbf{perplexity} measures how surprising $x$ is to the base model~\citep{secret-sharer, gpt2_extraction}, $\text{PPL}(x) = \exp\left(\frac{1}{T}\sum_{t=1}^{T} -\log p_\theta(x_t \mid x_{<t})\right)$, and
        \item \textbf{base-loss variance} measures how uniformly that surprise is distributed across tokens, $\text{Var}_t[\ell_t]$ where $\ell_t = -\log p_\theta(x_t \mid x_{<t})$.
    \end{enumerate}
    \item \textbf{Gradient-based features} additionally require a backward pass:
    \begin{enumerate}
        \item \textbf{gradient norm} measures the magnitude of the per-example gradient, $\|\nabla_\theta \mathcal{L}(\theta; x)\|_2$;
        \item \textbf{one-step absolute loss change} measures how much a single gradient update on $x$ reduces the model's loss on $x$ itself, $\Delta\mathcal{L}(x) = \mathcal{L}(\theta; x) - \mathcal{L}(\theta - \eta\nabla_\theta \mathcal{L}(\theta; x); x)$; and
        \item \textbf{one-step relative loss change} normalizes this by the initial loss, $\Delta\mathcal{L}_{\text{rel}}(x) = \Delta\mathcal{L}(x) / \mathcal{L}(\theta; x)$.
    \end{enumerate}
\end{enumerate}

To compute the one-step loss changes, we take a single plain gradient descent step on the base model with a learning rate of $\eta = 2\times10^{-5}$. The loss $\mathcal{L}$ is the cross-entropy on the 50-token continuation given the 50-token prefix, which are the same tokens we use to measure memorization. We probe each sequence separately starting from the same base model weights every time. After measuring the new loss, we discard the update, so probing one sequence does not affect any other.

\begin{table*}[t]
\centering\small
\begin{tabular}{lccccc}
\toprule
\multicolumn{6}{c}{\textit{Pythia-1.4B}} \\
\cmidrule(lr){1-6}
 & & \multicolumn{4}{c}{5-fold CV ROC-AUC} \\
\cmidrule(lr){3-6}
Dataset & Mem.\ rate & Text & Loss & Grad. & All \\
\midrule
FineWeb     & 0.015\% & $0.615 \pm 0.039$ & $0.962 \pm 0.001$ & $0.996 \pm 0.003$ & $0.997 \pm 0.004$ \\
PG-19       & 0.053\% & $0.603 \pm 0.024$ & $0.995 \pm 0.001$ & $0.999 \pm 0.000$ & $0.999 \pm 0.000$ \\
OpenWebMath & 0.31\%  & $0.738 \pm 0.006$ & $0.927 \pm 0.002$ & $0.950 \pm 0.004$ & $0.983 \pm 0.002$ \\
The Stack   & 2.56\%  & $0.633 \pm 0.001$ & $0.897 \pm 0.003$ & $0.932 \pm 0.001$ & $0.944 \pm 0.000$ \\
WildChat    & 4.09\%  & $0.777 \pm 0.004$ & $0.865 \pm 0.005$ & $0.822 \pm 0.006$ & $0.900 \pm 0.001$ \\
Enron       & 4.59\%  & $0.657 \pm 0.004$ & $0.970 \pm 0.001$ & $0.980 \pm 0.002$ & $0.985 \pm 0.001$ \\
\bottomrule
\end{tabular}
\caption{Predicting fine-tuning memorization from base model features (deduplicated setup). We report 5-fold cross-validation ROC-AUC by feature categories: Text (zlib entropy), Loss (perplexity and loss variance), Grad.\ (gradient norm and one-step loss change), and All (combined), for Pythia-1.4B. Values are mean $\pm$ standard deviation over three fine-tuning seeds. Results for Qwen-2.5 1.5B and Llama-3.2 1B are in Table~\ref{tab:main_qwen_llama} (Appendix~\ref{app:main_qwen_llama}). Random chance AUC is 0.5.}
\label{tab:main_pythia}
\end{table*}

\paragraph{Classifier Training and Evaluation} We train a GradientBoostingClassifier~\citep{friedman2000greedy} from scikit-learn~\citep{scikit-learn} on the extracted features to predict whether a sequence will be memorized. Since memorization rates vary substantially across datasets, we construct a balanced training set using a fixed 1:3 ratio of memorized to non-memorized examples, sampling all available memorized sequences per dataset along with three times as many randomly sampled non-memorized sequences. As our main setup, we additionally remove any sequence-level exact and approximate duplicates from the classifier's data identified via MinHash-LSH~\citep{minhash_1, minhash_2} (Jaccard Similarity $\geq$ 0.8). Because duplication is a well-established driver of memorization~\citep{lee-etal-2022-deduplicating, kandpal22a, huang-etal-2024-demystifying}, this deduplicated setup isolates our ability to predict \emph{nontrivial} memorization that is not explained by duplication. Unless stated otherwise, all reported results use this deduplicated setup, with the non-deduplicated results reported in Appendix~\ref{app:duplicates}. For evaluation, we report ROC-AUC for cross-validation, cross-domain, and cross-model setups. For evaluating our classifier on a disjoint and independently drawn 1M sequence sample, we keep duplicates so that the evaluation reflects the data a practitioner would actually inspect. We report recall@$k$ (the fraction of memorized sequences recovered within a top-$k$ inspection budget) and precision@$k$ (the fraction of the top $k\%$ that is truly memorized). This setup simulates the practical setting where a practitioner can only inspect a fixed number of sequences and must decide which ones to audit for memorization.

\section{Memorization is Predictable}
\label{sec:memorization_is_predictable}

We find that memorization follows a structured and predictable pattern rather than being stochastic. By using only features that we compute from the base model and the training data, we predict which sequences a target fine-tuned model will memorize. Under 5-fold cross-validation, the combined feature set attains ROC-AUC between 0.900 (WildChat) and 0.999 (PG-19) across all six domains for Pythia-1.4B (Table~\ref{tab:main_pythia}), far above the 0.5 random chance. This margin is important, especially as memorization isn't often very high. For instance, in our setup, memorization rates range from 0.015\% (FineWeb) to 4.6\% (Enron). The results are stable across three independent fine-tuning seeds (standard deviation $\le 0.003$). The same pattern holds across Qwen-2.5 1.5B and Llama-3.2 1B, where they show comparable AUCs on all six datasets (Table~\ref{tab:main_qwen_llama}). 

Predictability also holds across model scale, with the cross-validation AUC staying above 0.8 across the Qwen-2.5 ladder from 0.5B to 14B, although it declines for the 7B and 14B models (Table~\ref{tab:ladder_auc}). In the non-deduplicated setup (Appendix~\ref{app:duplicates}), where duplicate counts are available as features, the text and duplicate-count features alone are strongly predictive on the datasets where the model-based features are weakest. They reach about 0.95 AUC on WildChat for all three families, and about 0.88 on Enron for Qwen-2.5 1.5B and Llama-3.2 1B (Tables~\ref{tab:dup_pythia} and~\ref{tab:dup_qwen_llama}). This suggests that memorization in conversational and email data is driven largely by duplicated and templated text, a well-established driver of memorization~\citep{lee-etal-2022-deduplicating, kandpal22a, huang-etal-2024-demystifying}. Finally, we find that our results also transfer to a knowledge distillation setup (Appendix~\ref{app:distillation}) and to looser forms of memorization such as near-verbatim memorization (Appendix~\ref{app:approx_mem}).

\subsection{Flagging Memorization}
\label{subsec:cheaply_flagging}

\begin{table*}[t]
\centering\small
\setlength{\tabcolsep}{3pt}
\begin{tabular}{@{}l*{5}{c}@{\hspace{12pt}}*{5}{c}@{\hspace{12pt}}*{5}{c}@{}}
\toprule
 & \multicolumn{5}{c}{\textit{Pythia-1.4B}} & \multicolumn{5}{c}{\textit{Qwen-2.5 1.5B}} & \multicolumn{5}{c}{\textit{Llama-3.2 1B}} \\
\cmidrule(lr){2-6}\cmidrule(lr){7-11}\cmidrule(lr){12-16}
Dataset & $0.1\%$ & $0.5\%$ & $1\%$ & $5\%$ & $10\%$ & $0.1\%$ & $0.5\%$ & $1\%$ & $5\%$ & $10\%$ & $0.1\%$ & $0.5\%$ & $1\%$ & $5\%$ & $10\%$ \\
\midrule
FineWeb     & 62.0 & 89.1 & 94.6 & 97.8 & 100.0 & 60.7 & 98.4 & 99.2 & 99.6 & 99.6 & 52.2 & 90.5 & 92.3 & 93.5 & 97.3 \\
PG-19       & 26.8 & 95.5 & 97.9 & 99.7 & 99.8 & 37.3 & 78.3 & 85.4 & 94.9 & 95.5 & 27.5 & 73.6 & 85.4 & 96.6 & 97.7 \\
OpenWebMath & 22.6 & 43.8 & 73.5 & 87.6 & 92.1 & 4.4 & 51.0 & 71.5 & 88.1 & 96.7 & 0.0 & 41.0 & 70.3 & 90.2 & 93.2 \\
The Stack   & 4.2 & 18.5 & 32.8 & 69.0 & 75.3 & 3.5 & 16.4 & 29.3 & 75.6 & 86.0 & 3.3 & 14.7 & 26.5 & 60.8 & 70.9 \\
WildChat    & 1.0 & 4.2 & 7.2 & 23.1 & 37.5 & 1.3 & 5.5 & 9.0 & 24.8 & 37.1 & 0.9 & 4.0 & 6.5 & 19.2 & 32.0 \\
\bottomrule
\end{tabular}
\caption{\textbf{Flagging Memorization}: Recall@$k$ (\% of all memorized sequences contained in the top $k\%$ highest-scored) at inspection budgets $k$ for Pythia-1.4B, Qwen-2.5 1.5B, and Llama-3.2 1B on the held-out 1M set at natural prevalence. Each classifier is first trained on (memorization outcome, features) independent fine-tuning run and then evaluated on second run on a disjoint dataset. Since $k$ is the percentage of the results inspected, the random baseline for recall@$k$ is $k$. Enron is omitted as its corpus is exhausted by training.}
\label{tab:recallk}
\end{table*}

In the previous section, we found that it is possible to predict memorization with a high AUC score. This predictability can provide a very practical use case. Consider a practitioner about to fine-tune a model on millions of sequences who would like to know which sequences are likely to be memorized before training begins. We therefore use the classifier to \emph{prioritize} the sequences most-likely to get memorized from a large dataset. We score every sequence using our classifier, rank each sequence by the predicted risk score, and inspect only the top $k\%$. We evaluate this on an independent fine-tuning run over a disjoint non-deduplicated 1M-sequence sample from the same source and report recall@$k$, the fraction of all memorized sequences contained in the top $k\%$ (Table~\ref{tab:recallk}, Figure~\ref{fig:recallk}).

We find that our classifier recovers memorization far above the random baseline (recall@$k = k$) in nearly every case, and recovers most of it within a small budget on FineWeb, PG-19, and OpenWebMath. We want to highlight that memorization is often rare and that is why a practitioner may want to catch most of it. In this setting, inspecting just the top $0.5$--$1\%$ of sequences already recovers the large majority of memorized examples (for example, $89$--$99\%$ for FineWeb and $74$--$98\%$ for PG-19 across the three models). For WildChat, a small budget recovers less memorization than on other datasets, both because memorization is common ($4\%$) and because memorized and non-memorized sequences are harder to separate (Table~\ref{tab:main_pythia}). As a result, it requires a larger budget ($56\%$ recall at $k = 20\%$, Figure~\ref{fig:recallk}). Overall, our findings show that a practitioner can flag most memorization by inspecting a small \& prioritized fraction of the training data. This pattern holds across architectures, as Qwen-2.5 1.5B and Llama-3.2 1B show the same behavior (Table~\ref{tab:recallk}), with their recall-vs-budget curves in Figure~\ref{fig:recallk}. 

As a complement to recall@$k$, we report precision@$k$ (the fraction of the inspected top $k\%$ that is truly memorized) in Appendix~\ref{app:precisionk}. At tight audit budgets the classifier runs far above the random baseline, its natural memorization rate (for example, FineWeb precision@$0.1\%$ is about $380\times$ its $0.015\%$ rate, and for Qwen-2.5 1.5B the top $1\%$ of The Stack and WildChat is $42$ to $59\%$ truly memorized). The precision is low at larger budgets on datasets with lower memorization rates as nearly all memorized sequences are already found within a small budget and inspecting further only adds non-memorized ones. 

\subsubsection{Removing predicted sequences reduces memorization}
\label{subsubsec:removing_predicted_sequences_reduces_mem}
Next, we test whether our classifier can help reduce memorization in practice. We rank all sequences in the fine-tuning data according to their predicted risk, remove the top $k\%$ ($k = 1$, 5, or 10), and fine-tune Qwen-2.5 1.5B again on the remaining data. To ensure that the reduction in memorization is not simply due to training on less data, we also remove a random 10\% of the data and fine-tune again. Table~\ref{tab:loop_all} shows the results. Removing the top 10\% lowers memorization by 41--50\% on The Stack and WildChat. For example, on The Stack it goes from 3.72\% to 1.87\%. On OpenWebMath, FineWeb, and PG-19, it removes almost all memorization. In contrast, removing a random 10\% changes memorization by 7\% or less. This shows that the reduction in memorization comes from removing the sequences flagged by our classifier, and not from training on less data.

We also check whether removing these sequences hurts model utility. We measure perplexity on held-out set of 10{,}000 sequences from the same domain and 0-shot accuracy on five common benchmarks: LAMBADA~\citep{paperno-etal-2016-lambada}, WinoGrande~\citep{winogrande}, HellaSwag~\citep{zellers-etal-2019-hellaswag}, PIQA~\citep{bisk2019piqareasoningphysicalcommonsense}, and ARC-Easy~\citep{clark2018thinksolvedquestionanswering}, which we run with the \texttt{lm-evaluation-harness}~\citep{eval-harness}. Compared with fine-tuning on the full data, removing the top 10\% raises the in-domain perplexity by at most 1.6\% and changes the accuracy on each benchmark by at most 1.7 points (Table~\ref{tab:loop_utility}). Removing only the top 1\% or 5\% has a similarly small effect (Appendix~\ref{app:loop}, Table~\ref{tab:loop_utility_k}). Unlike training-time methods such as goldfish loss~\citep{hans2024likegoldfishdontmemorize}, which change the loss on all of the training data, our approach leaves the training procedure unchanged and only removes a small set of sequences flagged before fine-tuning. Each setting uses a single fine-tuning run. For The Stack, we also repeat the experiment over three seeds and see the same result (Appendix~\ref{app:loop}).

\begin{table}[t]
\centering\small
\setlength{\tabcolsep}{3.5pt}
\resizebox{\columnwidth}{!}{\begin{tabular}{@{}lrrrrr@{}}
\toprule
Dataset & Full & Rand.\ 10\% & Top 1\% & Top 5\% & Top 10\% \\
\midrule
The Stack   & 3.72  & 3.65  & 3.04  & 2.02 & 1.87 \\
OpenWebMath & 0.86  & 0.85  & 0.45  & 0.15 & 0.01 \\
FineWeb     & 0.039 & 0.040 & 0     & 0    & 0    \\
PG-19       & 0.072 & 0.067 & 0.003 & 0    & 0    \\
WildChat    & 4.87  & 4.54  & 4.32  & 3.55 & 2.90 \\
\bottomrule
\end{tabular}}
\caption{Memorization rate (\%) of Qwen-2.5 1.5B after fine-tuning on the full data, on the data with a random 10\% removed, and on the data with the top $k\%$ of sequences ranked by our classifier removed.}
\label{tab:loop_all}
\end{table}

\subsection{Results by Feature Type}
\label{subsec:feature_type}

\begin{figure*}[t]
\centering
\includegraphics[width=0.6\textwidth]{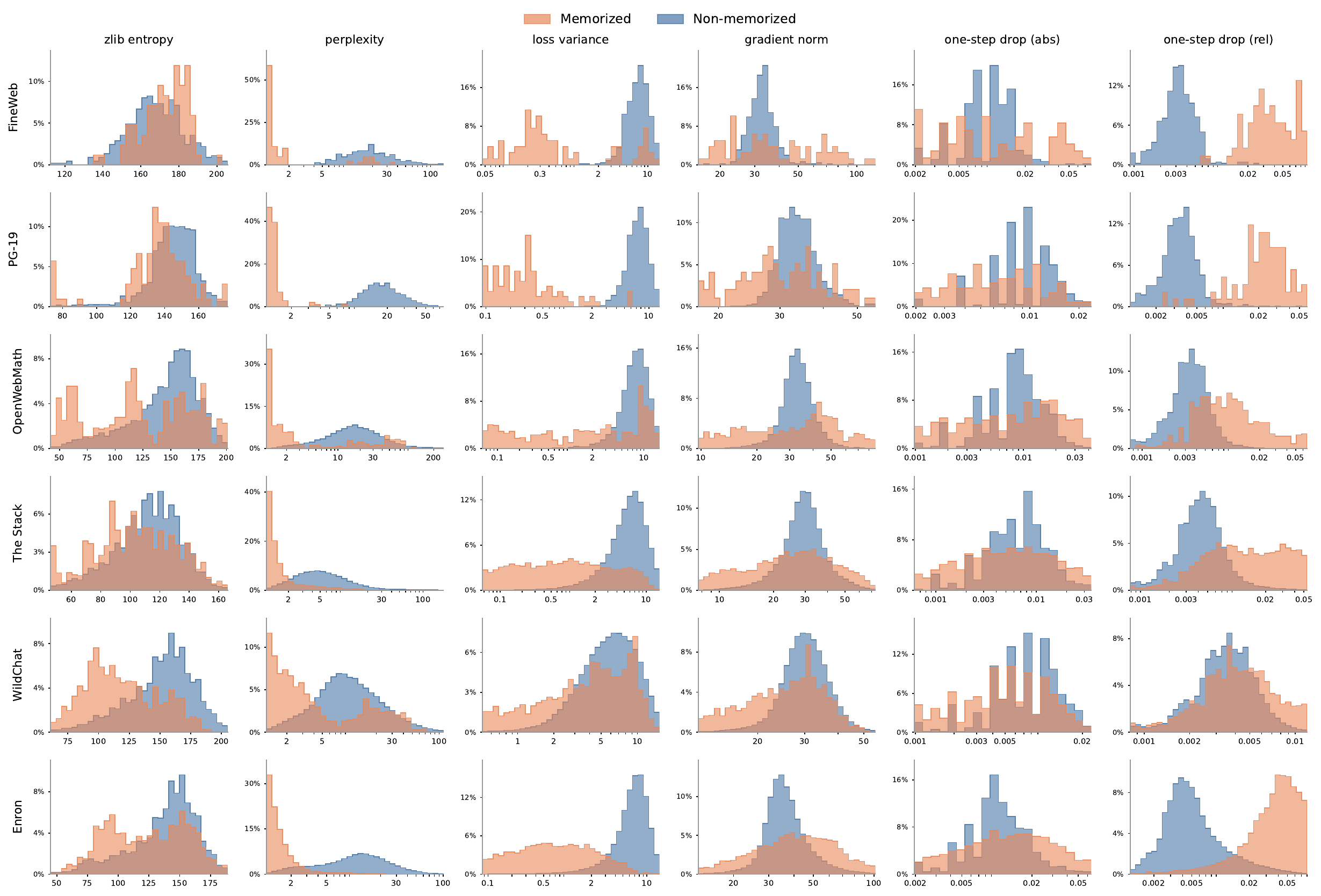}
\caption{Distribution of each of our six features for memorized versus non-memorized sequences under Pythia-1.4B across all six datasets. Each class is normalized so its bars sum to $100\%$, in the deduplicated, contamination-removed setup. All axes except zlib entropy are log-scaled. On average, memorized sequences are more compressible (lower zlib entropy), less surprising (lower perplexity), more uniform (lower loss variance), and faster to learn (larger one-step relative loss drop). The same distributions for Qwen-2.5 1.5B and Llama-3.2 1B are in Appendix~\ref{app:feature_dists}.}
\label{fig:feat_pythia}
\end{figure*}

We now break these results down by feature type (Table~\ref{tab:main_pythia}). Figure~\ref{fig:feat_pythia} shows how each of our six features are distributed for memorized versus non-memorized sequences for Pythia-1.4B, with the same distributions for Qwen-2.5 1.5B and Llama-3.2 1B in Appendix~\ref{app:feature_dists} (Figure~\ref{fig:feat_qwenllama}). Text-based feature (zlib entropy) appears to be the weakest one. For Pythia-1.4B, it is close to chance on FineWeb and PG-19 (AUC around 0.6) and only moderate elsewhere, reaching its highest value on OpenWebMath (0.738) and WildChat (0.777). Across all three models, it ranges from 0.46 to 0.78. A plausible explanation is that in these datasets memorized sequences are more compressible than non-memorized ones, for instance because they contain more templated or formulaic text~\citep{prashanth2025recite}. This makes them separable by compressibility alone, though text features remain substantially weaker than the model-based features.

In contrast, loss-based and gradient-based features computed from the base model are both strong. These findings also hold for Qwen-2.5 1.5B and Llama-3.2 1B (Table~\ref{tab:main_qwen_llama}). Most notably, gradient-based features alone are comparable to loss-based features and nearly match the full feature set (for example, above 0.99 on FineWeb and PG-19). We examine why they work through the lens of \emph{learnability} in Section~\ref{sec:learnability}.

\section{Memorization as Learnability}
\label{sec:learnability}

In this section, we study \emph{why} sequences get memorized, using the one-step loss change features from Section~\ref{sec:exp_setup}. 

Prior work has primarily relied on the perplexity of a sequence as an indicator of memorization~\citep{secret-sharer, gpt2_extraction, prashanth2025recite}, where the model finds memorized sequences less surprising. While predictive, perplexity offers limited insight into the role of learning behind memorization and \emph{why} certain sequences are memorized over others. Gradient-based signals have been used to attribute model predictions~\citep{axiomatic, du-etal-2023-generalizing} and to identify examples that are important during training~\citep{data_diet}. \citet{stoehr2024localizingparagraphmemorizationlanguage} show that gradients carry a memorization signal in LLMs, which they use to localize stored memorization post-hoc. Other works study how individual examples are learned during fine-tuning. \citet{swayamdipta-etal-2020-dataset} track each example over training and find that some examples are learned much faster than others, and \citet{ren2025learning} study how a single gradient step on one example changes an LLM's predictions during fine-tuning. However, these analyses are all computed \emph{during} or \emph{after} training on the sequences in question. We instead ask which signals from the \emph{base} model predict memorization before fine-tuning, and what they reveal about the mechanism behind memorization.

\begin{figure*}[t]
\centering
\includegraphics[width=0.8\textwidth]{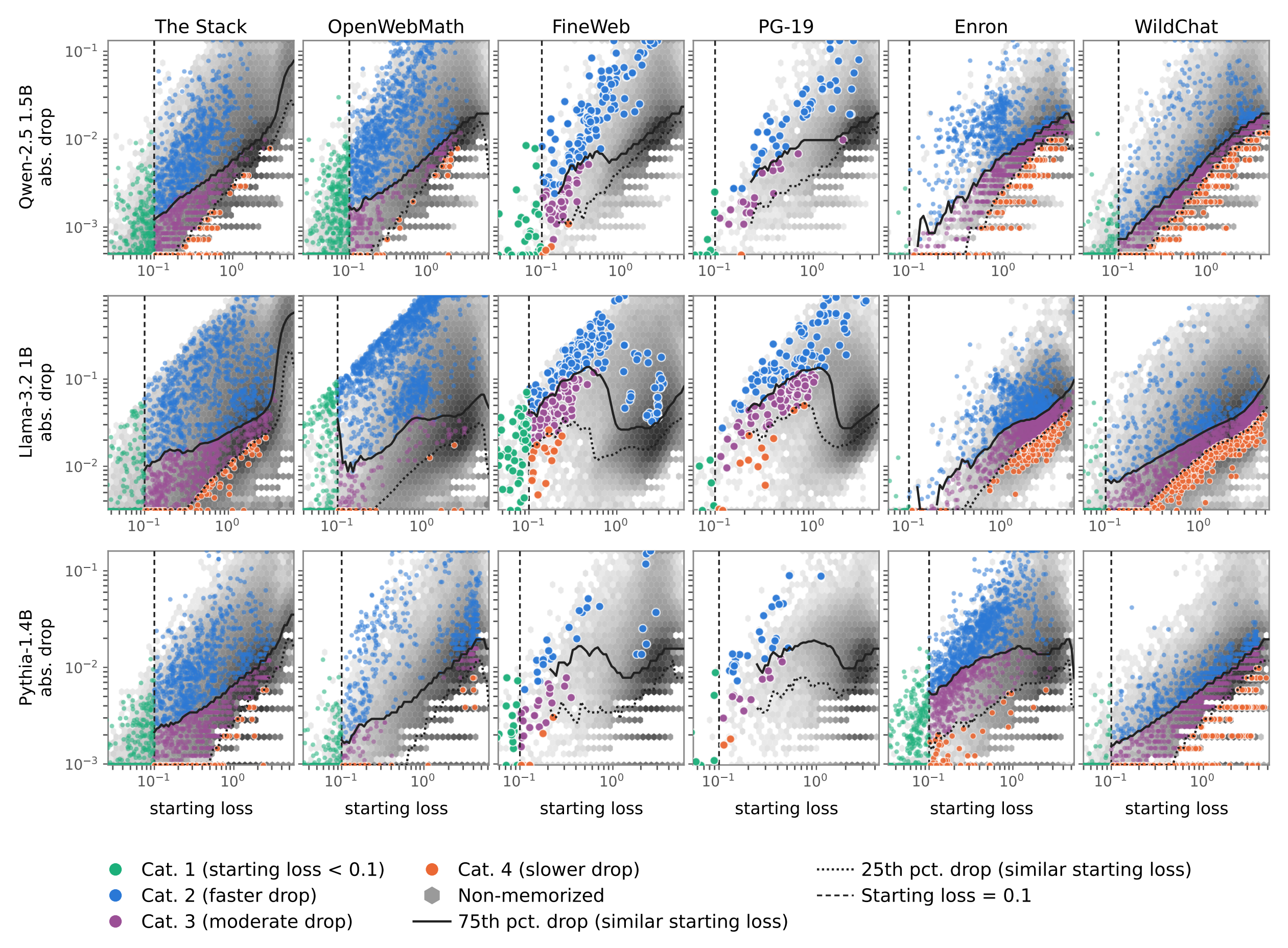}
\caption{Starting loss under the base model vs. the absolute one-step loss drop for each model (rows) \& dataset (columns). The grey shading shows the density of non-memorized sequences. The solid \& dotted black lines show the 75th and 25th percentiles of the absolute one-step loss drop among sequences with a similar starting loss. Memorized sequences are colored by their learning category, with \low{\textbf{Category 1}} having a starting loss below 0.1 (green). Among sequences with a starting loss of at least 0.1, \fast{\textbf{Category 2}} has a drop above the 75th percentile (blue), \moderate{\textbf{Category 3}} has a drop between the 25th and 75th percentiles (purple), \& \slow{\textbf{Category 4}} has a drop below the 25th percentile (orange).}
\label{fig:learnability_matched}
\end{figure*}

Motivated by the finding that memorized sequences are those a model fits most easily~\citep{borkar2026memorizationdynamicsknowledgedistillation}, we hypothesize that memorized sequences are \emph{easier to learn} than non-memorized ones, and that this is detectable from a single gradient step on the base model. We measure this via the one-step loss change, specifically, how much the loss on a sequence $x$ drops after taking a single gradient step on $x$. We find that sequences follow different routes to getting memorized depending on how well the base model already fits them and how strongly they are learned after a single gradient update.

\subsection{Not All Sequences Are Memorized the Same Way}
\label{subsec:not_all_seq_are_memorized_the_same_way} 

\begin{table}[!t]
\centering\small
\setlength{\tabcolsep}{2pt}
\begin{tabular}{@{}lcccc@{}}
\toprule
 & Low & Faster & Moderate & Slower \\
Dataset & start & drop & drop & drop \\
\midrule
The Stack   & 25\hspace{0.1em}/\hspace{0.1em}11\hspace{0.1em}/\hspace{0.1em}18 & 42\hspace{0.1em}/\hspace{0.1em}53\hspace{0.1em}/\hspace{0.1em}44 & 23\hspace{0.1em}/\hspace{0.1em}31\hspace{0.1em}/\hspace{0.1em}29 & 10\hspace{0.1em}/\hspace{0.1em}5\hspace{0.1em}/\hspace{0.1em}9 \\
OpenWebMath & 27\hspace{0.1em}/\hspace{0.1em}13\hspace{0.1em}/\hspace{0.1em}20 & 57\hspace{0.1em}/\hspace{0.1em}78\hspace{0.1em}/\hspace{0.1em}61 & 13\hspace{0.1em}/\hspace{0.1em}8\hspace{0.1em}/\hspace{0.1em}16 & 3\hspace{0.1em}/\hspace{0.1em}1\hspace{0.1em}/\hspace{0.1em}3 \\
FineWeb     & 16\hspace{0.1em}/\hspace{0.1em}16\hspace{0.1em}/\hspace{0.1em}23 & 60\hspace{0.1em}/\hspace{0.1em}50\hspace{0.1em}/\hspace{0.1em}39 & 21\hspace{0.1em}/\hspace{0.1em}29\hspace{0.1em}/\hspace{0.1em}31 & 3\hspace{0.1em}/\hspace{0.1em}5\hspace{0.1em}/\hspace{0.1em}7 \\
PG-19       & 12\hspace{0.1em}/\hspace{0.1em}3\hspace{0.1em}/\hspace{0.1em}14 & 63\hspace{0.1em}/\hspace{0.1em}52\hspace{0.1em}/\hspace{0.1em}56 & 24\hspace{0.1em}/\hspace{0.1em}38\hspace{0.1em}/\hspace{0.1em}25 & 1\hspace{0.1em}/\hspace{0.1em}7\hspace{0.1em}/\hspace{0.1em}5 \\
Enron       & 1\hspace{0.1em}/\hspace{0.1em}1\hspace{0.1em}/\hspace{0.1em}16 & 38\hspace{0.1em}/\hspace{0.1em}43\hspace{0.1em}/\hspace{0.1em}57 & 42\hspace{0.1em}/\hspace{0.1em}43\hspace{0.1em}/\hspace{0.1em}23 & 19\hspace{0.1em}/\hspace{0.1em}13\hspace{0.1em}/\hspace{0.1em}4 \\
WildChat    & 11\hspace{0.1em}/\hspace{0.1em}4\hspace{0.1em}/\hspace{0.1em}6 & 27\hspace{0.1em}/\hspace{0.1em}25\hspace{0.1em}/\hspace{0.1em}28 & 38\hspace{0.1em}/\hspace{0.1em}45\hspace{0.1em}/\hspace{0.1em}45 & 24\hspace{0.1em}/\hspace{0.1em}26\hspace{0.1em}/\hspace{0.1em}21 \\
\bottomrule
\end{tabular}
\caption{Share of memorized sequences (\%) in each learning category for Qwen-2.5 1.5B / Llama-3.2 1B / Pythia-1.4B.}
\label{tab:three_routes}
\end{table}

We find that there are four different ways a sequence may get learned and eventually memorized during fine-tuning. Below, we first describe how sequences are learned, and then what shapes which of them get memorized verbatim.

\paragraph{How sequences are learned} We place every sequence in the fine-tuning data into one of four categories. \low{\textbf{(1) Low starting loss.}} A sequence may already have a low loss under the base model, which means the base model already finds it easy to fit. We call a starting loss low if the base model's average per-token loss on the sequence is below 0.1. We choose this cutoff because below it the base model already nearly reproduces many of these sequences, with a BLEU~\citep{papineni-etal-2002-bleu} of at least 0.75~\citep{ippolito-etal-2023-preventing}, even though we remove any verbatim completions (Section~\ref{sec:exp_setup}). \fast{\textbf{(2) Higher starting loss and a faster drop.}} A sequence may start with a higher loss, but its loss drops more than usual after a single gradient step. \moderate{\textbf{(3) Higher starting loss and a moderate drop.}} A sequence may start with a higher loss and its loss drops about as much as usual. \slow{\textbf{(4) Higher starting loss and a slower drop.}} A sequence may also start with a higher loss, but its loss drops less than usual after that step.

To obtain the last three categories, we exclude the first category and only consider sequences with a starting loss of at least 0.1. We then compare each sequence only with sequences that have roughly the same starting loss (within 10\%). For example, a sequence with a starting loss of 0.34 is compared with all sequences whose starting loss is between 0.31 and 0.37. If less than 50 sequences fall in this range, we instead use the 50 sequences with closest starting loss. For each sequence, we compute its absolute one-step loss drop, which measures how much its loss falls after one gradient step on it (Section~\ref{sec:exp_setup}). We then rank this drop against the drops of the sequences it is compared with to get a percentile. We say that a sequence has a faster drop if its percentile is above 75, a slower drop if it is below 25, and a moderate drop otherwise. 

Table~\ref{tab:three_routes} shows how memorized sequences are distributed across the four categories. Figure~\ref{fig:learnability_matched} shows these categories visually and Appendix~\ref{app:route_examples} shows randomly chosen examples of memorized sequences from each category. In The Stack, OpenWebMath, FineWeb, and PG-19, a large portion of memorized sequences has a \fast{\textbf{faster}} drop (39--78\%), and only a small portion has a \slow{\textbf{slower}} drop (1--10\%). \low{\textbf{Category 1}} makes up 1--27\% and \moderate{\textbf{Category 3}} accounts for 8--45\%. WildChat is the exception, where most memorized sequences have a \moderate{\textbf{moderate}} drop (38--45\%), and the rest are split about evenly between a \fast{\textbf{faster}} drop (25--28\%) and a \slow{\textbf{slower}} drop (21--26\%). This may explain why memorization is harder to predict in WildChat than in other domains (Table~\ref{tab:main_pythia}). Sequences in \slow{\textbf{Category 4}} have a smaller one-step drop than most sequences with a similar starting loss, so our features find them harder to tell apart from non-memorized sequences. In contrast, our features find \low{\textbf{Category 1}} and \fast{\textbf{Category 2}} sequences easier to catch. Memorized sequences in \fast{\textbf{Category 2}} typically have one of the largest relative loss drops in the data. Memorized sequences in \low{\textbf{Category 1}} are caught by their low starting loss and, in most settings, also by their relative loss drop. Their loss drops only a little in absolute terms, but since their starting loss is already very low, this small drop is still a large fraction of it. 

\paragraph{Overlap with pretraining data} Next, we try to understand why some sequences naturally have a low starting loss (\low{\textbf{Category 1}}), and why others are learned fast (\fast{\textbf{Category 2}}), moderately (\moderate{\textbf{Category 3}}), or slowly (\slow{\textbf{Category 4}}). Since both the starting loss and the one-step loss drop are computed under the base model, we hypothesize that these categories can be better understood by looking at the data the base model was pretrained on. This is in line with prior work that looks into the training data to trace model behavior~\citep{struggle, elazar2024whats, infinigram}. Among the base model families we use in this work, only Pythia has public training data, the Pile~\citep{gao2020pile800gbdatasetdiverse}. So we take the sequences whose starting loss and one-step drop are computed under the Pythia-1.4B base model. 

For each sequence, we count how many times the 8-grams in its 50-token suffix also occur in the Pile, and add these counts up. Table~\ref{tab:pile_overlap} reports the median of this count for each category. We find that \low{\textbf{Category 1}} sequences have the most overlap, which is consistent with the base model already finding them easy to fit. \fast{\textbf{Category 2}} sequences have less overlap than \low{\textbf{Category 1}}, which is consistent with their higher starting loss. But their overlap is still higher than that of \moderate{\textbf{Category 3}} and \slow{\textbf{Category 4}}, which is consistent with their faster drop in loss. \moderate{\textbf{Category 3}} sequences also have more overlap than \slow{\textbf{Category 4}}. The only exception is PG-19, where \fast{\textbf{Categories 2}}, \moderate{\textbf{3}}, and \slow{\textbf{4}} have about the same overlap, possibly because PG-19 is itself part of the Pile. These results suggest that how much a sequence overlaps with the pretraining data shapes how it is learned during fine-tuning.

\begin{table}[t]
\centering\small
\setlength{\tabcolsep}{4pt}
\begin{tabular}{@{}lrrrr@{}}
\toprule
Dataset & \low{Cat. 1} & \fast{Cat. 2} & \moderate{Cat. 3} & \slow{Cat. 4} \\
\midrule
The Stack   & 97.5k & 637  & 223 & 114 \\
OpenWebMath & 1.8k  & 19   & 13  & 10  \\
FineWeb     & 40.8k & 6    & 4   & 3   \\
PG-19       & 35.0k & 116  & 113 & 113 \\
Enron       & 75.7k & 1.3k & 595 & 467 \\
WildChat    & 7.2k  & 16   & 10  & 8   \\
\bottomrule
\end{tabular}
\caption{Overlap with the pretraining data (the Pile) for Pythia-1.4B. For each sequence, we count how many times the 8-grams in its 50-token suffix also occur in the Pile, and we report the median of this count over all sequences in each category.}
\label{tab:pile_overlap}
\end{table}

\paragraph{What shapes memorization} After studying how sequences are learned under the base model, we next look at what shapes their verbatim memorization during fine-tuning. To do this, we compare memorized and non-memorized sequences that are learned in a similar way. For each memorized sequence, we find non-memorized sequences from the same category that have nearly the same starting loss and one-step loss drop, both within about 10\% of its own. If fewer than five such sequences exist, we use the five closest ones from the same category. For each sequence, we then count how many times the 8-grams in its 50-token suffix occur elsewhere in the fine-tuning data, and add these counts up. Table~\ref{tab:ft_overlap} reports the median of this count for Pythia-1.4B. 

We find that memorized sequences have a much higher overlap with the rest of the fine-tuning data than non-memorized sequences that are learned in the same way. For instance, in The Stack, the median in \low{\textbf{Category 1}} is 6.0k for memorized sequences and 3.1k for non-memorized ones. In OpenWebMath, the median in \fast{\textbf{Category 2}} is 4.7k for memorized sequences and 35 for non-memorized ones. In WildChat, it is 5.1k vs.\ 105 in \moderate{\textbf{Category 3}} and 5.6k vs.\ 84 in \slow{\textbf{Category 4}}. We see the same results for Qwen-2.5 1.5B and Llama-3.2 1B (Appendix~\ref{app:ft_overlap}). The only exception is Enron for Pythia-1.4B, where non-memorized sequences have a higher 8-gram overlap. We believe this is because many Enron emails share the same templates (e.g., headers, disclaimers, and forwarded text), so short 8-grams may overlap heavily with the rest of the data for both memorized and non-memorized sequences. When we instead count overlapping 32-grams, we find that memorized sequences in Enron again have a higher overlap in all four categories (Table~\ref{tab:ft32} in Appendix~\ref{app:ft_overlap}). Together, these results suggest that how a sequence is learned is shaped by its overlap with the pretraining data, while whether it ends up memorized is shaped by how much of it is repeated in the fine-tuning data. This is in line with prior work, which finds that overlapping n-grams in the training data affect memorization~\citep{borkar-etal-2025-privacy}, and that in supervised fine-tuning, repetition alone is not enough to explain which secrets get memorized~\citep{schwarzschild2026unmaskingsecretcodeusing}.

\begin{table}[t]
\centering\small
\setlength{\tabcolsep}{2pt}
\resizebox{\columnwidth}{!}{\begin{tabular}{@{}lrrrr@{}}
\toprule
Dataset & \low{Cat. 1} & \fast{Cat. 2} & \moderate{Cat. 3} & \slow{Cat. 4} \\
\midrule
The Stack   & 6.0k\hspace{0.1em}/\hspace{0.1em}3.1k & 7.3k\hspace{0.1em}/\hspace{0.1em}2.6k & 13.0k\hspace{0.1em}/\hspace{0.1em}298 & 9.7k\hspace{0.1em}/\hspace{0.1em}229 \\
OpenWebMath & 199\hspace{0.1em}/\hspace{0.1em}148 & 4.7k\hspace{0.1em}/\hspace{0.1em}35 & 3.3k\hspace{0.1em}/\hspace{0.1em}21 & 1.5k\hspace{0.1em}/\hspace{0.1em}38 \\
FineWeb     & 392\hspace{0.1em}/\hspace{0.1em}27 & 539\hspace{0.1em}/\hspace{0.1em}34 & 72\hspace{0.1em}/\hspace{0.1em}26 & 43\hspace{0.1em}/\hspace{0.1em}5 \\
PG-19       & 924\hspace{0.1em}/\hspace{0.1em}36 & 1.4k\hspace{0.1em}/\hspace{0.1em}23 & 51\hspace{0.1em}/\hspace{0.1em}9 & 708\hspace{0.1em}/\hspace{0.1em}1 \\
Enron       & 19.4k\hspace{0.1em}/\hspace{0.1em}17.4k & 3.2k\hspace{0.1em}/\hspace{0.1em}5.1k & 7.2k\hspace{0.1em}/\hspace{0.1em}14.2k & 10.0k\hspace{0.1em}/\hspace{0.1em}17.8k \\
WildChat    & 750\hspace{0.1em}/\hspace{0.1em}265 & 4.5k\hspace{0.1em}/\hspace{0.1em}106 & 5.1k\hspace{0.1em}/\hspace{0.1em}105 & 5.6k\hspace{0.1em}/\hspace{0.1em}84 \\
\bottomrule
\end{tabular}}
\caption{Overlap with the fine-tuning data for Pythia-1.4B, shown as memorized\hspace{0.1em}/\hspace{0.1em}non-memorized. For each sequence, we count how many times the 8-grams in its 50-token suffix occur elsewhere in the fine-tuning data. We report the median count. Appendix~\ref{app:ft_overlap_details} gives more details on how we compute this count.}
\label{tab:ft_overlap}
\end{table}

\section{Generalization Across Domains, Models, and Scale}
\label{sec:generalization}

\begin{figure*}[t]
\centering
\includegraphics[width=0.8\linewidth]{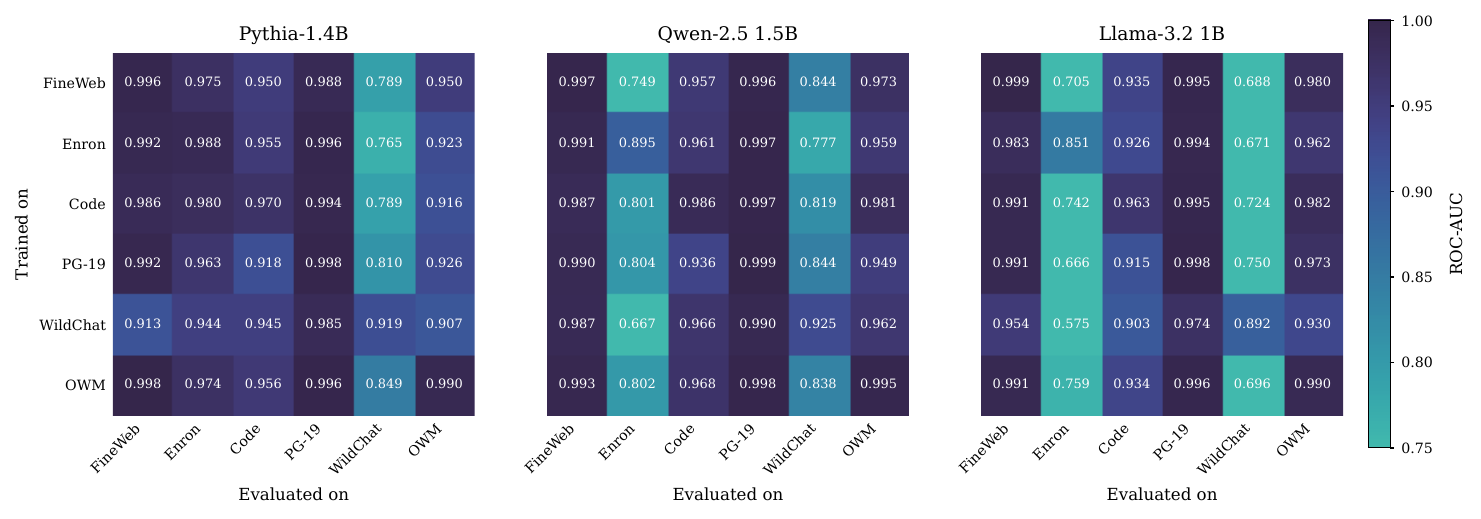}
\caption{Cross-domain transfer for Pythia-1.4B, Qwen-2.5 1.5B, and Llama-3.2 1B (deduplicated setup). Each cell is the ROC-AUC of a classifier trained on the row domain and evaluated on the column domain. The diagonal is the in-domain reference from a 70/30 held-out split, and off-diagonal cells are cross-domain transfer. Pythia values are the mean over three fine-tuning seeds, and Qwen and Llama are from a single run.}
\label{fig:cross_domain}
\end{figure*}
\begin{figure}[!t]
\centering\small
\begin{tabular}{lccc}
\toprule
 & \multicolumn{3}{c}{Evaluated on} \\
\cmidrule(lr){2-4}
Trained on & Pythia & Qwen & Llama \\
\midrule
Pythia & 0.978 & 0.934 & 0.906 \\
Qwen   & 0.949 & 0.967 & 0.891 \\
Llama  & 0.943 & 0.921 & 0.949 \\
\bottomrule
\end{tabular}
\captionof{table}{\textbf{Cross-model transfer}, mean ROC-AUC over the six datasets (deduplicated setup, all features). Each cell trains on the row model's features and labels and predicts the column model's memorization. The diagonal is the in-domain reference (70/30 held-out split).}
\label{tab:cross_model}
\vspace{10pt}
\centering
\includegraphics[width=\linewidth]{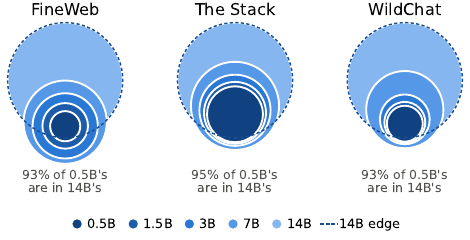}
\captionof{figure}{Memorization for the Qwen-2.5 models from 0.5B to 14B, all fine-tuned on the same data. Circles are proportional to the number of memorized sequences and each circle overlaps the 14B circle (dashed edge) by the measured fraction. The memorization in smaller models falls almost entirely inside that of the 14B.}
\label{fig:mem_venn}
\end{figure}

The results so far train and evaluate a classifier within a single dataset domain and a single model architecture. We now study whether a classifier trained on a particular domain, model family, and model size generalizes to other dataset domains, model families, and model sizes. This would help us understand whether the memorization patterns the classifiers capture are universal or tied to specific datasets, model architectures, or sizes. If they are universal, this could allow a practitioner to train the classifier once and reuse it on new datasets, new base models, or new model sizes. Throughout this section we use the deduplicated setup. We call this setting \emph{transfer}, where we train the classifier on one setting (the \emph{source}) and evaluate it on another (the \emph{target}). For each transfer, we train the classifier on all of the source's balanced sequences (the 1:3 sample from Section~\ref{sec:exp_setup}) and report its ROC-AUC on all of the target's. The diagonal entries in Figure~\ref{fig:cross_domain} and Table~\ref{tab:cross_model}, where the source and target are the same, act as an in-domain reference. Since we cannot evaluate on the same data the classifier was trained on, these entries instead use a 70/30 split for train and test. 

\subsection{Cross-Domain Transfer}
\label{subsec:cross_domain}
We first hold the base model fixed (Pythia-1.4B) and vary the dataset, training on one domain and predicting memorization in each of the others (Figure~\ref{fig:cross_domain}). We observe a strong transfer, where, setting aside the conversational WildChat target, every off-diagonal transfer stays between 0.907 and 0.998 AUC (mean 0.961), close to the in-domain diagonal (mean 0.977). Thus, the memorization signals that the classifier captures appear to be largely domain independent. For instance, a classifier trained on web text (FineWeb) predicts memorization in code and mathematics nearly as well as one trained in-domain. WildChat is the one harder target (0.765 to 0.849 when predicted from other domains), which is consistent with it also being the hardest dataset in-domain (Table~\ref{tab:main_pythia}), as its multi-turn conversational format is the most distinct of the six. In practice, a practitioner can train the classifier on a single labeled domain and use it to flag any potential memorization across new domains. Qwen-2.5 1.5B and Llama-3.2 1B show similar trends with off-diagonal mean AUC of 0.916 and 0.876 respectively (Figure~\ref{fig:cross_domain}).

\subsection{Cross-Model Transfer}
\label{subsec:cross_model}
We next hold the dataset fixed and vary the base model, asking whether a classifier trained on one architecture predicts another architecture's memorization (Table~\ref{tab:cross_model}). The classifier is trained on the source model's pre-fine-tuning features and labels, then applied to the target model's own features and scored against the target model's memorization labels. In other words, we use the feature \emph{weights} trained on the source model and compute the feature \emph{values} using the target base model. Averaged over the six datasets, cross-model transfer reaches 0.924 AUC, only modestly below the in-domain reference of 0.964. This transfer is carried by the model-dependent features. The loss-based and gradient-based features transfer across model families almost as well as they work in-domain. Loss-based features reach an AUC of 0.924 across models, compared with 0.942 in-domain, and gradient-based features reach 0.907, compared with 0.946. In contrast, text-based features are weak everywhere (0.69 to 0.72). Thus, the signals that determine what sequences get memorized are largely shared across architectures rather than specific to any one model. 

\subsection{Cross-Scale Transfer}
\label{subsec:cross_scale} 
\begin{figure*}[t]
\centering
\includegraphics[width=\textwidth]{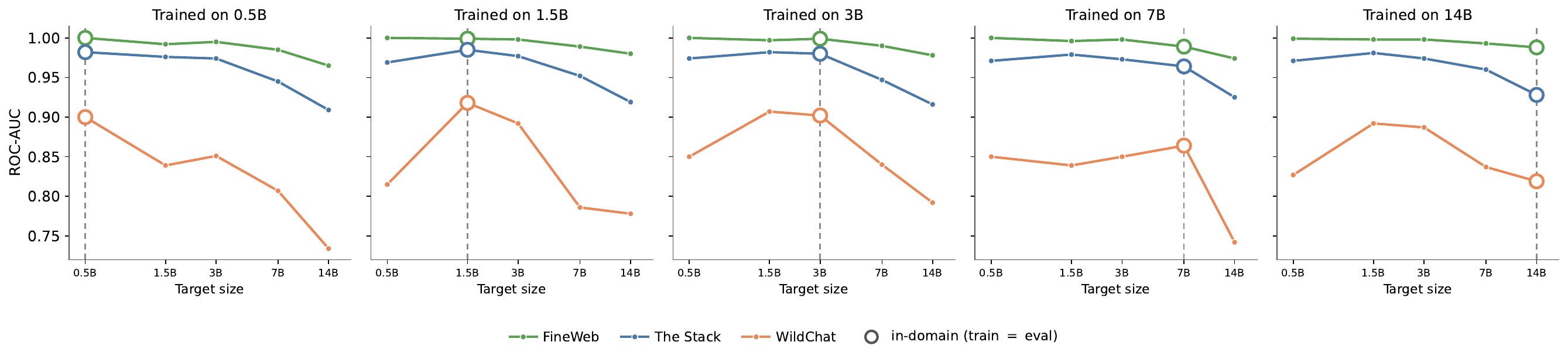}
\caption{Cross-scale transfer across Qwen-2.5 ladder, one panel per source (training) model (deduplicated setup). Within each panel, each line is a dataset and shows the ROC-AUC of that source model's classifier evaluated across all target model sizes on the $x$-axis. The circle and dashed line mark the in-domain reference where the source and target models are the same size, so points to its left predict memorization in a smaller model and points to its right predict memorization in a larger one. Predictability stays high for smaller counterparts but declines for larger ones.}
\label{fig:cross_scale}
\end{figure*}
Next, we study whether a classifier trained on the memorization labels of a smaller model can predict memorization in larger models, and whether a classifier trained on a larger model's labels can find memorization in smaller models. We consider the Qwen-2.5 0.5B, 1.5B, 3B, 7B, and 14B variants, and fine-tune each on The Stack, FineWeb, and WildChat datasets. We fine-tune three seeds for the 0.5B, 1.5B, and 3B models and a single seed for the 7B and 14B models, and we average over seeds. Figure~\ref{fig:cross_scale} reports the results.

We find that a classifier trained on a smaller model can predict memorization in its larger counterparts well on The Stack and FineWeb, and this predictability drops as the size of the larger counterpart grows, most on WildChat. For a classifier trained on 0.5B's labels and predicting memorization in 14B, the AUC still reaches 0.965 on FineWeb, 0.909 on the Stack, and drops to 0.734 on WildChat. 

In the reverse direction, a classifier trained on a larger model predicts its smaller counterparts almost as well as one trained on them directly. Using the 14B model to predict the 0.5B model reaches 0.999 on FineWeb, 0.971 on The Stack, and 0.827 on WildChat. To understand this asymmetry, we compare the sequences memorized by the 0.5B and 14B models, which are fine-tuned on the same data. Almost all sequences memorized by the 0.5B model are also memorized by the 14B model (93 to 95\%, Figure~\ref{fig:mem_venn}), while the 14B model memorizes many sequences that the 0.5B model does not, so the smaller model's memorization is largely a subset of the larger model's~\citep{ borkar2026memorizationdynamicsknowledgedistillation}. The larger model does not transfer better simply because it contains more memorized examples. When we subsample its memorized sequences to the 0.5B count, transfer barely changes (for example, from 0.970 to 0.967 on The Stack). Furthermore, it stays the same even when we remove all sequences that both models memorize from training, so that the classifier never sees any of the 0.5B model's memorized sequences. This suggests that the larger model's memorization spans a broader range of sequences, so a classifier trained on it also recognizes the narrower set that a smaller model memorizes, but not the other way around. Put together, our results show that one can predict which sequences a larger model will memorize after fine-tuning by simply using a classifier trained on a smaller model's memorization labels. This means a single fine-tuning run of a small model may be enough to flag memorization in its larger counterparts.

\section{Scaling Laws for Memorization}
\label{sec:scaling_laws}
In Section~\ref{subsec:cross_scale} we found that memorization predictability carries across model sizes. We now ask how the memorization rate and its predictability change with model size, and whether they follow a scaling law. Prior works on scaling laws have studied the relationship between model's loss and performance with model size~\citep{kaplan2020scalinglawsneurallanguage, brown2020languagemodelsfewshotlearners, hernandez2021scalinglawstransfer, scao2022languagemodeltrainmillion, hoffmann2022trainingcomputeoptimallargelanguage, rae2022scalinglanguagemodelsmethods, chowdhery2022palmscalinglanguagemodeling, hernandez2022scalinglawsinterpretabilitylearning, Bahri_2024}, which have helped inform training-related decisions to produce efficient and better performing language models. 

In this section, we study whether memorization rate and the ability to predict memorization in fine-tuned language models follow a scaling law. \citet{tirumala2022memorization} show that larger models memorize more and faster, but characterize this as a training dynamics rather than a fitted law. \citet{quantifying_memorization} found that memorization rate in pre-trained language models follows a log-linear relationship with model size. \citet{lu-etal-2024-scaling} find that fact memorization in language models grows linearly with model size and follows a negative exponential law with the number of training epochs.  For predicting memorization in pre-trained models, \citet{emergent_memorization} report a non-linear trend for precision and recall while forecasting memorization in a 12B model using the smaller counterparts. 

Despite these important works on scaling laws for studying loss and memorization in pre-trained language models, we find that the literature is relatively poor on studying laws for verbatim memorization in fine-tuned models, which we study. We highlight that \citet{lu-etal-2024-scaling}  study scaling laws for fact memorization, which is different than verbatim memorization of training data sequences. We fine-tune Qwen 2.5 ladder that includes 0.5B, 1.5B, 3B, 7B, and 14B variants on The Stack, WildChat, and FineWeb datasets using the setup in Section~\ref{sec:exp_setup} and report our findings below. 

\begin{figure*}[t]
\centering
\includegraphics[width=0.8\textwidth]{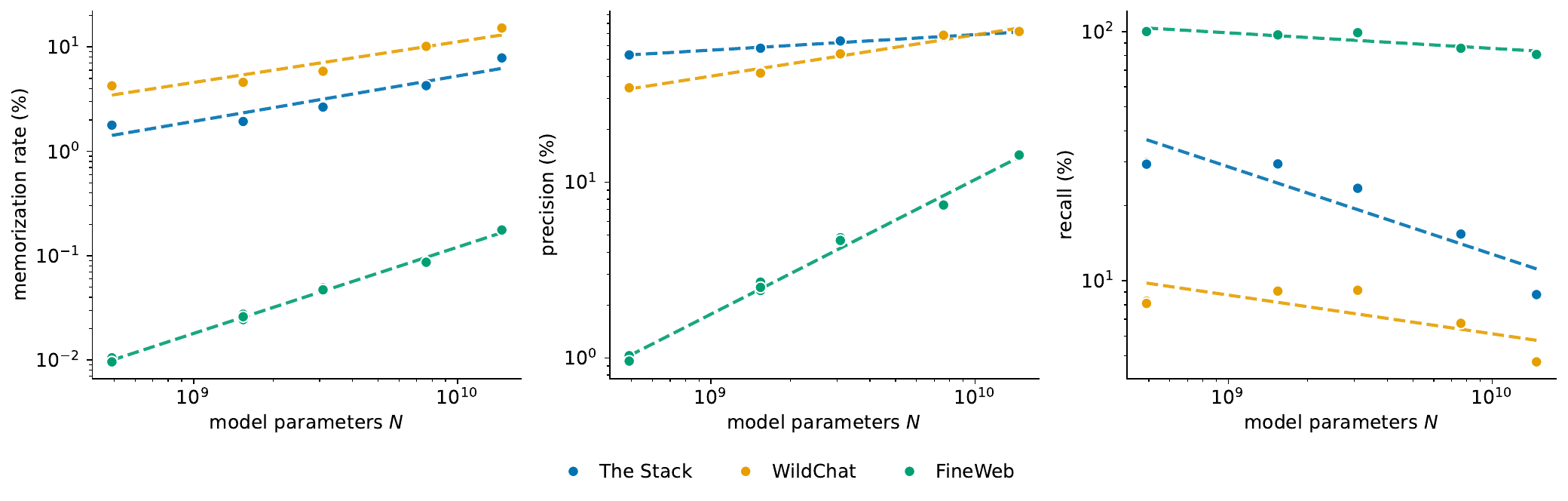}
\caption{Scaling of memorization and its predictability across Qwen-2.5 family (0.5B, 1.5B, 3B, 7B, and 14B), on The Stack, WildChat, and FineWeb datasets. \textbf{Left}: the memorization rate rises as a power law in model size. \textbf{Center}: precision at a $1\%$ audit budget rises as a clean power law. \textbf{Right}: recall at a $1\%$ audit budget declines. Points are individual fine-tuning-seed runs and dashed lines are the fitted power laws.}
\label{fig:scaling}
\end{figure*}

\subsection{Memorization Rate}
We find that memorization rate $m$ follows a power law in model size ($m \propto N^b$), where $N$ is the number of model parameters. Equivalently, the relationship is linear on a log-log plot with slope $b$. We fit on the mean memorization rate per model size using three fine-tuning seeds for the 0.5B, 1.5B, and 3B variants and a single seed for 7B and 14B. To check that the exponent is robust to the choice of seeds, we refit the power law for every combination of seeds across the 0.5B, 1.5B, and 3B models, giving $3\times3\times3=27$ fits per dataset. The fitted exponent $b$ is highly stable across these 27 fits. Its standard deviation is at most $0.012$ (FineWeb), and $\le 0.005$ for The Stack and WildChat. Table~\ref{tab:mem_law} reports the fit for all three datasets, and Figure~\ref{fig:scaling} (left) shows the fitted power laws.

\begin{table}[t]
\centering
\small   
\begin{tabular}{cccc}
\toprule
Dataset & Mem. Law & 95\% CI on $b$ & $R^2$ \\
\midrule
The Stack & $N^{0.43}$ & $[0.15, 0.72]$ & 0.887 \\
WildChat & $N^{0.39}$ & $[0.15, 0.63]$ & 0.896 \\
FineWeb & $N^{0.83}$ & $[0.74, 0.92]$ & 0.997 \\
\bottomrule
\end{tabular}
\caption{Memorization rate follows a power law in model size ($m\propto N^{b}$) over the five Qwen-2.5 sizes (0.5--14B). We report the exponent $b$, its 95\% confidence interval, and the power-law $R^2$.}
\label{tab:mem_law}
\end{table}

Across all three datasets the exponent is positive and below one, with $95\%$ confidence intervals excluding both $0$ and $1$. So memorization increases with model size but grows more slowly than the number of parameters. FineWeb has the steepest exponent ($N^{0.83}$), while The Stack and WildChat grow more slowly ($N^{0.43}$ and $N^{0.39}$). This differs from the log-linear relationship reported for memorization in pre-trained models~\citep{quantifying_memorization}. In our fine-tuning setup, the power law fits better than log-linear form on every dataset, with $R^2$ of $0.89$, $0.90$, and $0.997$ for the power law against $0.76$, $0.82$, and $0.83$ for the log-linear fit. Our best estimate is that each tenfold increase in model size multiplies the memorization rate by about $2.7\times$ on The Stack ($10^{0.43}$), $2.5\times$ on WildChat ($10^{0.39}$), and $6.8\times$ on FineWeb ($10^{0.83}$).

Following best practices for reporting scaling laws~\citep{li2025misfitting}, we report a $95\%$ confidence interval on each exponent. Because we use only five model sizes for the fit, the exponent is estimated with limited precision. Adding more larger models would tighten these intervals and test whether the sublinear trend continues at larger scale. We describe the full fitting and confidence interval computation procedure in Appendix~\ref{app:scaling_fit}.

\subsection{Predictive Power} 
We next ask how our ability to predict memorization changes as the model size increases. We study the classifier's precision@$k$ and recall@$k$ at a fixed audit budget of $k=1\%$. Put simply, we inspect the top $1\%$ of sequences ranked by predicted risk. We observe that precision@$1\%$ follows a clean power law in model size (Figure~\ref{fig:scaling}, center), while recall@$1\%$ decreases with scale but is not well described by a power law (Figure~\ref{fig:scaling}, right). The two metrics also move in opposite directions (Table~\ref{tab:pred_law}). The underlying recall@$k$ and precision@$k$ values for every model size and at all audit budgets are reported in Tables~\ref{tab:ladder_recallk} and~\ref{tab:ladder_precisionk}.

\begin{table}[t]
\centering
\small
\resizebox{\columnwidth}{!}{%
\begin{tabular}{llccc}
\toprule
Dataset & Metric & Law & 95\% CI & $R^2$ \\
\midrule
The Stack & recall@1\%    & $N^{-0.35}$ & $[-0.64, -0.06]$ & 0.828 \\
          & precision@1\% & $N^{+0.09}$ & $[+0.07, +0.10]$ & 0.989 \\
\midrule
WildChat  & recall@1\%    & $N^{-0.16}$ & $[-0.41, +0.10]$ & 0.559 \\
          & precision@1\% & $N^{+0.23}$ & $[+0.17, +0.30]$ & 0.975 \\
\midrule
FineWeb   & recall@1\%    & $N^{-0.06}$ & $[-0.13, +0.00]$ & 0.769 \\
          & precision@1\% & $N^{+0.77}$ & $[+0.64, +0.89]$ & 0.992 \\
\bottomrule
\end{tabular}%
}
\caption{Scaling of predictive power with model size at a $1\%$ audit budget. For each dataset we fit power laws $\text{metric}\propto N^{b}$ for recall@$1\%$ and precision@$1\%$ over the five Qwen-2.5 sizes, and report the exponent $b$, its 95\% confidence interval, and the power-law $R^2$. Precision rises with scale (positive $b$), while recall at a fixed budget falls (negative $b$).}
\label{tab:pred_law}
\end{table}

Precision@$1\%$ increases with model size on every dataset and the fit is very clean ($R^2 \ge 0.97$). Recall@$1\%$ decreases with model size, but this trend is noisy and not well described by a power law ($R^2$ between $0.56$ and $0.83$). For WildChat and FineWeb the interval includes zero, so the decline is clearest on The Stack. This non-linear behavior of recall is consistent with the non-linear precision and recall trends that \citet{emergent_memorization} report when forecasting memorization.

Both trends follow directly from the memorization rate rising with scale (Table~\ref{tab:mem_law}). As memorization increases with model size, the top $1\%$ of sequences flagged by the classifier contains a larger fraction of memorized sequences, which leads to precision goes up. At the same time, the fixed $1\%$ budget can hold only so many sequences. As a result, it captures a smaller fraction of the growing memorized set that results in recall going down. We report these exponents across a range of audit budgets in Appendix~\ref{app:budget_sweep}.

The classifier's ability to separate memorized from non-memorized sequences in fact stays strong across the ladder, although it declines for the largest models. The cross-validation ROC-AUC (all features) remains high at every size, between $0.81$ and $1.0$ across datasets (for example, $0.91$ to $0.97$ on The Stack and $0.99$ to $1.0$ on FineWeb). Table~\ref{tab:ladder_auc} in Appendix~\ref{app:ladder_auc} reports the AUC for every model size and dataset.

\section{Conclusion}
We show that memorization of a fine-tuning dataset can be predicted before training a model on it, using only features from the base model, memorization labels from a single fine-tuning run, and the target dataset. We find that memorization can be understood through the lens of \emph{learnability}. Some sequences are already nearly learned by the base model, while others start at a higher loss and are learned quickly, moderately, or slowly, as measured by a single gradient step on the base model. A classifier trained on one domain, model family, and model size also predicts memorization in other domains, model families, and model sizes. Finally, we observe that the memorization rate in a fine-tuned language model and precision@$k$ of predicting memorization follow a power law in model size.

\section*{Limitations}
First, we focus specifically on memorization prediction for a fine-tuning language modeling task, and do not cover other forms of fine-tuning such as instruction tuning or reinforcement learning from human feedback (RLHF)~\citep{NEURIPS2022_b1efde53}. We study fine-tuning rather than pretraining because of the high compute cost of pretraining at scale. Understanding how our results translate to other forms of fine-tuning and pretraining is a promising direction for future work. Second, we adopt a specific definition of memorization, namely an exact match of the 50-token suffix generated by greedy decoding from a 50-token prefix. While our findings transfer to looser forms of memorization, such as approximate memorization~\citep{ippolito-etal-2023-preventing}, extending the prediction question to other non-verbatim categories of memorization would be an interesting follow-up. 

Third, for our scaling analysis we experiment with models in the 0.5B to 14B range. Covering additional larger models could improve the robustness of these scaling laws. And while we cover six domains, results may differ for domains not studied here. Finally, our recall@$k$ results are computed on a 1M-sequence sample. Although this is a substantial size, real-world or production datasets that are considerably larger may yield different recall@$k$ values than those we report.

\section*{Ethical Considerations}
We use only publicly available models and datasets. Some of our fine-tuning datasets such as Enron and WildChat may contain sensitive or personally identifiable information. However, these datasets are already public, and our objective is to predict which sequences a model would memorize before fine-tuning begins. Our method is intended as a preventive measure. By flagging the sequences most at risk of memorization before training begins, practitioners can audit or remove high-risk content. 

\section*{Reproducibility Statement}
All datasets and base models we use are publicly available, and we report the key training and evaluation hyperparameters in the paper. To support reproducibility, we release our code, our fine-tuned Pythia, Qwen-2.5, and Llama-3.2 models, the trained memorization classifiers, and the base-model features and memorization labels used to train them. 

\section*{AI Usage Statement}
We acknowledge the use of Generative AI as coding assistant and have verified all resulting code for correctness and quality. We also used Generative AI to polish the writing of this paper. We did not use Generative AI to generate any sections or parts of this paper.

\bibliography{custom,anthology-1,anthology-2}

\appendix

\twocolumn[{%
\noindent{\Large\bfseries Appendix Contents}\par\vspace{8pt}
\noindent\begin{tabular*}{\textwidth}{@{}l@{\extracolsep{\fill}}r@{}}
\multicolumn{1}{@{}p{0.9\textwidth}}{\textbf{\ref{app:grp_predict}\hspace{0.5em}More Results on Predicting Memorization}} & \pageref{app:grp_predict} \\[2pt]
\multicolumn{1}{@{}p{0.9\textwidth}}{\hspace{1.2em}\ref{app:main_qwen_llama}\hspace{0.5em}Results for Qwen-2.5 1.5B and Llama-3.2 1B} & \pageref{app:main_qwen_llama} \\[1pt]
\multicolumn{1}{@{}p{0.9\textwidth}}{\hspace{1.2em}\ref{app:duplicates}\hspace{0.5em}Training Classifier Including Duplicates} & \pageref{app:duplicates} \\[1pt]
\multicolumn{1}{@{}p{0.9\textwidth}}{\hspace{1.2em}\ref{app:feature_dists}\hspace{0.5em}Per-Feature Distributions} & \pageref{app:feature_dists} \\[1pt]
\multicolumn{1}{@{}p{0.9\textwidth}}{\hspace{1.2em}\ref{app:distillation}\hspace{0.5em}Knowledge Distillation} & \pageref{app:distillation} \\[1pt]
\multicolumn{1}{@{}p{0.9\textwidth}}{\hspace{1.2em}\ref{app:approx_mem}\hspace{0.5em}Approximate Memorization} & \pageref{app:approx_mem} \\[1pt]
\multicolumn{1}{@{}p{0.9\textwidth}}{\textbf{\ref{app:grp_flag}\hspace{0.5em}More Results on Flagging Memorization}} & \pageref{app:grp_flag} \\[2pt]
\multicolumn{1}{@{}p{0.9\textwidth}}{\hspace{1.2em}\ref{app:precisionk}\hspace{0.5em}Precision at Audit Budgets} & \pageref{app:precisionk} \\[1pt]
\multicolumn{1}{@{}p{0.9\textwidth}}{\hspace{1.2em}\ref{app:baseline}\hspace{0.5em}Perplexity Ranking Baseline} & \pageref{app:baseline} \\[1pt]
\multicolumn{1}{@{}p{0.9\textwidth}}{\hspace{1.2em}\ref{app:loop}\hspace{0.5em}Reducing Memorization by Removing Flagged Sequences} & \pageref{app:loop} \\[1pt]
\multicolumn{1}{@{}p{0.9\textwidth}}{\textbf{\ref{app:grp_learn}\hspace{0.5em}More Details on Memorization as Learnability}} & \pageref{app:grp_learn} \\[2pt]
\multicolumn{1}{@{}p{0.9\textwidth}}{\hspace{1.2em}\ref{app:learnability}\hspace{0.5em}One-Step Loss Distributions} & \pageref{app:learnability} \\[1pt]
\multicolumn{1}{@{}p{0.9\textwidth}}{\hspace{1.2em}\ref{app:route_examples}\hspace{0.5em}Example Sequences from Each Learning Category} & \pageref{app:route_examples} \\[1pt]
\multicolumn{1}{@{}p{0.9\textwidth}}{\hspace{1.2em}\ref{app:ft_overlap_details}\hspace{0.5em}Computing Fine-Tuning Data Overlap} & \pageref{app:ft_overlap_details} \\[1pt]
\multicolumn{1}{@{}p{0.9\textwidth}}{\hspace{1.2em}\ref{app:ft_overlap}\hspace{0.5em}Overlap with the Fine-Tuning Data} & \pageref{app:ft_overlap} \\[1pt]
\multicolumn{1}{@{}p{0.9\textwidth}}{\textbf{\ref{app:grp_scaling}\hspace{0.5em}More Details on Scaling Laws}} & \pageref{app:grp_scaling} \\[2pt]
\multicolumn{1}{@{}p{0.9\textwidth}}{\hspace{1.2em}\ref{app:scaling_fit}\hspace{0.5em}Fitting the Memorization Scaling Laws} & \pageref{app:scaling_fit} \\[1pt]
\multicolumn{1}{@{}p{0.9\textwidth}}{\hspace{1.2em}\ref{app:ladder_auc}\hspace{0.5em}Predictive Power Across Qwen-2.5 Ladder} & \pageref{app:ladder_auc} \\[1pt]
\multicolumn{1}{@{}p{0.9\textwidth}}{\hspace{1.2em}\ref{app:budget_sweep}\hspace{0.5em}Predictive-Power Exponents Across Audit Budgets} & \pageref{app:budget_sweep} \\[1pt]
\multicolumn{1}{@{}p{0.9\textwidth}}{\textbf{\ref{app:gpu_hours}\hspace{0.5em}GPU Hours}} & \pageref{app:gpu_hours} \\[2pt]
\end{tabular*}\par\vspace{16pt}
}]

\section{More Results on Predicting Memorization}
\label{app:grp_predict}

\subsection{Results for Qwen-2.5 1.5B and Llama-3.2 1B}
\label{app:main_qwen_llama}

Table~\ref{tab:main_qwen_llama} shows the 5-fold cross-validation ROC-AUC for Qwen-2.5 1.5B and Llama-3.2 1B, using the same setup and feature categories as Table~\ref{tab:main_pythia}.

\begin{table*}[t]
\centering\small
\begin{tabular}{lcccccccccc}
\toprule
 & \multicolumn{5}{c}{\textit{Qwen-2.5 1.5B}} & \multicolumn{5}{c}{\textit{Llama-3.2 1B}} \\
\cmidrule(lr){2-6} \cmidrule(lr){7-11}
Dataset & Mem.\ rate & Text & Loss & Grad. & All & Mem.\ rate & Text & Loss & Grad. & All \\
\midrule
FineWeb     & 0.034\% & 0.462 & 0.976 & 0.990 & 0.991 & 0.048\% & 0.566 & 0.978 & 0.993 & 0.994 \\
PG-19       & 0.061\% & 0.773 & 0.994 & 0.998 & 0.994 & 0.16\%  & 0.753 & 0.996 & 0.996 & 0.996 \\
OpenWebMath & 0.82\%  & 0.666 & 0.961 & 0.968 & 0.991 & 0.69\%  & 0.676 & 0.941 & 0.984 & 0.986 \\
The Stack   & 3.97\%  & 0.624 & 0.936 & 0.951 & 0.972 & 3.23\%  & 0.643 & 0.858 & 0.914 & 0.938 \\
WildChat    & 4.60\%  & 0.723 & 0.850 & 0.860 & 0.899 & 6.12\%  & 0.733 & 0.847 & 0.807 & 0.875 \\
Enron       & 2.94\%  & 0.684 & 0.849 & 0.842 & 0.892 & 4.11\%  & 0.621 & 0.795 & 0.792 & 0.850 \\
\bottomrule
\end{tabular}
\caption{Predicting fine-tuning memorization from base model features (deduplicated setup). We report 5-fold cross-validation ROC-AUC by feature categories: Text (zlib entropy), Loss (perplexity and loss variance), Grad.\ (gradient norm and one-step loss change), and All (combined), for Qwen-2.5 1.5B and Llama-3.2 1B. Values are from a single fine-tuning run. Results for Pythia-1.4B are in Table~\ref{tab:main_pythia}. Random chance AUC is 0.5.}
\label{tab:main_qwen_llama}
\end{table*}

\subsection{Training Classifier Including Duplicates}
\label{app:duplicates}

We use the deduplicated setup as our main setup in Section~\ref{sec:memorization_is_predictable}, where we remove exact and near-duplicate sequences and drop the duplicate-count features to predict memorization that is not explained by duplication. In this section, we report the results for the non-deduplicated setup. We add the exact and approximate duplicate counts as additional features to capture duplication. We detect approximate duplicates via MinHash-LSH. We first shingle each sequence into token 5-grams, then group any two sequences whose estimated Jaccard similarity is at least 0.8.

\begin{figure*}[t]
\centering
\includegraphics[width=\textwidth]{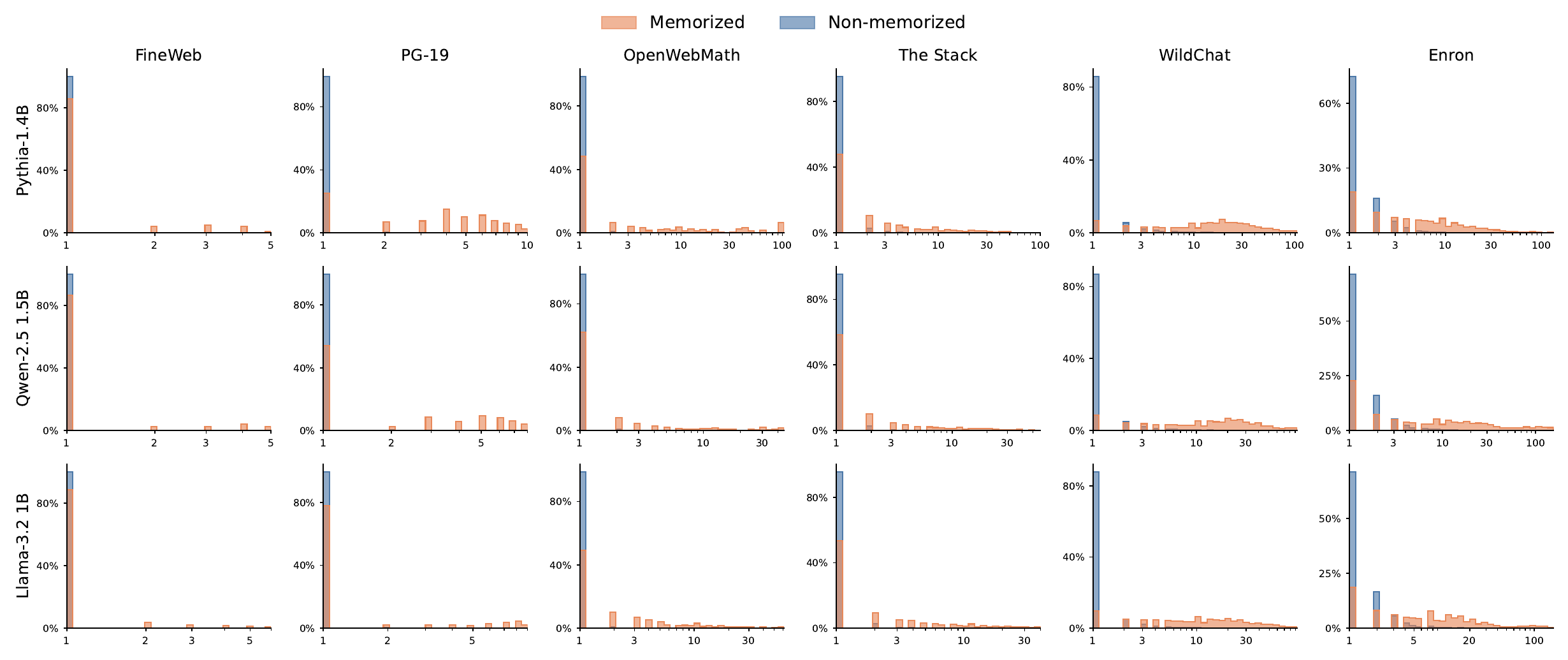}
\caption{Distribution of the approximate duplicate count for memorized versus non-memorized sequences in the contamination-free setup, for all three model families (rows) and six datasets (columns). Each class is normalized so its bars sum to $100\%$, and the duplicate-count axis is log-scaled. Non-memorized sequences are almost all unique (count $1$) while memorized sequences are frequently duplicated.}
\label{fig:dup_hist}
\end{figure*}

Tables~\ref{tab:dup_pythia} and~\ref{tab:dup_qwen_llama} report 5-fold cross-validation ROC-AUC by feature tier for Pythia-1.4B, Qwen-2.5 1.5B, and Llama-3.2 1B. The combined feature set (All) continues to perform strongly (0.93 to 0.999). However, the duplication features now seem to show high ROC-AUC scores on datasets that have a high level of duplication. For instance, for Pythia-1.4B on WildChat the AUC is 0.96, compared to 0.79 for Loss and 0.76 for gradient-based features. Similarly, the duplicate features combined with text beat the other loss and gradient features on WildChat and Enron for Llama and Qwen models. As seen in Figure~\ref{fig:dup_hist}, memorized sequences seem to have a high amount of duplication, especially for the WildChat and Enron datasets. This explains why duplication-driven features largely capture the majority of memorization.

\begin{table}[t]
\centering\small
\begin{tabular}{lcccc}
\toprule
Dataset & Text + Dup & Loss & Gradient & All \\
\midrule
FineWeb     & 0.676 & 0.979 & 0.987 & 0.992 \\
PG-19       & 0.888 & 0.997 & 0.998 & 0.996 \\
OpenWebMath & 0.912 & 0.963 & 0.974 & 0.993 \\
The Stack   & 0.855 & 0.935 & 0.953 & 0.977 \\
WildChat    & 0.960 & 0.791 & 0.765 & 0.972 \\
Enron       & 0.880 & 0.956 & 0.970 & 0.985 \\
\bottomrule
\end{tabular}
\caption{Pythia-1.4B non-deduplicated setup: 5-fold cross-validation ROC-AUC by feature tier. Text + Dup = zlib entropy and the exact and approximate duplicate counts; Loss = perplexity and loss variance; Gradient = gradient norm and one-step loss change; All = combined (8 features).}
\label{tab:dup_pythia}
\end{table}

\begin{table*}[t!]
\centering\small
\begin{tabular}{lcccccccc}
\toprule
 & \multicolumn{4}{c}{\textit{Qwen-2.5 1.5B}} & \multicolumn{4}{c}{\textit{Llama-3.2 1B}} \\
\cmidrule(lr){2-5} \cmidrule(lr){6-9}
Dataset & Text + Dup & Loss & Gradient & All & Text + Dup & Loss & Gradient & All \\
\midrule
FineWeb     & 0.661 & 0.988 & 0.996 & 0.997 & 0.655 & 0.990 & 0.996 & 0.998 \\
PG-19       & 0.858 & 0.988 & 0.996 & 0.999 & 0.801 & 0.989 & 0.994 & 0.997 \\
OpenWebMath & 0.849 & 0.985 & 0.988 & 0.996 & 0.868 & 0.954 & 0.982 & 0.994 \\
The Stack   & 0.824 & 0.960 & 0.968 & 0.986 & 0.836 & 0.893 & 0.928 & 0.970 \\
WildChat    & 0.950 & 0.761 & 0.782 & 0.972 & 0.948 & 0.755 & 0.756 & 0.965 \\
Enron       & 0.876 & 0.802 & 0.810 & 0.943 & 0.878 & 0.763 & 0.758 & 0.932 \\
\bottomrule
\end{tabular}
\caption{Non-deduplicated setup, 5-fold cross-validation ROC-AUC by feature tier for Qwen-2.5 1.5B and Llama-3.2 1B. Text + Dup = zlib entropy and the exact and approximate duplicate counts; Loss = perplexity and loss variance; Gradient = gradient norm and one-step loss change; All = combined.}
\label{tab:dup_qwen_llama}
\end{table*}

\subsection{Per-Feature Distributions}
\label{app:feature_dists}

\begin{figure*}[p]
\centering
{\small\textbf{Qwen-2.5 1.5B}}\par\vspace{2pt}
\includegraphics[width=\textwidth]{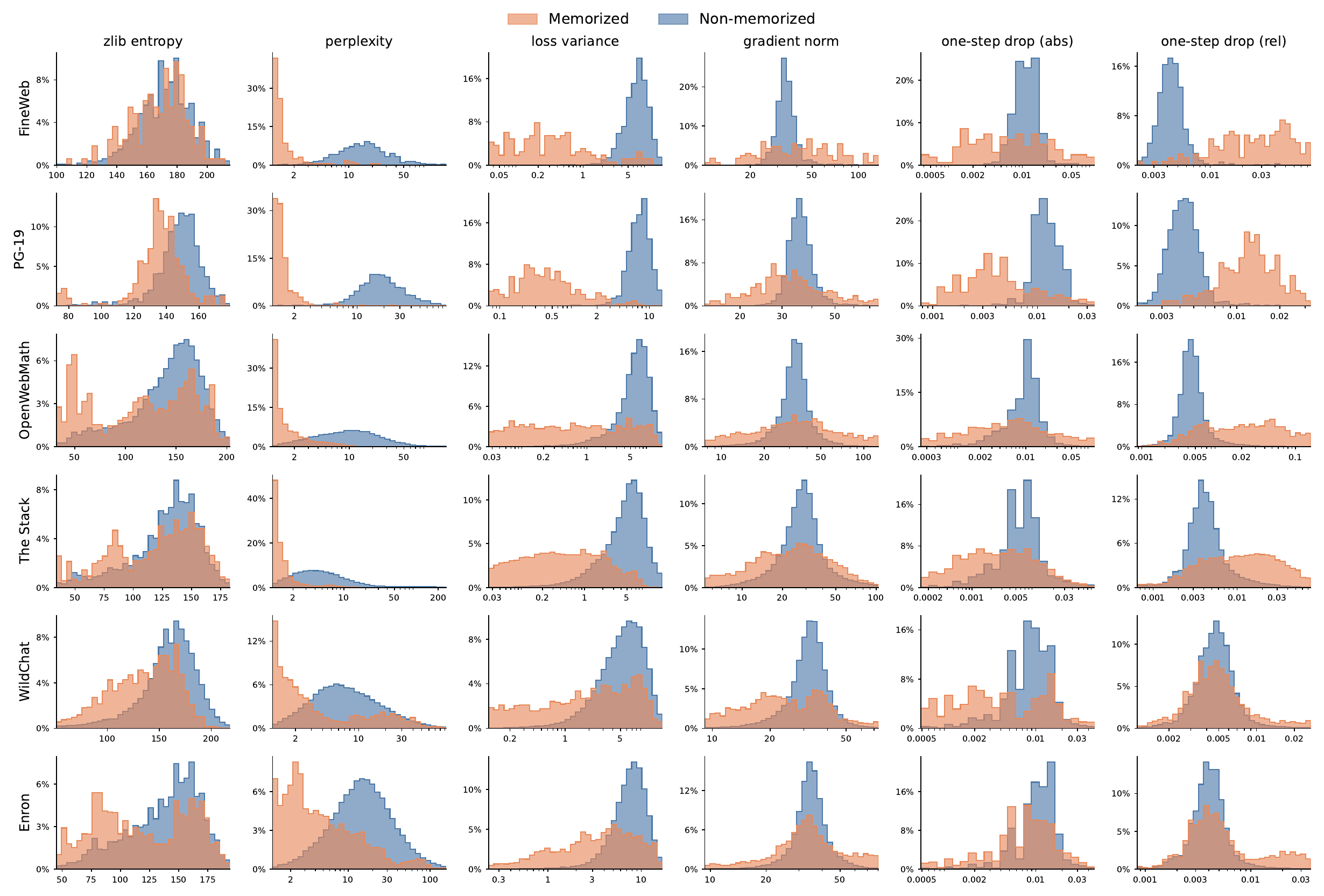}\par\vspace{8pt}
{\small\textbf{Llama-3.2 1B}}\par\vspace{2pt}
\includegraphics[width=\textwidth]{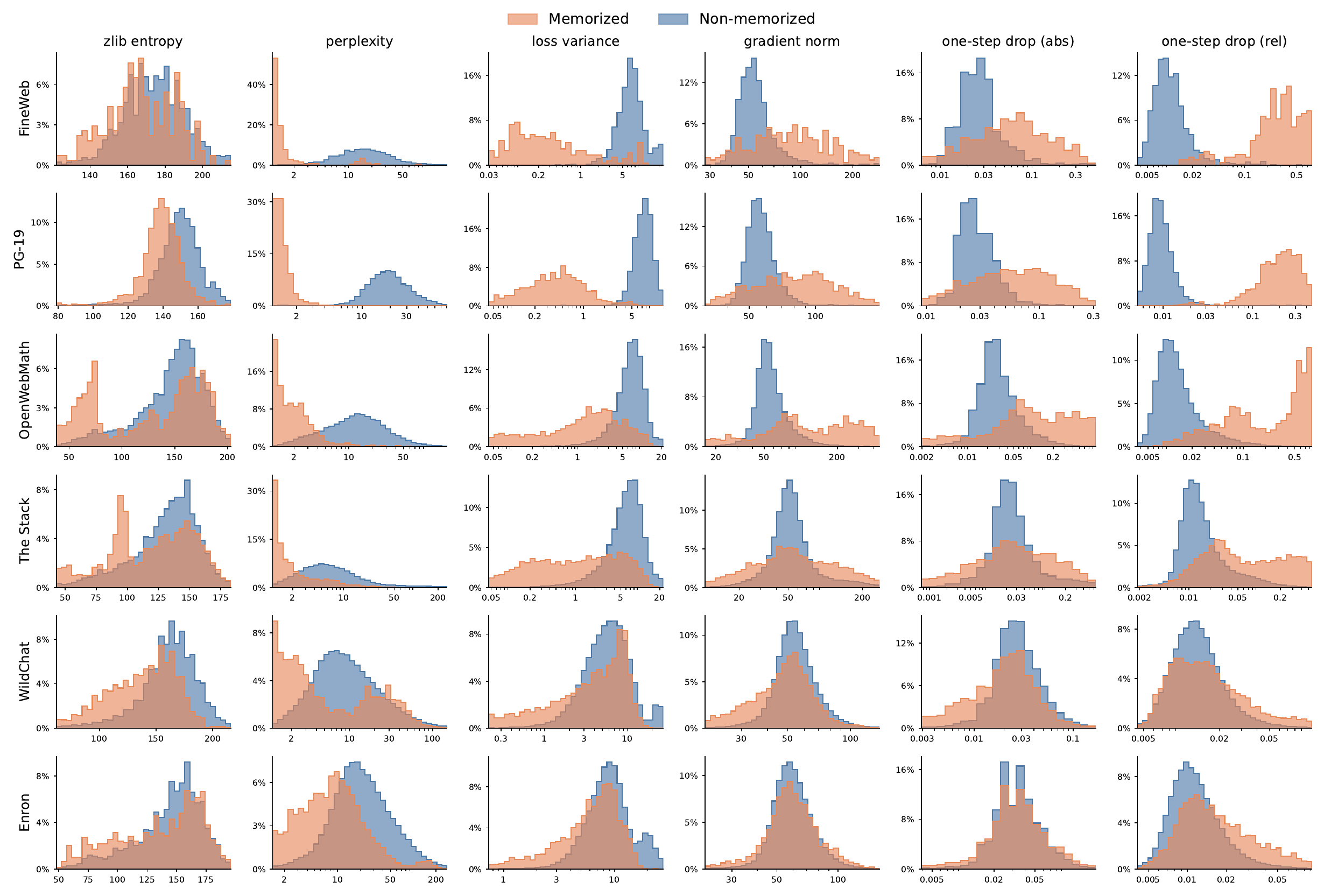}
\caption{Per-feature distributions for memorized versus non-memorized sequences under Qwen-2.5 1.5B (top) and Llama-3.2 1B (bottom), across all six datasets (rows) and all six features (columns). Results are for deduplicated and contamination-free setup. All axes except zlib entropy are log-scaled.}
\label{fig:feat_qwenllama}
\end{figure*}

Section~\ref{subsec:feature_type} shows the distribution of each of our six features for memorized versus non-memorized sequences under Pythia-1.4B (Figure~\ref{fig:feat_pythia}). Figure~\ref{fig:feat_qwenllama} shows the same for Qwen-2.5 1.5B and Llama-3.2 1B. We use the deduplicated and contamination-free setup from Table~\ref{tab:main_pythia}.

\subsection{Knowledge Distillation}
\label{app:distillation}

Our main experiments study whether we can predict memorization in a target model fine-tuned using conventional cross-entropy loss. We further study whether we can predict memorization in a knowledge distillation setup where a smaller model is distilled from a larger teacher model~\citep{hinton2015distillingknowledgeneuralnetwork} using forward KL divergence loss~\citep{Kullback1951OnIA, kodistillm}. We use a Qwen-2.5 7B model fine-tuned on The Stack as the teacher and distill it into a Qwen-2.5 1.5B student with a forward-KL objective for one epoch and temperature of 1. We then apply the exact classifier training and evaluation setup from Section~\ref{sec:exp_setup} where we try to predict the student's memorization from the 1.5B base model's features.

The distilled student memorizes $3.70\%$ of its training data, close to the $3.72\%$ we observe when fine-tuning Qwen-2.5 1.5B directly on the same data. The 5-fold cross-validation ROC-AUC is $0.984$ with all features and $0.976$ with gradient features alone. For recall@$k$, our classifier's performance aligns with what we observe for direct fine-tuning (Table~\ref{tab:distill}). For example, recall@$10\%$ is $90.7\%$ here against $86.0\%$ for direct fine-tuning on the same dataset (Table~\ref{tab:recallk}). 

\begin{table}[t]
\centering\small
\setlength{\tabcolsep}{4pt}
\begin{tabular}{@{}lcccccc@{}}
\toprule
Features & AUC & $0.1\%$ & $0.5\%$ & $1\%$ & $5\%$ & $10\%$ \\
\midrule
All (6)       & 0.971 & 2.1 & 12.8 & 26.0 & 80.2 & 90.7 \\
Gradient only & 0.967 & 2.5 & 13.3 & 26.6 & 78.2 & 89.7 \\
\bottomrule
\end{tabular}
\caption{Predicting memorization in a knowledge distillation setup (Qwen-2.5 7B teacher distilled into a 1.5B student on The Stack). We report the held-out ROC-AUC and recall@$k$ (\%) at audit budgets $k$, using the same setup as Table~\ref{tab:recallk}.}
\label{tab:distill}
\end{table}

\subsection{Approximate Memorization}
\label{app:approx_mem}
We next test whether our classifier also predicts looser forms of memorization, such as approximate memorization. Closely following \citet{ippolito-etal-2023-preventing}, we call a sequence memorized if the BLEU score~\citep{papineni-etal-2002-bleu} between its true continuation and the model's generation is at least 0.75. We run this study on Qwen-2.5 1.5B fine-tuned on The Stack and WildChat in the deduplicated setup. Under this definition, about three times as many sequences count as memorized as under exact match, with a memorization rate of $7.2\%$ on The Stack and $1.9\%$ on WildChat. Our results transfer cleanly to this setup (Table~\ref{tab:approx_mem}). The 5-fold cross-validation ROC-AUC with all six features stays high at $0.960$ on The Stack and $0.931$ on WildChat.

\begin{table}[t]
\centering\small
\begin{tabular}{lc}
\toprule
Dataset & Approximate \\
\midrule
The Stack & 0.960 \\
WildChat  & 0.931 \\
\bottomrule
\end{tabular}
\caption{5-fold cross-validation ROC-AUC (all six features) for Qwen-2.5 1.5B under approximate memorization (BLEU $\geq 0.75$) in the deduplicated setup.}
\label{tab:approx_mem}
\end{table}

\section{More Results on Flagging Memorization}
\label{app:grp_flag}

Here we report two additional experiments on how well our classifier flags memorization. Section~\ref{app:precisionk} reports precision@$k$, the fraction of the inspected sequences that are truly memorized. Section~\ref{app:baseline} compares our classifier against a simple baseline that ranks sequences by their base-model perplexity alone.

\subsection{Precision at Audit Budgets}
\label{app:precisionk}

Table~\ref{tab:recallk} reports the recall@$k$ performance of our classifier. Here we report precision@$k$, the fraction of the inspected top $k\%$ that is truly memorized, at the same budgets and on the same held-out run (Table~\ref{tab:precisionk}). On datasets with low memorization rates, most of the memorized sequences are already captured at a small budget. As the budget increases, precision falls as majority of the memorized sequences are already flagged. This is evident on the low-prevalence datasets (FineWeb, PG-19, OpenWebMath). At tight budgets, the classifier operates far above the natural memorization rate. For example, on FineWeb the precision@$0.1\%$ of $5.7\%$ for Pythia-1.4B is roughly $380\times$ the $0.015\%$ base rate. On the more frequently memorized The Stack and WildChat, the top $1\%$ is $42$ to $59\%$ truly memorized for Qwen-2.5 1.5B. The natural memorization rate of each dataset is reported in Table~\ref{tab:main_pythia}.

\begin{table*}[t]
\centering\small
\begin{minipage}[t]{0.48\linewidth}\centering
\begin{tabular}{lccccc}
\toprule
\multicolumn{6}{c}{\textit{Pythia-1.4B}} \\
\cmidrule(lr){1-6}
Dataset & $0.1\%$ & $0.5\%$ & $1\%$ & $5\%$ & $10\%$ \\
\midrule
FineWeb     & 5.7  & 1.6  & 0.9  & 0.2  & 0.1  \\
PG-19       & 16.8 & 11.8 & 6.1  & 1.2  & 0.6  \\
OpenWebMath & 33.0 & 12.8 & 10.7 & 2.6  & 1.3  \\
The Stack   & 58.6 & 52.1 & 46.0 & 19.4 & 10.6 \\
WildChat    & 40.1 & 33.2 & 28.2 & 18.0 & 14.6 \\
\bottomrule
\end{tabular}
\end{minipage}\hfill
\begin{minipage}[t]{0.48\linewidth}\centering
\begin{tabular}{lccccc}
\toprule
\multicolumn{6}{c}{\textit{Qwen-2.5 1.5B}} \\
\cmidrule(lr){1-6}
Dataset & $0.1\%$ & $0.5\%$ & $1\%$ & $5\%$ & $10\%$ \\
\midrule
FineWeb     & 14.8 & 4.8  & 2.4  & 0.5  & 0.2  \\
PG-19       & 26.3 & 11.1 & 6.0  & 1.3  & 0.7  \\
OpenWebMath & 15.8 & 36.8 & 25.8 & 6.4  & 3.5  \\
The Stack   & 69.8 & 65.4 & 58.5 & 30.2 & 17.2 \\
WildChat    & 59.9 & 50.4 & 41.8 & 23.0 & 17.1 \\
\bottomrule
\end{tabular}
\end{minipage}

\vspace{1.2em}

\begin{tabular}{lccccc}
\toprule
\multicolumn{6}{c}{\textit{Llama-3.2 1B}} \\
\cmidrule(lr){1-6}
Dataset & $0.1\%$ & $0.5\%$ & $1\%$ & $5\%$ & $10\%$ \\
\midrule
FineWeb     & 21.0 & 7.3  & 3.7  & 0.8  & 0.4  \\
PG-19       & 24.9 & 13.3 & 7.7  & 1.7  & 0.9  \\
OpenWebMath & 0.0  & 39.9 & 34.1 & 8.8  & 4.5  \\
The Stack   & 68.3 & 60.2 & 54.3 & 24.9 & 14.5 \\
WildChat    & 57.9 & 50.0 & 39.9 & 23.8 & 19.8 \\
\bottomrule
\end{tabular}
\caption{\textbf{Precision at audit budgets}: precision@$k$ (\% of the inspected top $k\%$ of sequences that is truly memorized) at budgets $k$ for Pythia-1.4B, Qwen-2.5 1.5B, and Llama-3.2 1B, on the held-out dataset of 1M sequences.}
\label{tab:precisionk}
\end{table*}

\subsection{Perplexity Ranking Baseline}
\label{app:baseline}

Next, we compare the performance of our learned classifier with perplexity-only baseline where we simply rank the sequences according to their base model perplexity. Perplexity is the most common memorization heuristic in prior work~\citep{secret-sharer, gpt2_extraction, prashanth2025recite}, and does not require a labeled fine-tuning run. Table~\ref{tab:baseline} compares our classifier vs perplexity-only baseline at a $1\%$ audit budget on the same held-out run as Table~\ref{tab:recallk}. The first ranks sequences by base-model perplexity alone and the second is our full classifier. 

Our full classifier recovers more memorization than the perplexity baseline on every dataset and model family except Pythia-1.4B on WildChat, where the two are tied, and the gap is largest on OpenWebMath, where recall@$1\%$ roughly doubles for Pythia-1.4B and Llama-3.2 1B (from $39.1\%$ to $73.5\%$ and from $35.4\%$ to $70.3\%$), and rises from $50.9\%$ to $71.5\%$ for Qwen-2.5 1.5B. 

\begin{table}[t]
\centering\small
\begin{tabular}{lcc}
\toprule
Dataset & Base ppl & Full clf \\
\midrule
\multicolumn{3}{c}{\textit{Pythia-1.4B}} \\
\cmidrule(lr){1-3}
FineWeb     & 89.1 & 94.6 \\
PG-19       & 96.4 & 97.9 \\
OpenWebMath & 39.1 & 73.5 \\
The Stack   & 27.5 & 32.8 \\
WildChat    & 7.3  & 7.2  \\
\midrule
\multicolumn{3}{c}{\textit{Qwen-2.5 1.5B}} \\
\cmidrule(lr){1-3}
FineWeb     & 87.3 & 99.2 \\
PG-19       & 69.4 & 85.4 \\
OpenWebMath & 50.9 & 71.5 \\
The Stack   & 24.2 & 29.3 \\
WildChat    & 7.7  & 9.0  \\
\midrule
\multicolumn{3}{c}{\textit{Llama-3.2 1B}} \\
\cmidrule(lr){1-3}
FineWeb     & 87.3 & 92.3 \\
PG-19       & 68.4 & 85.4 \\
OpenWebMath & 35.4 & 70.3 \\
The Stack   & 20.3 & 26.5 \\
WildChat    & 6.2  & 6.5  \\
\bottomrule
\end{tabular}
\caption{Recall@$1\%$ comparison between our classifier and perplexity-only baseline. \emph{Base ppl} ranks sequences by base-model perplexity alone and \emph{Full clf} is our full classifier. The full classifier beats the perplexity baseline on every dataset and model family except Pythia-1.4B on WildChat, where the two are tied.}
\label{tab:baseline}
\end{table}

\subsection{Reducing Memorization by Removing Flagged Sequences}
\label{app:loop}

We conduct this experiment on Qwen-2.5 1.5B fine-tuned on The Stack over three fine-tuning seeds (Table~\ref{tab:loop}). We find that removing the top $10\%$ predicted sequences nearly halves the memorization rate, from $3.71\%$ to $1.86\%$, while removing a random $10\%$ barely changes it ($3.65\%$). Since both conditions remove the same amount of data, the reduction in memorization comes from removing the predicted sequences and not from training on less data.  

\begin{table}[t]
\centering\small
\begin{tabular}{lc}
\toprule
Remove $10\%$, re-fine-tune & Mem.\ rate (\%) \\
\midrule
Full data (no removal) & $3.71 \pm 0.02$ \\
Random $10\%$           & $3.65 \pm 0.01$ \\
Predicted top $10\%$    & $\mathbf{1.86 \pm 0.00}$ \\
\bottomrule
\end{tabular}
\caption{Reducing memorization on Qwen-2.5 1.5B and The Stack. We remove $10\%$ of the fine-tuning set and fine-tune the model again, reporting the memorization rate (mean $\pm$ standard deviation over three fine-tuning seeds). Removing the top $10\%$ ranked by predicted risk nearly halves memorization, while removing a random $10\%$ of the same size barely changes it.}
\label{tab:loop}
\end{table}

\begin{table*}[t]
\centering\small
\setlength{\tabcolsep}{4pt}
\resizebox{\textwidth}{!}{\begin{tabular}{@{}lcccccccccccc@{}}
\toprule
 & \multicolumn{2}{c}{Held-out ppl} & \multicolumn{2}{c}{LAMBADA} & \multicolumn{2}{c}{WinoGrande} & \multicolumn{2}{c}{HellaSwag} & \multicolumn{2}{c}{PIQA} & \multicolumn{2}{c}{ARC-Easy} \\
\cmidrule(lr){2-3} \cmidrule(lr){4-5} \cmidrule(lr){6-7} \cmidrule(lr){8-9} \cmidrule(lr){10-11} \cmidrule(lr){12-13}
Dataset & Full & Removed & Full & Removed & Full & Removed & Full & Removed & Full & Removed & Full & Removed \\
\midrule
The Stack   & 6.13 & 6.14 & 58.2$\pm$0.7 & 58.5$\pm$0.7 & 63.1$\pm$1.4 & 63.2$\pm$1.4 & 67.6$\pm$0.5 & 67.6$\pm$0.5 & 74.5$\pm$1.0 & 75.4$\pm$1.0 & 73.6$\pm$0.9 & 72.4$\pm$0.9 \\
OpenWebMath & 10.89 & 10.92 & 62.5$\pm$0.7 & 62.6$\pm$0.7 & 65.3$\pm$1.3 & 64.5$\pm$1.3 & 66.6$\pm$0.5 & 66.4$\pm$0.5 & 75.8$\pm$1.0 & 75.3$\pm$1.0 & 73.4$\pm$0.9 & 73.7$\pm$0.9 \\
FineWeb     & 18.39 & 18.45 & 62.3$\pm$0.7 & 62.4$\pm$0.7 & 65.1$\pm$1.3 & 65.2$\pm$1.3 & 66.0$\pm$0.5 & 65.7$\pm$0.5 & 75.8$\pm$1.0 & 76.1$\pm$1.0 & 71.0$\pm$0.9 & 70.7$\pm$0.9 \\
PG-19       & 18.83 & 18.90 & 53.8$\pm$0.7 & 54.1$\pm$0.7 & 60.5$\pm$1.4 & 62.0$\pm$1.4 & 64.4$\pm$0.5 & 64.4$\pm$0.5 & 74.3$\pm$1.0 & 74.4$\pm$1.0 & 73.6$\pm$0.9 & 72.0$\pm$0.9 \\
WildChat    & 6.17 & 6.27 & 58.5$\pm$0.7 & 58.6$\pm$0.7 & 64.7$\pm$1.3 & 63.1$\pm$1.4 & 66.7$\pm$0.5 & 66.6$\pm$0.5 & 76.2$\pm$1.0 & 75.5$\pm$1.0 & 72.4$\pm$0.9 & 72.7$\pm$0.9 \\
\bottomrule
\end{tabular}}
\caption{Utility of Qwen-2.5 1.5B after fine-tuning on the full data (Full) and after removing the top 10\% of sequences predicted by our classifier as likely to be memorized (Removed). We report perplexity on 10{,}000 held-out sequences from the same domain and zero-shot accuracy (\%) $\pm$ standard error on five benchmarks. We report length-normalized accuracy for HellaSwag, PIQA, and ARC-Easy, and accuracy for LAMBADA and WinoGrande.}
\label{tab:loop_utility}
\end{table*}

Table~\ref{tab:loop_utility_k} shows the utility of Qwen-2.5 1.5B when we remove only the top 1\% or 5\% of sequences ranked by our classifier, for the same domains as Table~\ref{tab:loop_utility}.

\begin{table*}[t]
\centering\small
\setlength{\tabcolsep}{4pt}
\resizebox{\textwidth}{!}{\begin{tabular}{@{}lcccccccccccccccccc@{}}
\toprule
 & \multicolumn{3}{c}{Held-out ppl} & \multicolumn{3}{c}{LAMBADA} & \multicolumn{3}{c}{WinoGrande} & \multicolumn{3}{c}{HellaSwag} & \multicolumn{3}{c}{PIQA} & \multicolumn{3}{c}{ARC-Easy} \\
\cmidrule(lr){2-4} \cmidrule(lr){5-7} \cmidrule(lr){8-10} \cmidrule(lr){11-13} \cmidrule(lr){14-16} \cmidrule(lr){17-19}
Dataset & Full & Top 1\% & Top 5\% & Full & Top 1\% & Top 5\% & Full & Top 1\% & Top 5\% & Full & Top 1\% & Top 5\% & Full & Top 1\% & Top 5\% & Full & Top 1\% & Top 5\% \\
\midrule
The Stack   & 6.13 & 6.13 & 6.13 & 58.2 & 57.7 & 58.5 & 63.1 & 63.3 & 63.7 & 67.6 & 67.7 & 67.7 & 74.5 & 74.6 & 75.0 & 73.6 & 72.6 & 72.8 \\
OpenWebMath & 10.89 & 10.90 & 10.91 & 62.5 & 62.6 & 62.8 & 65.3 & 65.0 & 65.3 & 66.6 & 66.2 & 66.4 & 75.8 & 75.5 & 75.4 & 73.4 & 73.3 & 74.2 \\
FineWeb     & 18.39 & 18.39 & 18.41 & 62.3 & 62.4 & 62.5 & 65.1 & 64.6 & 65.3 & 66.0 & 65.9 & 65.7 & 75.8 & 75.8 & 75.9 & 71.0 & 70.9 & 70.6 \\
PG-19       & 18.83 & 18.86 & 18.87 & 53.8 & 53.4 & 53.1 & 60.5 & 61.2 & 60.3 & 64.4 & 64.4 & 64.4 & 74.3 & 74.3 & 74.4 & 73.6 & 73.9 & 71.8 \\
WildChat    & 6.17 & 6.18 & 6.21 & 58.5 & 58.4 & 58.5 & 64.7 & 64.4 & 64.6 & 66.7 & 66.9 & 66.8 & 76.2 & 76.0 & 75.5 & 72.4 & 73.7 & 73.6 \\
\bottomrule
\end{tabular}}
\caption{Utility of Qwen-2.5 1.5B after fine-tuning on the full data (Full) and after removing the top 1\% or 5\% of sequences ranked by our classifier. We report perplexity on 10{,}000 held-out sequences from the same domain and zero-shot accuracy (\%) on five benchmarks.}
\label{tab:loop_utility_k}
\end{table*}

\section{More Details on Memorization as Learnability}
\label{app:grp_learn}

\subsection{One-Step Loss Distributions}
\label{app:learnability}

\begin{figure*}[t]
\centering
\includegraphics[width=\textwidth]{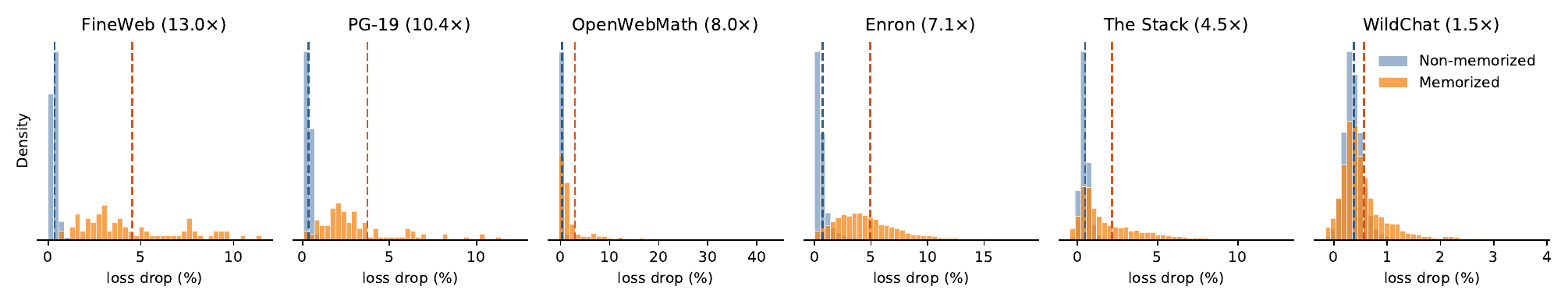}
\caption{Distribution of the one-step \emph{relative} loss drop for memorized versus non-memorized sequences under Pythia-1.4B on all six datasets. Dashed lines mark the class means. On average, memorized sequences are learned faster in every case. The corresponding means and ratios are in Table~\ref{tab:learnability}.}
\label{fig:learnability}
\end{figure*}

\begin{figure*}[t]
\centering
\includegraphics[width=\textwidth]{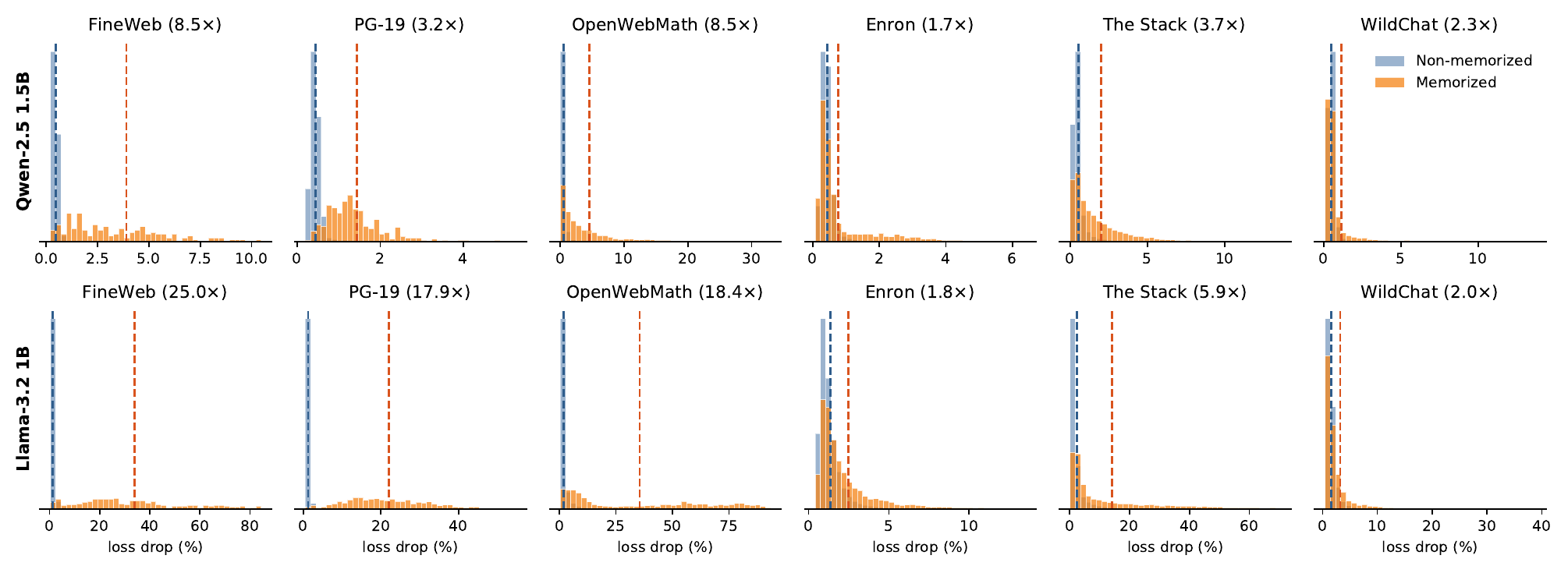}
\caption{One-step \emph{relative} loss drop for memorized versus non-memorized sequences for Qwen-2.5 1.5B (top) and Llama-3.2 1B (bottom) on all six datasets. Dashed line denotes the mean. On average, memorized sequences are learned faster in every case. Results are for deduplicated and contamination-free setup.}
\label{fig:learnability_grid}
\end{figure*}

Table~\ref{tab:learnability} reports the mean one-step relative loss reduction for memorized and non-memorized sequences for all models. Figures~\ref{fig:learnability} and~\ref{fig:learnability_grid} show the distributions for Pythia-1.4B, Qwen-2.5 1.5B and Llama-3.2 1B. 

\begin{table*}[t]
\centering\small
\setlength{\tabcolsep}{5pt}
\begin{tabular}{@{}lccccccccc@{}}
\toprule
 & \multicolumn{3}{c}{Pythia-1.4B} & \multicolumn{3}{c}{Qwen-2.5 1.5B} & \multicolumn{3}{c}{Llama-3.2 1B} \\
\cmidrule(lr){2-4}\cmidrule(lr){5-7}\cmidrule(lr){8-10}
Dataset & Mem. & Non-mem.\ & Factor & Mem. & Non-mem.\ & Factor & Mem. & Non-mem.\ & Factor \\
\midrule
FineWeb     & 4.53 & 0.35 & $13.0\times$ & 3.91 & 0.46 & $8.5\times$ & 34.03 & 1.36 & $25.0\times$ \\
PG-19       & 3.73 & 0.36 & $10.4\times$ & 1.44 & 0.45 & $3.2\times$ & 22.04 & 1.23 & $17.9\times$ \\
OpenWebMath & 2.93 & 0.36 & $8.0\times$ & 4.53 & 0.53 & $8.5\times$ & 35.53 & 1.93 & $18.4\times$ \\
The Stack   & 2.14 & 0.48 & $4.5\times$ & 1.99 & 0.54 & $3.7\times$ & 14.24 & 2.42 & $5.9\times$ \\
Enron       & 4.93 & 0.70 & $7.1\times$ & 0.77 & 0.44 & $1.7\times$ & 2.47 & 1.36 & $1.8\times$ \\
WildChat    & 0.57 & 0.37 & $1.5\times$ & 1.16 & 0.51 & $2.3\times$ & 3.22 & 1.58 & $2.0\times$ \\
\bottomrule
\end{tabular}
\caption{Mean one-step \emph{relative} loss reduction for memorized versus non-memorized sequences in the deduplicated and contamination-free setup. Factor is the ratio of the two means.}
\label{tab:learnability}
\end{table*}

\subsection{Example Sequences from Each Learning Category}
\label{app:route_examples}

\definecolor{rtLow}{HTML}{1BAF7A}\definecolor{rtFast}{HTML}{2A78D6}\definecolor{rtSlow}{HTML}{EB6834}

We show three memorized sequences from each group for Pythia-1.4B on The Stack, OpenWebMath, FineWeb, and PG-19. The examples come from the same sequences as Table~\ref{tab:three_routes} and are drawn in a random order (seed 0). Each box shows the last 20 tokens of the prefix in grey, then the 50-token continuation on which the loss is measured.  

\medskip\noindent{\large\bfseries The Stack}\par\nopagebreak\smallskip\nopagebreak

\noindent\begin{minipage}{\linewidth}
\noindent{\color{rtLow}\rule[-0.1ex]{0.7em}{0.7em}}\hspace{0.45em}{\small\bfseries\color{rtLow!80!black}Category 1 (low starting loss)}\hfill{\scriptsize\color{black!55}$n=1{,}237$}\par\nopagebreak\vspace{2pt}\nopagebreak

\noindent{\setlength{\fboxsep}{4pt}\colorbox{rtLow!7}{\parbox{\dimexpr\linewidth-2\fboxsep\relax}{\raggedright{\scriptsize\sffamily\color{black!65}loss \textbf{0.094}\quad drop \textbf{0.0001}}\\[2pt]{\scriptsize\ttfamily\textcolor{black!45}{\_setting\allowbreak{}s.GetCfl\allowbreak{}agsC(con\allowbreak{}figname){\color{black!35}$\hookleftarrow$}\allowbreak{}  cflags\_cc = self}{\color{rtLow}\bfseries\,$\Vert$\,}.xcode\_s\allowbreak{}ettings.\allowbreak{}GetCflag\allowbreak{}sCC(conf\allowbreak{}igname){\color{black!35}$\hookleftarrow$}\allowbreak{}  cflags\_objc = self.xco\allowbreak{}de\_setti\allowbreak{}ngs.GetC\allowbreak{}flagsObj\allowbreak{}C(config\allowbreak{}name){\color{black!35}$\hookleftarrow$}\allowbreak{}  cflags\_objcc = self.\par}}}}\par
\end{minipage}\par\vspace{4pt}

\noindent{\setlength{\fboxsep}{4pt}\colorbox{rtLow!7}{\parbox{\dimexpr\linewidth-2\fboxsep\relax}{\raggedright{\scriptsize\sffamily\color{black!65}loss \textbf{0.056}\quad drop \textbf{0.0011}}\\[2pt]{\scriptsize\ttfamily\textcolor{black!45}{, route\_fi\allowbreak{}lter\_par\allowbreak{}ameters, custom\_h\allowbreak{}eaders=N\allowbreak{}one, raw=False, **operation\_}{\color{rtLow}\bfseries\,$\Vert$\,}config):{\color{black!35}$\hookleftarrow$}\allowbreak{}  \# Construct URL{\color{black!35}$\hookleftarrow$}\allowbreak{}  url = self.cre\allowbreak{}ate\_or\_u\allowbreak{}pdate.me\allowbreak{}tadata[\textquotesingle{}\allowbreak{}url\textquotesingle{}]{\color{black!35}$\hookleftarrow$}\allowbreak{}  path\_for\allowbreak{}mat\_argu\allowbreak{}ments = \{{\color{black!35}$\hookleftarrow$}\allowbreak{}  \textquotesingle{}resourc\allowbreak{}eGroupNa\allowbreak{}me\textquotesingle{}: self.\_se\allowbreak{}rialize.\allowbreak{}url(\textquotedbl{}res\allowbreak{}ource\_gr\allowbreak{}oup\_name\allowbreak{}\textquotedbl{},\par}}}}\par\vspace{4pt}

\noindent{\setlength{\fboxsep}{4pt}\colorbox{rtLow!7}{\parbox{\dimexpr\linewidth-2\fboxsep\relax}{\raggedright{\scriptsize\sffamily\color{black!65}loss \textbf{0.088}\quad drop \textbf{0.0019}}\\[2pt]{\scriptsize\ttfamily\textcolor{black!45}{0,{\color{black!35}$\hookleftarrow$}\allowbreak{}  number=1, type=11, cpp\_type=10, label}{\color{rtLow}\bfseries\,$\Vert$\,}=3,{\color{black!35}$\hookleftarrow$}\allowbreak{}  has\_defa\allowbreak{}ult\_valu\allowbreak{}e=False, default\_\allowbreak{}value=[]\allowbreak{},{\color{black!35}$\hookleftarrow$}\allowbreak{}  message\_\allowbreak{}type=Non\allowbreak{}e, enum\_typ\allowbreak{}e=None, containi\allowbreak{}ng\_type=\allowbreak{}None,{\color{black!35}$\hookleftarrow$}\allowbreak{}  is\_exten\allowbreak{}sion=Fal\allowbreak{}se, extensio\allowbreak{}n\_scope=\par}}}}\par\vspace{4pt}

\noindent\begin{minipage}{\linewidth}
\noindent{\color{rtFast}\rule[-0.1ex]{0.7em}{0.7em}}\hspace{0.45em}{\small\bfseries\color{rtFast!80!black}Category 2 (faster drop)}\hfill{\scriptsize\color{black!55}$n=3{,}130$}\par\nopagebreak\vspace{2pt}\nopagebreak

\noindent{\setlength{\fboxsep}{4pt}\colorbox{rtFast!7}{\parbox{\dimexpr\linewidth-2\fboxsep\relax}{\raggedright{\scriptsize\sffamily\color{black!65}loss \textbf{2.877}\quad drop \textbf{0.0215}\quad drop larger than for 95\% of sequences with a similar starting loss}\\[2pt]{\scriptsize\ttfamily\textcolor{black!45}{ured must be of a type compatible with boolean\textquotedbl{}\textquotedbl{}\textquotedbl{},{\color{black!35}$\hookleftarrow$}\allowbreak{}  \textquotesingle{}defined\allowbreak{}-{}type\textquotesingle{}: \textquotedbl{}boolean}{\color{rtFast}\bfseries\,$\Vert$\,}\textquotedbl{},{\color{black!35}$\hookleftarrow$}\allowbreak{}  \textquotesingle{}generat\allowbreak{}ed-{}type\textquotesingle{}\allowbreak{}: \textquotedbl{}\textquotedbl{}\textquotedbl{}YANGD\allowbreak{}ynClass(\allowbreak{}base=YAN\allowbreak{}GBool, is\_leaf=\allowbreak{}True, yang\_nam\allowbreak{}e=\textquotedbl{}lsp-{}c\allowbreak{}onfig-{}re\allowbreak{}optimize\allowbreak{}-{}timer-{}c\allowbreak{}onfigure\allowbreak{}d\textquotedbl{}, rest\_name=\textquotedbl{}\par}}}}\par
\end{minipage}\par\vspace{4pt}

\noindent{\setlength{\fboxsep}{4pt}\colorbox{rtFast!7}{\parbox{\dimexpr\linewidth-2\fboxsep\relax}{\raggedright{\scriptsize\sffamily\color{black!65}loss \textbf{1.271}\quad drop \textbf{0.1016}\quad drop larger than for 100\% of sequences with a similar starting loss}\\[2pt]{\scriptsize\ttfamily\textcolor{black!45}{ the source reference on Chart Studio Cloud for{\color{black!35}$\hookleftarrow$}\allowbreak{}  \textasciigrave{}color\textasciigrave{}.{\color{black!35}$\hookleftarrow$}\allowbreak{}  family{\color{black!35}$\hookleftarrow$}\allowbreak{}  HTML font}{\color{rtFast}\bfseries\,$\Vert$\,} family -{} the typeface that will be applied by{\color{black!35}$\hookleftarrow$}\allowbreak{}  the web browser. The web browser will only be able to{\color{black!35}$\hookleftarrow$}\allowbreak{}  apply a font if it is available on the system which it{\color{black!35}$\hookleftarrow$}\allowbreak{}  operates. Provide multiple font families, separated\par}}}}\par\vspace{4pt}

\noindent{\setlength{\fboxsep}{4pt}\colorbox{rtFast!7}{\parbox{\dimexpr\linewidth-2\fboxsep\relax}{\raggedright{\scriptsize\sffamily\color{black!65}loss \textbf{0.267}\quad drop \textbf{0.0051}\quad drop larger than for 86\% of sequences with a similar starting loss}\\[2pt]{\scriptsize\ttfamily\textcolor{black!45}{contrib.\allowbreak{}auth.pas\allowbreak{}sword\_va\allowbreak{}lidation\allowbreak{}.CommonP\allowbreak{}asswordV\allowbreak{}alidator\allowbreak{}\textquotesingle{},{\color{black!35}$\hookleftarrow$}\allowbreak{}  \},{\color{black!35}$\hookleftarrow$}\allowbreak{}  \{{\color{black!35}$\hookleftarrow$}\allowbreak{}  }{\color{rtFast}\bfseries\,$\Vert$\,}\textquotesingle{}NAME\textquotesingle{}: \textquotesingle{}django.\allowbreak{}contrib.\allowbreak{}auth.pas\allowbreak{}sword\_va\allowbreak{}lidation\allowbreak{}.Numeric\allowbreak{}Password\allowbreak{}Validato\allowbreak{}r\textquotesingle{},{\color{black!35}$\hookleftarrow$}\allowbreak{}  \},{\color{black!35}$\hookleftarrow$}\allowbreak{} ]{\color{black!35}$\hookleftarrow$}\allowbreak{} \# Internat\allowbreak{}ionaliza\allowbreak{}tion{\color{black!35}$\hookleftarrow$}\allowbreak{} \# https://\allowbreak{}docs.dja\allowbreak{}ngoproje\allowbreak{}ct.com/e\allowbreak{}n/2.2/to\allowbreak{}pics/\par}}}}\par\vspace{4pt}

\noindent\begin{minipage}{\linewidth}
\noindent{\color{rtModerate}\rule[-0.1ex]{0.7em}{0.7em}}\hspace{0.45em}{\small\bfseries\color{rtModerate!80!black}Category 3 (moderate drop)}\hfill{\scriptsize\color{black!55}$n=2{,}021$}\par\nopagebreak\vspace{2pt}\nopagebreak

\noindent{\setlength{\fboxsep}{4pt}\colorbox{rtModerate!7}{\parbox{\dimexpr\linewidth-2\fboxsep\relax}{\raggedright{\scriptsize\sffamily\color{black!65}loss \textbf{0.414}\quad drop \textbf{0.0024}\quad drop larger than for 58\% of sequences with a similar starting loss}\\[2pt]{\scriptsize\ttfamily\textcolor{black!45}{\{\textquotesingle{}en\textquotesingle{}: \textquotesingle{}Weifang, Shandong\textquotesingle{}, \textquotesingle{}zh\textquotesingle{}: u(\textquotesingle{}\textbackslash{}u5}{\color{rtModerate}\bfseries\,$\Vert$\,}c71\textbackslash{}u4e1\allowbreak{}c\textbackslash{}u7701\textbackslash{}\allowbreak{}u6f4d\textbackslash{}u5\allowbreak{}74a\textbackslash{}u5e0\allowbreak{}2\textquotesingle{})\},{\color{black!35}$\hookleftarrow$}\allowbreak{}  \textquotesingle{}8615264\allowbreak{}44\textquotesingle{}:\{\textquotesingle{}en\allowbreak{}\textquotesingle{}: \textquotesingle{}Weifang, Shandong\textquotesingle{}, \textquotesingle{}zh\par}}}}\par
\end{minipage}\par\vspace{4pt}

\noindent{\setlength{\fboxsep}{4pt}\colorbox{rtModerate!7}{\parbox{\dimexpr\linewidth-2\fboxsep\relax}{\raggedright{\scriptsize\sffamily\color{black!65}loss \textbf{0.157}\quad drop \textbf{0.0020}\quad drop larger than for 67\% of sequences with a similar starting loss}\\[2pt]{\scriptsize\ttfamily\textcolor{black!45}{SocialEx\allowbreak{}pression\allowbreak{}EventTop\allowbreak{}icMessag\allowbreak{}e.{\color{black!35}$\hookleftarrow$}\allowbreak{}  :return: The type of this QueueCon\allowbreak{}versatio\allowbreak{}nSocial}{\color{rtModerate}\bfseries\,$\Vert$\,}Expressi\allowbreak{}onEventT\allowbreak{}opicMess\allowbreak{}age.{\color{black!35}$\hookleftarrow$}\allowbreak{}  :rtype: str{\color{black!35}$\hookleftarrow$}\allowbreak{}  \textquotedbl{}\textquotedbl{}\textquotedbl{}{\color{black!35}$\hookleftarrow$}\allowbreak{}  return self.\_type{\color{black!35}$\hookleftarrow$}\allowbreak{}  @type.setter{\color{black!35}$\hookleftarrow$}\allowbreak{}  def type(self, type):{\color{black!35}$\hookleftarrow$}\allowbreak{}  \textquotedbl{}\textquotedbl{}\textquotedbl{}{\color{black!35}$\hookleftarrow$}\allowbreak{}  Sets the type of this QueueCon\par}}}}\par\vspace{4pt}

\noindent{\setlength{\fboxsep}{4pt}\colorbox{rtModerate!7}{\parbox{\dimexpr\linewidth-2\fboxsep\relax}{\raggedright{\scriptsize\sffamily\color{black!65}loss \textbf{0.886}\quad drop \textbf{0.0039}\quad drop larger than for 46\% of sequences with a similar starting loss}\\[2pt]{\scriptsize\ttfamily\textcolor{black!45}{Instance\allowbreak{}Id\textasciigrave{}\textasciigrave{}.{\color{black!35}$\hookleftarrow$}\allowbreak{}  :see: http://d\allowbreak{}ocs.aws.\allowbreak{}amazon.c\allowbreak{}om/A}{\color{rtModerate}\bfseries\,$\Vert$\,}WSCloudF\allowbreak{}ormation\allowbreak{}/latest/\allowbreak{}UserGuid\allowbreak{}e/aws-{}re\allowbreak{}source-{}s\allowbreak{}sm-{}assoc\allowbreak{}iation.h\allowbreak{}tml\#cfn-{}\allowbreak{}ssm-{}asso\allowbreak{}ciation-{}\allowbreak{}instance\allowbreak{}id{\color{black!35}$\hookleftarrow$}\allowbreak{}  \textquotedbl{}\textquotedbl{}\textquotedbl{}{\color{black!35}$\hookleftarrow$}\allowbreak{}  return jsii.get\allowbreak{}(self, \textquotedbl{}instanc\allowbreak{}eId\textquotedbl{})\par}}}}\par\vspace{4pt}

\noindent\begin{minipage}{\linewidth}
\noindent{\color{rtSlow}\rule[-0.1ex]{0.7em}{0.7em}}\hspace{0.45em}{\small\bfseries\color{rtSlow!80!black}Category 4 (slower drop)}\hfill{\scriptsize\color{black!55}$n=656$}\par\nopagebreak\vspace{2pt}\nopagebreak

\noindent{\setlength{\fboxsep}{4pt}\colorbox{rtSlow!7}{\parbox{\dimexpr\linewidth-2\fboxsep\relax}{\raggedright{\scriptsize\sffamily\color{black!65}loss \textbf{1.000}\quad drop \textbf{0.0024}\quad drop larger than for 21\% of sequences with a similar starting loss}\\[2pt]{\scriptsize\ttfamily\textcolor{black!45}{().strip\allowbreak{}().split\allowbreak{}(){\color{black!35}$\hookleftarrow$}\allowbreak{}  P = []{\color{black!35}$\hookleftarrow$}\allowbreak{}  for tok in toks:{\color{black!35}$\hookleftarrow$}\allowbreak{}  }{\color{rtSlow}\bfseries\,$\Vert$\,}P.append\allowbreak{}(float(t\allowbreak{}ok) / 100){\color{black!35}$\hookleftarrow$}\allowbreak{}  E = []{\color{black!35}$\hookleftarrow$}\allowbreak{}  for i in range(N):{\color{black!35}$\hookleftarrow$}\allowbreak{}  e = 1.0 / (1 -{} P[i]){\color{black!35}$\hookleftarrow$}\allowbreak{}  E.append(e){\color{black!35}$\hookleftarrow$}\allowbreak{}  \par}}}}\par
\end{minipage}\par\vspace{4pt}

\noindent{\setlength{\fboxsep}{4pt}\colorbox{rtSlow!7}{\parbox{\dimexpr\linewidth-2\fboxsep\relax}{\raggedright{\scriptsize\sffamily\color{black!65}loss \textbf{1.424}\quad drop \textbf{0.0039}\quad drop larger than for 23\% of sequences with a similar starting loss}\\[2pt]{\scriptsize\ttfamily\textcolor{black!45}{def accept(self, visitor:\allowbreak{}ParseTre\allowbreak{}eVisitor\allowbreak{}):{\color{black!35}$\hookleftarrow$}\allowbreak{}  if hasattr( visitor, \textquotedbl{}}{\color{rtSlow}\bfseries\,$\Vert$\,}visitGro\allowbreak{}upingSet\allowbreak{}\textquotedbl{} ):{\color{black!35}$\hookleftarrow$}\allowbreak{}  return visitor.\allowbreak{}visitGro\allowbreak{}upingSet\allowbreak{}(self){\color{black!35}$\hookleftarrow$}\allowbreak{}  else:{\color{black!35}$\hookleftarrow$}\allowbreak{}  return visitor.\allowbreak{}visitChi\allowbreak{}ldren(se\allowbreak{}lf){\color{black!35}$\hookleftarrow$}\allowbreak{}  def grouping\allowbreak{}Set(self\allowbreak{}):{\color{black!35}$\hookleftarrow$}\allowbreak{}  localctx = SqlBaseP\allowbreak{}arser.Gr\allowbreak{}oup\par}}}}\par\vspace{4pt}

\noindent{\setlength{\fboxsep}{4pt}\colorbox{rtSlow!7}{\parbox{\dimexpr\linewidth-2\fboxsep\relax}{\raggedright{\scriptsize\sffamily\color{black!65}loss \textbf{1.943}\quad drop \textbf{0.0020}\quad drop larger than for 4\% of sequences with a similar starting loss}\\[2pt]{\scriptsize\ttfamily\textcolor{black!45}{Client(c\allowbreak{}redentia\allowbreak{}ls=ga\_cr\allowbreak{}edential\allowbreak{}s.Anonym\allowbreak{}ousCrede\allowbreak{}ntials()\allowbreak{},){\color{black!35}$\hookleftarrow$}\allowbreak{}  \# Any value}{\color{rtSlow}\bfseries\,$\Vert$\,} that is part of the HTTP/1.1 URI should be sent as{\color{black!35}$\hookleftarrow$}\allowbreak{}  \# a field header. Set these to a non-{}empty value.{\color{black!35}$\hookleftarrow$}\allowbreak{}  request = dlp.GetS\allowbreak{}toredInf\allowbreak{}oTypeReq\allowbreak{}uest(){\color{black!35}$\hookleftarrow$}\allowbreak{}  request.name\par}}}}\par\vspace{4pt}

\medskip\noindent{\large\bfseries OpenWebMath}\par\nopagebreak\smallskip\nopagebreak

\noindent\begin{minipage}{\linewidth}
\noindent{\color{rtLow}\rule[-0.1ex]{0.7em}{0.7em}}\hspace{0.45em}{\small\bfseries\color{rtLow!80!black}Category 1 (low starting loss)}\hfill{\scriptsize\color{black!55}$n=139$}\par\nopagebreak\vspace{2pt}\nopagebreak

\noindent{\setlength{\fboxsep}{4pt}\colorbox{rtLow!7}{\parbox{\dimexpr\linewidth-2\fboxsep\relax}{\raggedright{\scriptsize\sffamily\color{black!65}loss \textbf{0.095}\quad drop \textbf{0.0029}}\\[2pt]{\scriptsize\ttfamily\textcolor{black!45}{ Traveled By No. 1{\color{black!35}$\hookleftarrow$}\allowbreak{} Photography{\color{black!35}$\hookleftarrow$}\allowbreak{} Series: The One Less Traveled By{\color{black!35}$\hookleftarrow$}\allowbreak{} The}{\color{rtLow}\bfseries\,$\Vert$\,} One Less Traveled By No. 2{\color{black!35}$\hookleftarrow$}\allowbreak{} Photography{\color{black!35}$\hookleftarrow$}\allowbreak{} Series: The One Less Traveled By{\color{black!35}$\hookleftarrow$}\allowbreak{} The One Less Traveled By No. 3{\color{black!35}$\hookleftarrow$}\allowbreak{} Photography{\color{black!35}$\hookleftarrow$}\allowbreak{} Series: The One Less Traveled By{\color{black!35}$\hookleftarrow$}\allowbreak{} The One Less Traveled By No\par}}}}\par
\end{minipage}\par\vspace{4pt}

\noindent{\setlength{\fboxsep}{4pt}\colorbox{rtLow!7}{\parbox{\dimexpr\linewidth-2\fboxsep\relax}{\raggedright{\scriptsize\sffamily\color{black!65}loss \textbf{0.063}\quad drop \textbf{0.0012}}\\[2pt]{\scriptsize\ttfamily\textcolor{black!45}{,50,51,5\allowbreak{}2,53,54,\allowbreak{}55,56,57\allowbreak{},58,59}{\color{rtLow}\bfseries\,$\Vert$\,},60),(61\allowbreak{},62,63,6\allowbreak{}4,65,66,\allowbreak{}67,68,69\allowbreak{},70,71,7\allowbreak{}2),(73,7\allowbreak{}4,75,76,\allowbreak{}77,78,79\allowbreak{},80,81,8\allowbreak{}2,83,84\par}}}}\par\vspace{4pt}

\noindent{\setlength{\fboxsep}{4pt}\colorbox{rtLow!7}{\parbox{\dimexpr\linewidth-2\fboxsep\relax}{\raggedright{\scriptsize\sffamily\color{black!65}loss \textbf{0.095}\quad drop \textbf{0.0035}}\\[2pt]{\scriptsize\ttfamily\textcolor{black!45}{,38,39,4\allowbreak{}0,41,42)\allowbreak{}(43,44,4\allowbreak{}5,46,47}{\color{rtLow}\bfseries\,$\Vert$\,},48)(49,\allowbreak{}50,51,52\allowbreak{},53,54)(\allowbreak{}55,56,57\allowbreak{},58,59,6\allowbreak{}0)(61,62\allowbreak{},63,64,6\allowbreak{}5,66)(67\allowbreak{},68,69,7\allowbreak{}0,71,72\par}}}}\par\vspace{4pt}

\noindent\begin{minipage}{\linewidth}
\noindent{\color{rtFast}\rule[-0.1ex]{0.7em}{0.7em}}\hspace{0.45em}{\small\bfseries\color{rtFast!80!black}Category 2 (faster drop)}\hfill{\scriptsize\color{black!55}$n=426$}\par\nopagebreak\vspace{2pt}\nopagebreak

\noindent{\setlength{\fboxsep}{4pt}\colorbox{rtFast!7}{\parbox{\dimexpr\linewidth-2\fboxsep\relax}{\raggedright{\scriptsize\sffamily\color{black!65}loss \textbf{4.305}\quad drop \textbf{0.0391}\quad drop larger than for 98\% of sequences with a similar starting loss}\\[2pt]{\scriptsize\ttfamily\textcolor{black!45}{; hard ones only{\color{black!35}$\hookleftarrow$}\allowbreak{}  It is currently 16 Oct 2019, 17:24{\color{black!35}$\hookleftarrow$}\allowbreak{} \#\#\# GMAT Club}{\color{rtFast}\bfseries\,$\Vert$\,} Daily Prep{\color{black!35}$\hookleftarrow$}\allowbreak{} \#\#\#\# Thank you for using the timer -{} this advanced tool can estimate your performance and suggest more practice questions. We have subscribed you to Daily Prep Questions via email.{\color{black!35}$\hookleftarrow$}\allowbreak{} Customized{\color{black!35}$\hookleftarrow$}\allowbreak{} for You{\color{black!35}$\hookleftarrow$}\allowbreak{} we will pick new questions that match\par}}}}\par
\end{minipage}\par\vspace{4pt}

\noindent{\setlength{\fboxsep}{4pt}\colorbox{rtFast!7}{\parbox{\dimexpr\linewidth-2\fboxsep\relax}{\raggedright{\scriptsize\sffamily\color{black!65}loss \textbf{2.910}\quad drop \textbf{0.0176}\quad drop larger than for 93\% of sequences with a similar starting loss}\\[2pt]{\scriptsize\ttfamily\textcolor{black!45}{for You{\color{black!35}$\hookleftarrow$}\allowbreak{} we will pick new questions that match your level based on your Timer History{\color{black!35}$\hookleftarrow$}\allowbreak{} Track{\color{black!35}$\hookleftarrow$}\allowbreak{} }{\color{rtFast}\bfseries\,$\Vert$\,}every week, we’ll send you an estimated GMAT score based on your performance{\color{black!35}$\hookleftarrow$}\allowbreak{} Practice{\color{black!35}$\hookleftarrow$}\allowbreak{} Pays{\color{black!35}$\hookleftarrow$}\allowbreak{} we will pick new questions that match your level based on your Timer History{\color{black!35}$\hookleftarrow$}\allowbreak{} \# Events \& Promotions{\color{black!35}$\hookleftarrow$}\allowbreak{} \#\#\#\#\#\# Events \& Promotions\par}}}}\par\vspace{4pt}

\noindent{\setlength{\fboxsep}{4pt}\colorbox{rtFast!7}{\parbox{\dimexpr\linewidth-2\fboxsep\relax}{\raggedright{\scriptsize\sffamily\color{black!65}loss \textbf{0.259}\quad drop \textbf{0.0087}\quad drop larger than for 93\% of sequences with a similar starting loss}\\[2pt]{\scriptsize\ttfamily\textcolor{black!45}{),(174,4\allowbreak{}50,474,3\allowbreak{}95,285,4\allowbreak{}14),(175\allowbreak{},415,286\allowbreak{},396}{\color{rtFast}\bfseries\,$\Vert$\,},475,451\allowbreak{}),(176,4\allowbreak{}52,476,3\allowbreak{}97,287,4\allowbreak{}16),(177\allowbreak{},417,288\allowbreak{},398,477\allowbreak{},453),(1\allowbreak{}78,454,4\allowbreak{}78,399,2\allowbreak{}89,418),\allowbreak{}(179,419\allowbreak{},290,400\allowbreak{},479\par}}}}\par\vspace{4pt}

\noindent\begin{minipage}{\linewidth}
\noindent{\color{rtModerate}\rule[-0.1ex]{0.7em}{0.7em}}\hspace{0.45em}{\small\bfseries\color{rtModerate!80!black}Category 3 (moderate drop)}\hfill{\scriptsize\color{black!55}$n=113$}\par\nopagebreak\vspace{2pt}\nopagebreak

\noindent{\setlength{\fboxsep}{4pt}\colorbox{rtModerate!7}{\parbox{\dimexpr\linewidth-2\fboxsep\relax}{\raggedright{\scriptsize\sffamily\color{black!65}loss \textbf{3.727}\quad drop \textbf{0.0137}\quad drop larger than for 60\% of sequences with a similar starting loss}\\[2pt]{\scriptsize\ttfamily\textcolor{black!45}{{\color{black!35}$\hookleftarrow$}\allowbreak{} Thank you for using the timer -{} this advanced tool can estimate your performance and suggest more practice questions}{\color{rtModerate}\bfseries\,$\Vert$\,}. We have subscribed you to Daily Prep Questions via email.{\color{black!35}$\hookleftarrow$}\allowbreak{} Customized{\color{black!35}$\hookleftarrow$}\allowbreak{} for You{\color{black!35}$\hookleftarrow$}\allowbreak{} we will pick new questions that match your level based on your Timer History{\color{black!35}$\hookleftarrow$}\allowbreak{} Track{\color{black!35}$\hookleftarrow$}\allowbreak{} every week, we’ll send you an estimated GMAT score\par}}}}\par
\end{minipage}\par\vspace{4pt}

\noindent{\setlength{\fboxsep}{4pt}\colorbox{rtModerate!7}{\parbox{\dimexpr\linewidth-2\fboxsep\relax}{\raggedright{\scriptsize\sffamily\color{black!65}loss \textbf{0.407}\quad drop \textbf{0.0012}\quad drop larger than for 35\% of sequences with a similar starting loss}\\[2pt]{\scriptsize\ttfamily\textcolor{black!45}{40){\color{black!35}$\hookleftarrow$}\allowbreak{} Boot mode: Normal{\color{black!35}$\hookleftarrow$}\allowbreak{} Running processes:{\color{black!35}$\hookleftarrow$}\allowbreak{} C:\textbackslash{}WINDO\allowbreak{}WS\textbackslash{}syste\allowbreak{}m32\textbackslash{}}{\color{rtModerate}\bfseries\,$\Vert$\,}smss.exe{\color{black!35}$\hookleftarrow$}\allowbreak{} C:\textbackslash{}WINDO\allowbreak{}WS\textbackslash{}syste\allowbreak{}m32\textbackslash{}winl\allowbreak{}ogon.exe{\color{black!35}$\hookleftarrow$}\allowbreak{} C:\textbackslash{}WINDO\allowbreak{}WS\textbackslash{}syste\allowbreak{}m32\textbackslash{}serv\allowbreak{}ices.exe{\color{black!35}$\hookleftarrow$}\allowbreak{} C:\textbackslash{}WINDO\allowbreak{}WS\textbackslash{}syste\allowbreak{}m32\textbackslash{}lsas\allowbreak{}s.exe{\color{black!35}$\hookleftarrow$}\allowbreak{} C:\textbackslash{}WINDO\allowbreak{}WS\textbackslash{}syste\allowbreak{}m\par}}}}\par\vspace{4pt}

\noindent{\setlength{\fboxsep}{4pt}\colorbox{rtModerate!7}{\parbox{\dimexpr\linewidth-2\fboxsep\relax}{\raggedright{\scriptsize\sffamily\color{black!65}loss \textbf{0.126}\quad drop \textbf{0.0010}\quad drop larger than for 55\% of sequences with a similar starting loss}\\[2pt]{\scriptsize\ttfamily\textcolor{black!45}{,167,131\allowbreak{},153)(11\allowbreak{}8,168,13\allowbreak{}2,154)(1\allowbreak{}19,141,1\allowbreak{}33}{\color{rtModerate}\bfseries\,$\Vert$\,},155)(12\allowbreak{}0,142,13\allowbreak{}4,156)(1\allowbreak{}21,143,1\allowbreak{}35,157)(\allowbreak{}122,144,\allowbreak{}136,158)\allowbreak{}(123,145\allowbreak{},137,159\allowbreak{})(124,14\allowbreak{}6,138,16\allowbreak{}0)(125,1\allowbreak{}47,139,1\allowbreak{}61\par}}}}\par\vspace{4pt}

\noindent\begin{minipage}{\linewidth}
\noindent{\color{rtSlow}\rule[-0.1ex]{0.7em}{0.7em}}\hspace{0.45em}{\small\bfseries\color{rtSlow!80!black}Category 4 (slower drop)}\hfill{\scriptsize\color{black!55}$n=20$}\par\nopagebreak\vspace{2pt}\nopagebreak

\noindent{\setlength{\fboxsep}{4pt}\colorbox{rtSlow!7}{\parbox{\dimexpr\linewidth-2\fboxsep\relax}{\raggedright{\scriptsize\sffamily\color{black!65}loss \textbf{0.111}\quad drop \textbf{$-$0.0002}\quad drop larger than for 10\% of sequences with a similar starting loss}\\[2pt]{\scriptsize\ttfamily\textcolor{black!45}{ (2003) 10, No. 4 (2002) 10, No. 3 (2002) 10}{\color{rtSlow}\bfseries\,$\Vert$\,}, No. 2 (2002) 10, No. 1 (2002) 9, No. 4 (2001) 9, No. 3 (2001) 9, No. 2 (2001) 9, No. 1 (2001) 8, No\par}}}}\par
\end{minipage}\par\vspace{4pt}

\noindent{\setlength{\fboxsep}{4pt}\colorbox{rtSlow!7}{\parbox{\dimexpr\linewidth-2\fboxsep\relax}{\raggedright{\scriptsize\sffamily\color{black!65}loss \textbf{4.023}\quad drop \textbf{0.0078}\quad drop larger than for 15\% of sequences with a similar starting loss}\\[2pt]{\scriptsize\ttfamily\textcolor{black!45}{{\color{black!35}$\hookleftarrow$}\allowbreak{}  Geometry Search for the term Geometry in any field. Queries are case-{}ind\allowbreak{}ependent\allowbreak{}. Funct}{\color{rtSlow}\bfseries\,$\Vert$\,}* Wildcard queries are specified by * (e.g. functions, functorial, etc.). Otherwise the search is exact. \textquotedbl{}Topological group\textquotedbl{} Phrases (multi-{}w\allowbreak{}ords) should be set in \textquotedbl{}straight quotation marks\textquotedbl{}. au\par}}}}\par\vspace{4pt}

\noindent{\setlength{\fboxsep}{4pt}\colorbox{rtSlow!7}{\parbox{\dimexpr\linewidth-2\fboxsep\relax}{\raggedright{\scriptsize\sffamily\color{black!65}loss \textbf{3.932}\quad drop \textbf{0.0059}\quad drop larger than for 6\% of sequences with a similar starting loss}\\[2pt]{\scriptsize\ttfamily\textcolor{black!45}{ resource for mathematics{\color{black!35}$\hookleftarrow$}\allowbreak{} \#\#\#\#\# Examples{\color{black!35}$\hookleftarrow$}\allowbreak{}  Geometry Search for the term Geometry in any field. Queries are}{\color{rtSlow}\bfseries\,$\Vert$\,} case-{}ind\allowbreak{}ependent\allowbreak{}. Funct* Wildcard queries are specified by * (e.g. functions, functorial, etc.). Otherwise the search is exact. \textquotedbl{}Topological group\textquotedbl{} Phrases (multi-{}w\allowbreak{}ords) should be set in \textquotedbl{}\par}}}}\par\vspace{4pt}

\medskip\noindent{\large\bfseries FineWeb}\par\nopagebreak\smallskip\nopagebreak

\noindent\begin{minipage}{\linewidth}
\noindent{\color{rtLow}\rule[-0.1ex]{0.7em}{0.7em}}\hspace{0.45em}{\small\bfseries\color{rtLow!80!black}Category 1 (low starting loss)}\hfill{\scriptsize\color{black!55}$n=16$}\par\nopagebreak\vspace{2pt}\nopagebreak

\noindent{\setlength{\fboxsep}{4pt}\colorbox{rtLow!7}{\parbox{\dimexpr\linewidth-2\fboxsep\relax}{\raggedright{\scriptsize\sffamily\color{black!65}loss \textbf{0.080}\quad drop \textbf{0.0021}}\\[2pt]{\scriptsize\ttfamily\textcolor{black!45}{ website to function properly. This category only includes cookies that ensures basic function\allowbreak{}alities and security features of the}{\color{rtLow}\bfseries\,$\Vert$\,} website. These cookies do not store any personal information.{\color{black!35}$\hookleftarrow$}\allowbreak{} Any cookies that may not be particularly necessary for the website to function and is used specifically to collect user personal data via analytics, ads, other embedded contents are termed as non-{}nece\allowbreak{}ssary cookies.\par}}}}\par
\end{minipage}\par\vspace{4pt}

\noindent{\setlength{\fboxsep}{4pt}\colorbox{rtLow!7}{\parbox{\dimexpr\linewidth-2\fboxsep\relax}{\raggedright{\scriptsize\sffamily\color{black!65}loss \textbf{0.087}\quad drop \textbf{0.0022}}\\[2pt]{\scriptsize\ttfamily\textcolor{black!45}{ in its research publicat\allowbreak{}ions, and does not advise technology users to select only those vendors with the highest ratings}{\color{rtLow}\bfseries\,$\Vert$\,} or other designation. Gartner research publications consist of the opinions of Gartner’s research organization and should not be construed as statements of fact. Gartner disclaims all warranties, expressed or implied, with respect to this research,\par}}}}\par\vspace{4pt}

\noindent{\setlength{\fboxsep}{4pt}\colorbox{rtLow!7}{\parbox{\dimexpr\linewidth-2\fboxsep\relax}{\raggedright{\scriptsize\sffamily\color{black!65}loss \textbf{0.059}\quad drop \textbf{0.0020}}\\[2pt]{\scriptsize\ttfamily\textcolor{black!45}{ listing data appearing on this website, or contained in reports produced therefrom, is owned and copyrighted by California}{\color{rtLow}\bfseries\,$\Vert$\,} Regional Multiple Listing Service, Inc. (\textquotedbl{}CRMLS\textquotedbl{}) and is protected by all applicable copyright laws. Information provided is for viewer\textquotesingle{}s personal, non-{}comm\allowbreak{}ercial use and may not be used for any purpose other than to identify prospective properties the viewer\par}}}}\par\vspace{4pt}

\noindent\begin{minipage}{\linewidth}
\noindent{\color{rtFast}\rule[-0.1ex]{0.7em}{0.7em}}\hspace{0.45em}{\small\bfseries\color{rtFast!80!black}Category 2 (faster drop)}\hfill{\scriptsize\color{black!55}$n=27$}\par\nopagebreak\vspace{2pt}\nopagebreak

\noindent{\setlength{\fboxsep}{4pt}\colorbox{rtFast!7}{\parbox{\dimexpr\linewidth-2\fboxsep\relax}{\raggedright{\scriptsize\sffamily\color{black!65}loss \textbf{2.877}\quad drop \textbf{0.0371}\quad drop larger than for 100\% of sequences with a similar starting loss}\\[2pt]{\scriptsize\ttfamily\textcolor{black!45}{ maps, description for commercial real estate.{\color{black!35}$\hookleftarrow$}\allowbreak{} To support our service, we display Private Sponsored Links}{\color{rtFast}\bfseries\,$\Vert$\,} that are relevant to your search queries. These tracker-{}free affiliate links are not based on your personal information or browsing history, and they help us cover our costs without compromising your privacy. If you want to enjoy Ghostery without seeing sponsored results, you\par}}}}\par
\end{minipage}\par\vspace{4pt}

\noindent{\setlength{\fboxsep}{4pt}\colorbox{rtFast!7}{\parbox{\dimexpr\linewidth-2\fboxsep\relax}{\raggedright{\scriptsize\sffamily\color{black!65}loss \textbf{0.113}\quad drop \textbf{0.0059}\quad drop larger than for 86\% of sequences with a similar starting loss}\\[2pt]{\scriptsize\ttfamily\textcolor{black!45}{ especially linked to the most innovative technolo\allowbreak{}gies.{\color{black!35}$\hookleftarrow$}\allowbreak{} Privacy \& Cookies Policy{\color{black!35}$\hookleftarrow$}\allowbreak{} Necessary cookies are}{\color{rtFast}\bfseries\,$\Vert$\,} absolutely essential for the website to function properly. This category only includes cookies that ensures basic function\allowbreak{}alities and security features of the website. These cookies do not store any personal information.{\color{black!35}$\hookleftarrow$}\allowbreak{} Any cookies that may not be particularly necessary for the website to function and\par}}}}\par\vspace{4pt}

\noindent{\setlength{\fboxsep}{4pt}\colorbox{rtFast!7}{\parbox{\dimexpr\linewidth-2\fboxsep\relax}{\raggedright{\scriptsize\sffamily\color{black!65}loss \textbf{0.157}\quad drop \textbf{0.0072}\quad drop larger than for 81\% of sequences with a similar starting loss}\\[2pt]{\scriptsize\ttfamily\textcolor{black!45}{ ON{\color{black!35}$\hookleftarrow$}\allowbreak{} Stock \#PB2525N\allowbreak{}epean, ON{\color{black!35}$\hookleftarrow$}\allowbreak{} RV Liquidators is not responsible for}{\color{rtFast}\bfseries\,$\Vert$\,} any misprints, typos, or errors found in our website pages. Any price listed excludes sales tax, registration tags, and delivery fees. Manufacturer pictures, specific\allowbreak{}ations, and features may be used in place of actual units on our lot. Please\par}}}}\par\vspace{4pt}

\noindent\begin{minipage}{\linewidth}
\noindent{\color{rtModerate}\rule[-0.1ex]{0.7em}{0.7em}}\hspace{0.45em}{\small\bfseries\color{rtModerate!80!black}Category 3 (moderate drop)}\hfill{\scriptsize\color{black!55}$n=22$}\par\nopagebreak\vspace{2pt}\nopagebreak

\noindent{\setlength{\fboxsep}{4pt}\colorbox{rtModerate!7}{\parbox{\dimexpr\linewidth-2\fboxsep\relax}{\raggedright{\scriptsize\sffamily\color{black!65}loss \textbf{0.199}\quad drop \textbf{0.0067}\quad drop larger than for 65\% of sequences with a similar starting loss}\\[2pt]{\scriptsize\ttfamily\textcolor{black!45}{ highlights over the rest of the grass court season.{\color{black!35}$\hookleftarrow$}\allowbreak{} Privacy \& Cookies Policy{\color{black!35}$\hookleftarrow$}\allowbreak{} Necessary}{\color{rtModerate}\bfseries\,$\Vert$\,} cookies are absolutely essential for the website to function properly. This category only includes cookies that ensures basic function\allowbreak{}alities and security features of the website. These cookies do not store any personal information.{\color{black!35}$\hookleftarrow$}\allowbreak{} Any cookies that may not be particularly necessary for the website to\par}}}}\par
\end{minipage}\par\vspace{4pt}

\noindent{\setlength{\fboxsep}{4pt}\colorbox{rtModerate!7}{\parbox{\dimexpr\linewidth-2\fboxsep\relax}{\raggedright{\scriptsize\sffamily\color{black!65}loss \textbf{0.141}\quad drop \textbf{0.0032}\quad drop larger than for 42\% of sequences with a similar starting loss}\\[2pt]{\scriptsize\ttfamily\textcolor{black!45}{ of their friends and Brethren, or to fall themselves by their Hands. He has excited domestic ins}{\color{rtModerate}\bfseries\,$\Vert$\,}urrections amongst us, and has endeavoured to bring on the inhabitants of our frontiers, the merciless Indian Savages, whose known rule of warfare, is an undistin\allowbreak{}guished destruction of all ages, sexes and conditions.{\color{black!35}$\hookleftarrow$}\allowbreak{} In every stage\par}}}}\par\vspace{4pt}

\noindent{\setlength{\fboxsep}{4pt}\colorbox{rtModerate!7}{\parbox{\dimexpr\linewidth-2\fboxsep\relax}{\raggedright{\scriptsize\sffamily\color{black!65}loss \textbf{0.101}\quad drop \textbf{0.0029}\quad drop larger than for 53\% of sequences with a similar starting loss}\\[2pt]{\scriptsize\ttfamily\textcolor{black!45}{ary cookies are absolutely essential for the website to function properly. This category only includes cookies that ensures basic}{\color{rtModerate}\bfseries\,$\Vert$\,} function\allowbreak{}alities and security features of the website. These cookies do not store any personal information.{\color{black!35}$\hookleftarrow$}\allowbreak{} Any cookies that may not be particularly necessary for the website to function and is used specifically to collect user personal data via analytics, ads, other embedded contents are\par}}}}\par\vspace{4pt}

\noindent\begin{minipage}{\linewidth}
\noindent{\color{rtSlow}\rule[-0.1ex]{0.7em}{0.7em}}\hspace{0.45em}{\small\bfseries\color{rtSlow!80!black}Category 4 (slower drop)}\hfill{\scriptsize\color{black!55}$n=5$}\par\nopagebreak\vspace{2pt}\nopagebreak

\noindent{\setlength{\fboxsep}{4pt}\colorbox{rtSlow!7}{\parbox{\dimexpr\linewidth-2\fboxsep\relax}{\raggedright{\scriptsize\sffamily\color{black!65}loss \textbf{0.109}\quad drop \textbf{0.0009}\quad drop larger than for 11\% of sequences with a similar starting loss}\\[2pt]{\scriptsize\ttfamily\textcolor{black!45}{ary cookies are absolutely essential for the website to function properly. This category only includes cookies that ensures basic}{\color{rtSlow}\bfseries\,$\Vert$\,} function\allowbreak{}alities and security features of the website. These cookies do not store any personal information.{\color{black!35}$\hookleftarrow$}\allowbreak{} Any cookies that may not be particularly necessary for the website to function and is used specifically to collect user personal data via analytics, ads, other embedded contents are\par}}}}\par
\end{minipage}\par\vspace{4pt}

\noindent{\setlength{\fboxsep}{4pt}\colorbox{rtSlow!7}{\parbox{\dimexpr\linewidth-2\fboxsep\relax}{\raggedright{\scriptsize\sffamily\color{black!65}loss \textbf{0.127}\quad drop \textbf{0.0010}\quad drop larger than for 17\% of sequences with a similar starting loss}\\[2pt]{\scriptsize\ttfamily\textcolor{black!45}{iah.{\color{black!35}$\hookleftarrow$}\allowbreak{} Zechariah 12:10 And I will pour upon the house of David, and}{\color{rtSlow}\bfseries\,$\Vert$\,} upon the inhabitants of Jerusalem, the spirit of grace and of supplica\allowbreak{}tions: and they shall look upon me whom they have pierced, and they shall mourn for him, as one mourneth for his only son, and shall be in bitterness for him\par}}}}\par\vspace{4pt}

\noindent{\setlength{\fboxsep}{4pt}\colorbox{rtSlow!7}{\parbox{\dimexpr\linewidth-2\fboxsep\relax}{\raggedright{\scriptsize\sffamily\color{black!65}loss \textbf{0.175}\quad drop \textbf{0.0021}\quad drop larger than for 23\% of sequences with a similar starting loss}\\[2pt]{\scriptsize\ttfamily\textcolor{black!45}{ God he created them; male and female he created them. 28God blessed them, and God said}{\color{rtSlow}\bfseries\,$\Vert$\,} to them, “Be fruitful and multiply, and fill the earth and subdue it; and have dominion over the fish of the sea and over the birds of the air and over every living thing that moves upon the earth.” 29God said,\par}}}}\par\vspace{4pt}

\medskip\noindent{\large\bfseries PG-19}\par\nopagebreak\smallskip\nopagebreak

\noindent\begin{minipage}{\linewidth}
\noindent{\color{rtLow}\rule[-0.1ex]{0.7em}{0.7em}}\hspace{0.45em}{\small\bfseries\color{rtLow!80!black}Category 1 (low starting loss)}\hfill{\scriptsize\color{black!55}$n=6$}\par\nopagebreak\vspace{2pt}\nopagebreak

\noindent{\setlength{\fboxsep}{4pt}\colorbox{rtLow!7}{\parbox{\dimexpr\linewidth-2\fboxsep\relax}{\raggedright{\scriptsize\sffamily\color{black!65}loss \textbf{0.090}\quad drop \textbf{0.0089}}\\[2pt]{\scriptsize\ttfamily\textcolor{black!45}{ market{\color{black!35}$\hookleftarrow$}\allowbreak{} any commercial products without permission.{\color{black!35}$\hookleftarrow$}\allowbreak{} To create these etexts, the Project expends}{\color{rtLow}\bfseries\,$\Vert$\,} considerable{\color{black!35}$\hookleftarrow$}\allowbreak{} efforts to identify, transcribe and proofread public domain{\color{black!35}$\hookleftarrow$}\allowbreak{} works. Despite these efforts, the Project\textquotesingle{}s etexts and any{\color{black!35}$\hookleftarrow$}\allowbreak{} medium they may be on may contain \textquotedbl{}Defects\textquotedbl{}. Among other{\color{black!35}$\hookleftarrow$}\allowbreak{} things, De\par}}}}\par
\end{minipage}\par\vspace{4pt}

\noindent{\setlength{\fboxsep}{4pt}\colorbox{rtLow!7}{\parbox{\dimexpr\linewidth-2\fboxsep\relax}{\raggedright{\scriptsize\sffamily\color{black!65}loss \textbf{0.065}\quad drop \textbf{$-$0.0013}}\\[2pt]{\scriptsize\ttfamily\textcolor{black!45}{Book Sixty-{}five{\color{black!35}$\hookleftarrow$}\allowbreak{} Book Sixty-{}six{\color{black!35}$\hookleftarrow$}\allowbreak{} Book Sixty-{}seven{\color{black!35}$\hookleftarrow$}\allowbreak{} Book Sixty-{}eight{\color{black!35}$\hookleftarrow$}\allowbreak{} }{\color{rtLow}\bfseries\,$\Vert$\,}Book Sixty-{}nine{\color{black!35}$\hookleftarrow$}\allowbreak{} Book Seventy{\color{black!35}$\hookleftarrow$}\allowbreak{} Book Seventy-{}one{\color{black!35}$\hookleftarrow$}\allowbreak{} Book Seventy-{}two{\color{black!35}$\hookleftarrow$}\allowbreak{} Book Seventy-{}\allowbreak{}three{\color{black!35}$\hookleftarrow$}\allowbreak{} Book Seventy-{}four{\color{black!35}$\hookleftarrow$}\allowbreak{} Book Seventy-{}five{\color{black!35}$\hookleftarrow$}\allowbreak{} Book Seventy-{}six{\color{black!35}$\hookleftarrow$}\allowbreak{} Book Seventy-{}\allowbreak{}seven\par}}}}\par\vspace{4pt}

\noindent{\setlength{\fboxsep}{4pt}\colorbox{rtLow!7}{\parbox{\dimexpr\linewidth-2\fboxsep\relax}{\raggedright{\scriptsize\sffamily\color{black!65}loss \textbf{0.088}\quad drop \textbf{0.0051}}\\[2pt]{\scriptsize\ttfamily\textcolor{black!45}{pton Sinclair{\color{black!35}$\hookleftarrow$}\allowbreak{} Copyright laws are changing all over the world, be sure to check{\color{black!35}$\hookleftarrow$}\allowbreak{} the copyright laws}{\color{rtLow}\bfseries\,$\Vert$\,} for your country before posting these files!!{\color{black!35}$\hookleftarrow$}\allowbreak{} Please take a look at the important information in this header.{\color{black!35}$\hookleftarrow$}\allowbreak{} We encourage you to keep this file on your own disk, keeping an{\color{black!35}$\hookleftarrow$}\allowbreak{} electronic path open for the next readers. Do not remove this\par}}}}\par\vspace{4pt}

\noindent\begin{minipage}{\linewidth}
\noindent{\color{rtFast}\rule[-0.1ex]{0.7em}{0.7em}}\hspace{0.45em}{\small\bfseries\color{rtFast!80!black}Category 2 (faster drop)}\hfill{\scriptsize\color{black!55}$n=24$}\par\nopagebreak\vspace{2pt}\nopagebreak

\noindent{\setlength{\fboxsep}{4pt}\colorbox{rtFast!7}{\parbox{\dimexpr\linewidth-2\fboxsep\relax}{\raggedright{\scriptsize\sffamily\color{black!65}loss \textbf{0.147}\quad drop \textbf{0.0109}\quad drop larger than for 92\% of sequences with a similar starting loss}\\[2pt]{\scriptsize\ttfamily\textcolor{black!45}{ in our series by Plato{\color{black!35}$\hookleftarrow$}\allowbreak{} Copyright laws are changing all over the world, be sure to check{\color{black!35}$\hookleftarrow$}\allowbreak{} }{\color{rtFast}\bfseries\,$\Vert$\,}the copyright laws for your country before posting these files!!{\color{black!35}$\hookleftarrow$}\allowbreak{} Please take a look at the important information in this header.{\color{black!35}$\hookleftarrow$}\allowbreak{} We encourage you to keep this file on your own disk, keeping an{\color{black!35}$\hookleftarrow$}\allowbreak{} electronic path open for the next readers. Do\par}}}}\par
\end{minipage}\par\vspace{4pt}

\noindent{\setlength{\fboxsep}{4pt}\colorbox{rtFast!7}{\parbox{\dimexpr\linewidth-2\fboxsep\relax}{\raggedright{\scriptsize\sffamily\color{black!65}loss \textbf{0.138}\quad drop \textbf{0.0092}\quad drop larger than for 78\% of sequences with a similar starting loss}\\[2pt]{\scriptsize\ttfamily\textcolor{black!45}{Copyright laws are changing all over the world, be sure to check{\color{black!35}$\hookleftarrow$}\allowbreak{} the copyright laws for your country}{\color{rtFast}\bfseries\,$\Vert$\,} before posting these files!!{\color{black!35}$\hookleftarrow$}\allowbreak{} Please take a look at the important information in this header.{\color{black!35}$\hookleftarrow$}\allowbreak{} We encourage you to keep this file on your own disk, keeping an{\color{black!35}$\hookleftarrow$}\allowbreak{} electronic path open for the next readers. Do not remove this.{\color{black!35}$\hookleftarrow$}\allowbreak{} **\par}}}}\par\vspace{4pt}

\noindent{\setlength{\fboxsep}{4pt}\colorbox{rtFast!7}{\parbox{\dimexpr\linewidth-2\fboxsep\relax}{\raggedright{\scriptsize\sffamily\color{black!65}loss \textbf{0.127}\quad drop \textbf{0.0094}\quad drop larger than for 84\% of sequences with a similar starting loss}\\[2pt]{\scriptsize\ttfamily\textcolor{black!45}{Copyright laws are changing all over the world, be sure to check{\color{black!35}$\hookleftarrow$}\allowbreak{} the copyright laws for your country}{\color{rtFast}\bfseries\,$\Vert$\,} before posting these files!!{\color{black!35}$\hookleftarrow$}\allowbreak{} Please take a look at the important information in this header.{\color{black!35}$\hookleftarrow$}\allowbreak{} We encourage you to keep this file on your own disk, keeping an{\color{black!35}$\hookleftarrow$}\allowbreak{} electronic path open for the next readers. Do not remove this.{\color{black!35}$\hookleftarrow$}\allowbreak{} **\par}}}}\par\vspace{4pt}

\noindent\begin{minipage}{\linewidth}
\noindent{\color{rtModerate}\rule[-0.1ex]{0.7em}{0.7em}}\hspace{0.45em}{\small\bfseries\color{rtModerate!80!black}Category 3 (moderate drop)}\hfill{\scriptsize\color{black!55}$n=11$}\par\nopagebreak\vspace{2pt}\nopagebreak

\noindent{\setlength{\fboxsep}{4pt}\colorbox{rtModerate!7}{\parbox{\dimexpr\linewidth-2\fboxsep\relax}{\raggedright{\scriptsize\sffamily\color{black!65}loss \textbf{0.334}\quad drop \textbf{0.0078}\quad drop larger than for 47\% of sequences with a similar starting loss}\\[2pt]{\scriptsize\ttfamily\textcolor{black!45}{.org/gut\allowbreak{}enberg/1\allowbreak{}/1/4/0/1\allowbreak{}1404/114\allowbreak{}04}{\color{rtModerate}\bfseries\,$\Vert$\,}-{}h/11404\allowbreak{}-{}h.htm){\color{black!35}$\hookleftarrow$}\allowbreak{}  or{\color{black!35}$\hookleftarrow$}\allowbreak{}  (http://\allowbreak{}www.ibib\allowbreak{}lio.org/\allowbreak{}gutenber\allowbreak{}g/1/1/4/\allowbreak{}0/11404/\allowbreak{}11404-{}h.\allowbreak{}zip){\color{black!35}$\hookleftarrow$}\allowbreak{} THE\par}}}}\par
\end{minipage}\par\vspace{4pt}

\noindent{\setlength{\fboxsep}{4pt}\colorbox{rtModerate!7}{\parbox{\dimexpr\linewidth-2\fboxsep\relax}{\raggedright{\scriptsize\sffamily\color{black!65}loss \textbf{0.111}\quad drop \textbf{0.0030}\quad drop larger than for 38\% of sequences with a similar starting loss}\\[2pt]{\scriptsize\ttfamily\textcolor{black!45}{www.ibib\allowbreak{}lio.org/\allowbreak{}gutenber\allowbreak{}g/1/1/3/\allowbreak{}8/}{\color{rtModerate}\bfseries\,$\Vert$\,}11384/11\allowbreak{}384-{}h/11\allowbreak{}384-{}h.ht\allowbreak{}m){\color{black!35}$\hookleftarrow$}\allowbreak{}  or{\color{black!35}$\hookleftarrow$}\allowbreak{}  (http://\allowbreak{}www.ibib\allowbreak{}lio.org/\allowbreak{}gutenber\allowbreak{}g/1/1/3/\allowbreak{}8/11384/\allowbreak{}11384-{}h\par}}}}\par\vspace{4pt}

\noindent{\setlength{\fboxsep}{4pt}\colorbox{rtModerate!7}{\parbox{\dimexpr\linewidth-2\fboxsep\relax}{\raggedright{\scriptsize\sffamily\color{black!65}loss \textbf{0.438}\quad drop \textbf{0.0115}\quad drop larger than for 67\% of sequences with a similar starting loss}\\[2pt]{\scriptsize\ttfamily\textcolor{black!45}{ Rome as \textquotedbl{}the Babylonish harlot.\textquotedbl{}{\color{black!35}$\hookleftarrow$}\allowbreak{}  7. And when they shall have finished their}{\color{rtModerate}\bfseries\,$\Vert$\,} testimony, the beast{\color{black!35}$\hookleftarrow$}\allowbreak{}  that ascendeth out of the bottomless pit shall make war against{\color{black!35}$\hookleftarrow$}\allowbreak{}  them, and shall overcome them, and kill them.{\color{black!35}$\hookleftarrow$}\allowbreak{}  8. And their dead bodies shall lie in the street of the great{\color{black!35}$\hookleftarrow$}\allowbreak{} \par}}}}\par\vspace{4pt}

\noindent\begin{minipage}{\linewidth}
\noindent{\color{rtSlow}\rule[-0.1ex]{0.7em}{0.7em}}\hspace{0.45em}{\small\bfseries\color{rtSlow!80!black}Category 4 (slower drop)}\hfill{\scriptsize\color{black!55}$n=2$}\par\nopagebreak\vspace{2pt}\nopagebreak

\noindent{\setlength{\fboxsep}{4pt}\colorbox{rtSlow!7}{\parbox{\dimexpr\linewidth-2\fboxsep\relax}{\raggedright{\scriptsize\sffamily\color{black!65}loss \textbf{0.111}\quad drop \textbf{0.0016}\quad drop larger than for 12\% of sequences with a similar starting loss}\\[2pt]{\scriptsize\ttfamily\textcolor{black!45}{ I with Confed. Art. XII.]{\color{black!35}$\hookleftarrow$}\allowbreak{} 2. This Constitu\allowbreak{}tion, and the laws of the United}{\color{rtSlow}\bfseries\,$\Vert$\,} States which shall be{\color{black!35}$\hookleftarrow$}\allowbreak{} made in pursuance thereof, and all treaties made, or which shall be{\color{black!35}$\hookleftarrow$}\allowbreak{} made, under the authority of the United States, shall be the supreme{\color{black!35}$\hookleftarrow$}\allowbreak{} law of the land; and the judges in every State shall be bound\par}}}}\par
\end{minipage}\par\vspace{4pt}

\noindent{\setlength{\fboxsep}{4pt}\colorbox{rtSlow!7}{\parbox{\dimexpr\linewidth-2\fboxsep\relax}{\raggedright{\scriptsize\sffamily\color{black!65}loss \textbf{0.130}\quad drop \textbf{0.0018}\quad drop larger than for 19\% of sequences with a similar starting loss}\\[2pt]{\scriptsize\ttfamily\textcolor{black!45}{ version of this{\color{black!35}$\hookleftarrow$}\allowbreak{}  file which includes the original illustra\allowbreak{}tions.{\color{black!35}$\hookleftarrow$}\allowbreak{}  See 11076-{}h.}{\color{rtSlow}\bfseries\,$\Vert$\,}htm or 11076-{}h.zip:{\color{black!35}$\hookleftarrow$}\allowbreak{}  (http://\allowbreak{}www.ibib\allowbreak{}lio.org/\allowbreak{}gutenber\allowbreak{}g/1/1/0/\allowbreak{}7/11076/\allowbreak{}11076-{}h/\allowbreak{}11076-{}h.\allowbreak{}htm){\color{black!35}$\hookleftarrow$}\allowbreak{} \par}}}}\par\vspace{4pt}

\subsection{Computing Fine-Tuning Data Overlap}
\label{app:ft_overlap_details}
For each sequence, we take the 8-grams in its 50-token suffix and count how many times each of them occurs elsewhere in the fine-tuning data. We add these counts up to get the overlap of the sequence. Each memorized sequence is then compared with non-memorized sequences from the same category whose starting loss and one-step loss drop are both within about 10\% of its own. For \low{Category 1}, we only match on starting loss. If fewer than five such non-memorized sequences exist, we instead use the five closest ones from the same category. We report the median count over memorized sequences and over their matched non-memorized sequences, where each memorized sequence counts equally. Most memorized sequences needed the five closest ones only in some FineWeb and PG-19 cells, and in \low{Category 1} of Enron for Qwen-2.5 1.5B and Llama-3.2 1B. FineWeb and PG-19 have only 2--27 memorized sequences per cell for Pythia-1.4B.

\subsection{Overlap with the Fine-Tuning Data}
\label{app:ft_overlap}
Table~\ref{tab:ft_overlap_qwenllama} shows the fine-tuning overlap from Table~\ref{tab:ft_overlap} for Qwen-2.5 1.5B and Llama-3.2 1B. As with Pythia-1.4B, memorized sequences in \fast{Categories 2}, \moderate{3}, and \slow{4} have a much higher overlap than matched non-memorized ones in The Stack, OpenWebMath, Enron, and WildChat. FineWeb and PG-19 have few memorized sequences. Table~\ref{tab:ft32} shows results for 32-gram overlap for each 50-token suffix with the rest of the fine-tuning data on Enron.

\begin{table}[h]
\centering\small
\setlength{\tabcolsep}{3pt}
\resizebox{\columnwidth}{!}{\begin{tabular}{@{}lrrrr@{}}
\toprule
Dataset & \low{Cat. 1} & \fast{Cat. 2} & \moderate{Cat. 3} & \slow{Cat. 4} \\
\midrule
\multicolumn{5}{@{}l}{\textit{Qwen-2.5 1.5B}} \\
The Stack   & 1.7k\hspace{0.1em}/\hspace{0.1em}604 & 4.1k\hspace{0.1em}/\hspace{0.1em}461 & 10.7k\hspace{0.1em}/\hspace{0.1em}197 & 19.7k\hspace{0.1em}/\hspace{0.1em}284 \\
OpenWebMath & 620\hspace{0.1em}/\hspace{0.1em}480 & 955\hspace{0.1em}/\hspace{0.1em}245 & 1.6k\hspace{0.1em}/\hspace{0.1em}74 & 12.0k\hspace{0.1em}/\hspace{0.1em}91 \\
FineWeb     & 157\hspace{0.1em}/\hspace{0.1em}16 & 158\hspace{0.1em}/\hspace{0.1em}10 & 12\hspace{0.1em}/\hspace{0.1em}13 & 8\hspace{0.1em}/\hspace{0.1em}6 \\
PG-19       & 0\hspace{0.1em}/\hspace{0.1em}0 & 1.2k\hspace{0.1em}/\hspace{0.1em}5 & 2\hspace{0.1em}/\hspace{0.1em}10 & 0\hspace{0.1em}/\hspace{0.1em}0 \\
Enron       & 15.9k\hspace{0.1em}/\hspace{0.1em}22.1k & 39.0k\hspace{0.1em}/\hspace{0.1em}27.6k & 11.6k\hspace{0.1em}/\hspace{0.1em}1.2k & 15.2k\hspace{0.1em}/\hspace{0.1em}1.3k \\
WildChat    & 308\hspace{0.1em}/\hspace{0.1em}144 & 3.9k\hspace{0.1em}/\hspace{0.1em}53 & 3.5k\hspace{0.1em}/\hspace{0.1em}50 & 3.4k\hspace{0.1em}/\hspace{0.1em}76 \\
\midrule
\multicolumn{5}{@{}l}{\textit{Llama-3.2 1B}} \\
The Stack   & 1.4k\hspace{0.1em}/\hspace{0.1em}978 & 5.7k\hspace{0.1em}/\hspace{0.1em}1.6k & 16.2k\hspace{0.1em}/\hspace{0.1em}119 & 3.3k\hspace{0.1em}/\hspace{0.1em}55 \\
OpenWebMath & 738\hspace{0.1em}/\hspace{0.1em}161 & 1.0k\hspace{0.1em}/\hspace{0.1em}210 & 1.7k\hspace{0.1em}/\hspace{0.1em}108 & 480\hspace{0.1em}/\hspace{0.1em}101 \\
FineWeb     & 8\hspace{0.1em}/\hspace{0.1em}9 & 36\hspace{0.1em}/\hspace{0.1em}2 & 10\hspace{0.1em}/\hspace{0.1em}4 & 45\hspace{0.1em}/\hspace{0.1em}19 \\
PG-19       & 4\hspace{0.1em}/\hspace{0.1em}10 & 66\hspace{0.1em}/\hspace{0.1em}10 & 42\hspace{0.1em}/\hspace{0.1em}36 & 40\hspace{0.1em}/\hspace{0.1em}21 \\
Enron       & 9.6k\hspace{0.1em}/\hspace{0.1em}618 & 7.9k\hspace{0.1em}/\hspace{0.1em}858 & 8.0k\hspace{0.1em}/\hspace{0.1em}277 & 5.4k\hspace{0.1em}/\hspace{0.1em}86 \\
WildChat    & 385\hspace{0.1em}/\hspace{0.1em}193 & 1.8k\hspace{0.1em}/\hspace{0.1em}59 & 2.6k\hspace{0.1em}/\hspace{0.1em}35 & 2.7k\hspace{0.1em}/\hspace{0.1em}29 \\
\bottomrule
\end{tabular}}
\caption{Overlap with the fine-tuning data (8-grams) for Qwen-2.5 1.5B and Llama-3.2 1B, shown as memorized\hspace{0.1em}/\hspace{0.1em}non-memorized median count.}
\label{tab:ft_overlap_qwenllama}
\end{table}

\begin{table}[h]
\centering\small
\setlength{\tabcolsep}{3pt}
\resizebox{\columnwidth}{!}{\begin{tabular}{@{}lrrrr@{}}
\toprule
Model & \low{Cat. 1} & \fast{Cat. 2} & \moderate{Cat. 3} & \slow{Cat. 4} \\
\midrule
Pythia-1.4B   & 1.1k\hspace{0.1em}/\hspace{0.1em}721 & 426\hspace{0.1em}/\hspace{0.1em}109 & 324\hspace{0.1em}/\hspace{0.1em}87 & 414\hspace{0.1em}/\hspace{0.1em}82 \\
Qwen-2.5 1.5B & 251\hspace{0.1em}/\hspace{0.1em}2 & 4.5k\hspace{0.1em}/\hspace{0.1em}27 & 1.0k\hspace{0.1em}/\hspace{0.1em}12 & 1.0k\hspace{0.1em}/\hspace{0.1em}11 \\
Llama-3.2 1B  & 10\hspace{0.1em}/\hspace{0.1em}38 & 880\hspace{0.1em}/\hspace{0.1em}14 & 703\hspace{0.1em}/\hspace{0.1em}13 & 722\hspace{0.1em}/\hspace{0.1em}10 \\
\bottomrule
\end{tabular}}
\caption{Overlap with the fine-tuning data on Enron using 32-grams, shown as memorized\hspace{0.1em}/\hspace{0.1em}non-memorized median count. For each sequence, we count how many times the 32-grams in its 50-token suffix occur elsewhere in the fine-tuning data.}
\label{tab:ft32}
\end{table}

\section{More Details on Scaling Laws}
\label{app:grp_scaling}

\subsection{Fitting the Memorization Scaling Laws}
\label{app:scaling_fit}
In this section, we share more details on our scaling laws experiments in Section~\ref{sec:scaling_laws}. We fit each memorization-rate power law ($m=c\,N^{b}$) by ordinary least squares in log-log space, consistent with prior scaling-law fits~\citep{rosenfeld2019constructivepredictiongeneralizationerror}. This weights each model size by relative rather than absolute error. Taking logarithms gives $\log_{10}\text{mem} = \log_{10}c + b\log_{10}N$. The fit uses the five per-size mean memorization rates ($N\in\{0.5,1.5,3,7,14\}$B).

We report a $95\%$ confidence interval on $b$ as the standard $t$-interval for a least-squares slope obtained from the residuals of that single fit. With $x_i=\log_{10}N_i$, residuals $r_i$, and $n=5$ points, the standard error of the slope is
\begin{equation}
\mathrm{SE}(b)=\sqrt{\frac{\tfrac{1}{n-2}\sum_i r_i^2}{\sum_i (x_i-\bar{x})^2}},
\end{equation}
and the interval is then $b \pm t\,\mathrm{SE}(b)$ with $t \approx 3.18$, the standard value for a 95\% interval when fitting a line to 5 points. For example, on The Stack $b=0.43$ and $\mathrm{SE}(b)=0.09$, giving $0.43\pm3.18\times0.09=[0.15,0.72]$. The intervals are wide because we only use $n=5$ model sizes; additional model sizes would give tighter intervals. 

\subsection{Predictive Power Across Qwen-2.5 Ladder}
\label{app:ladder_auc}

We report the three classifier performance metrics used throughout the paper for every Qwen-2.5 model size on the three scaling datasets. Table~\ref{tab:ladder_recallk} gives the cross-validation ROC-AUC (all features) and the held-out recall@$k$, and Table~\ref{tab:ladder_precisionk} the held-out precision@$k$. For the 0.5B, 1.5B, and 3B sizes we report the mean $\pm$ standard deviation over three fine-tuning seeds, and for 7B and 14B a single run. The classifier separates memorized from non-memorized sequences well at every size.

\begin{table*}[t]
\centering\small
\begin{tabular}{llcccccc}
\toprule
 & & & \multicolumn{5}{c}{Recall@$k$ (\%)} \\
\cmidrule(lr){4-8}
Dataset & Size & ROC-AUC & $0.1\%$ & $0.5\%$ & $1\%$ & $5\%$ & $10\%$ \\
\midrule
The Stack  & 0.5B  & $0.969\pm0.000$ & $3.4\pm0.0$ & $16.5\pm0.1$ & $29.5\pm0.2$ & $71.1\pm0.1$ & $81.5\pm0.3$ \\
           & 1.5B  & $0.973\pm0.000$ & $3.5\pm0.0$ & $16.2\pm0.2$ & $29.2\pm0.2$ & $75.5\pm0.4$ & $85.7\pm0.3$ \\
           & 3B    & $0.968\pm0.000$ & $2.7\pm0.1$ & $12.7\pm0.1$ & $23.5\pm0.1$ & $70.2\pm0.2$ & $84.6\pm0.2$ \\
           & 7B    & 0.948 & 1.8 & 8.2 & 15.4 & 54.8 & 75.1 \\
           & 14B   & 0.909 & 1.0 & 4.6 & 8.8 & 36.1 & 57.7 \\
\midrule
WildChat   & 0.5B  & $0.877\pm0.001$ & $1.4\pm0.0$ & $5.0\pm0.1$ & $8.1\pm0.1$ & $22.4\pm0.2$ & $34.3\pm0.2$ \\
           & 1.5B  & $0.892\pm0.001$ & $1.3\pm0.0$ & $5.5\pm0.0$ & $9.1\pm0.1$ & $24.9\pm0.1$ & $37.3\pm0.2$ \\
           & 3B    & $0.882\pm0.002$ & $1.2\pm0.0$ & $5.3\pm0.0$ & $9.1\pm0.0$ & $24.3\pm0.1$ & $36.4\pm0.1$ \\
           & 7B    & 0.847 & 0.8 & 3.6 & 6.7 & 21.2 & 33.1 \\
           & 14B   & 0.809 & 0.6 & 2.5 & 4.7 & 17.3 & 28.7 \\
\midrule
FineWeb    & 0.5B  & $0.995\pm0.008$ & $80.5\pm0.4$ & $98.7\pm1.2$ & $99.0\pm1.0$ & $100.0\pm0.0$ & $100.0\pm0.0$ \\
           & 1.5B  & $0.998\pm0.002$ & $57.9\pm3.9$ & $95.2\pm2.9$ & $98.1\pm1.1$ & $99.1\pm0.4$ & $99.6\pm0.0$ \\
           & 3B    & $0.997\pm0.000$ & $60.8\pm1.0$ & $95.1\pm0.6$ & $98.8\pm0.3$ & $99.8\pm0.0$ & $99.8\pm0.0$ \\
           & 7B    & 0.992 & 37.8 & 82.2 & 85.7 & 87.9 & 88.4 \\
           & 14B   & 0.988 & 18.8 & 64.5 & 80.9 & 90.2 & 92.3 \\
\bottomrule
\end{tabular}
\caption{Cross-validation ROC-AUC (all features) and held-out recall@$k$ (\%) at audit budgets $k$ across the Qwen-2.5 ladder on three datasets. Same setup as Table~\ref{tab:recallk}. Values for 0.5B, 1.5B, and 3B are mean $\pm$ standard deviation over three fine-tuning seeds, and 7B and 14B are single runs.}
\label{tab:ladder_recallk}\label{tab:ladder_auc}
\end{table*}

\begin{table*}[t]
\centering\small
\begin{tabular}{llccccc}
\toprule
Dataset & Size & $0.1\%$ & $0.5\%$ & $1\%$ & $5\%$ & $10\%$ \\
\midrule
The Stack  & 0.5B  & $60.2\pm1.0$ & $59.1\pm0.6$ & $52.7\pm0.4$ & $25.4\pm0.4$ & $14.6\pm0.3$ \\
           & 1.5B  & $69.6\pm0.9$ & $64.5\pm0.8$ & $58.1\pm0.3$ & $30.0\pm0.2$ & $17.0\pm0.1$ \\
           & 3B    & $73.6\pm1.4$ & $68.8\pm0.5$ & $63.6\pm0.1$ & $38.1\pm0.1$ & $23.0\pm0.0$ \\
           & 7B    & 80.3 & 71.5 & 67.1 & 47.9 & 32.8 \\
           & 14B   & 79.3 & 74.0 & 71.0 & 58.3 & 46.6 \\
\midrule
WildChat   & 0.5B  & $59.8\pm0.2$ & $42.3\pm0.3$ & $34.3\pm0.4$ & $18.9\pm0.3$ & $14.5\pm0.3$ \\
           & 1.5B  & $58.9\pm1.9$ & $50.5\pm0.1$ & $41.9\pm0.3$ & $23.0\pm0.0$ & $17.2\pm0.1$ \\
           & 3B    & $69.7\pm1.8$ & $62.4\pm0.1$ & $53.8\pm0.1$ & $28.6\pm0.0$ & $21.4\pm0.0$ \\
           & 7B    & 80.9 & 74.1 & 68.5 & 43.2 & 33.7 \\
           & 14B   & 83.8 & 76.3 & 72.0 & 52.7 & 43.8 \\
\midrule
FineWeb    & 0.5B  & $8.0\pm0.5$ & $2.0\pm0.1$ & $1.0\pm0.0$ & $0.2\pm0.0$ & $0.1\pm0.0$ \\
           & 1.5B  & $15.0\pm1.3$ & $4.9\pm0.2$ & $2.5\pm0.1$ & $0.5\pm0.0$ & $0.3\pm0.0$ \\
           & 3B    & $29.0\pm1.2$ & $9.1\pm0.2$ & $4.7\pm0.1$ & $1.0\pm0.0$ & $0.5\pm0.0$ \\
           & 7B    & 32.7 & 14.2 & 7.4 & 1.5 & 0.8 \\
           & 14B   & 33.1 & 22.7 & 14.3 & 3.2 & 1.6 \\
\bottomrule
\end{tabular}
\caption{Held-out precision@$k$ (\%) across the Qwen-2.5 ladder on three datasets at audit budgets $k$. Same setup as Table~\ref{tab:recallk}. Values for 0.5B, 1.5B, and 3B are mean $\pm$ standard deviation over three fine-tuning seeds, and 7B and 14B are single runs.}
\label{tab:ladder_precisionk}
\end{table*}

\subsection{Predictive-Power Exponents Across Audit Budgets}
\label{app:budget_sweep}

Table~\ref{tab:budget_sweep} reports the recall@$k$ and precision@$k$ exponents as the audit budget $k$ changes. We fit them over the five Qwen-2.5 model sizes in the same way as in Section~\ref{sec:scaling_laws}.

\begin{table*}[t]
\centering\small
\begin{tabular}{lcccccc}
\toprule
 & \multicolumn{3}{c}{recall@$k$ exponent $b$} & \multicolumn{3}{c}{precision@$k$ exponent $b$} \\
\cmidrule(lr){2-4}\cmidrule(lr){5-7}
Budget $k$ & The Stack & WildChat & FineWeb & The Stack & WildChat & FineWeb \\
\midrule
0.1\% & $-0.35$ & $-0.27$ & $-0.39$ & $+0.08$ & $+0.12$ & $+0.44$ \\
0.5\% & $-0.37$ & $-0.20$ & $-0.11$ & $+0.07$ & $+0.19$ & $+0.71$ \\
1\%   & $-0.35$ & $-0.16$ & $-0.06$ & $+0.09$ & $+0.23$ & $+0.77$ \\
5\%   & $-0.19$ & $-0.07$ & $-0.04$ & $+0.25$ & $+0.32$ & $+0.79$ \\
10\%  & $-0.09$ & $-0.05$ & $-0.03$ & $+0.35$ & $+0.34$ & $+0.79$ \\
\bottomrule
\end{tabular}
\caption{Power-law exponents for recall@$k$ and precision@$k$ as the audit budget $k$ varies, fit over the five Qwen-2.5 model sizes.}
\label{tab:budget_sweep}
\end{table*}

\section{GPU Hours}
\label{app:gpu_hours}
We run all experiments on NVIDIA B200 and RTX PRO 6000 GPUs. In total, we estimate about 370 GPU hours for fine-tuning and memorization evaluation, and about 1{,}100 GPU hours for computing base-model features, most of which comes from the 7B and 14B models in our scaling experiments.

\end{document}